\documentclass[10pt,letterpaper]{article}

\usepackage[final,twocolumn]{labpreprint}
\usepackage{mathpazo,tgpagella}
\usepackage{xurl}
\setcitestyle{square}
\AddToHook{env/abstract/after}{\suppressfloats[t]}

\usepackage{cuted} %

\usepackage{currfile} %

\usepackage{amsthm}
\usepackage{placeins}
\usepackage{multirow}
\definecolor{dFourDtableblue}{RGB}{135,200,235}
\newcommand{\dFourDours}[1]{\cellcolor{dFourDtableblue!30}#1}
\newcommand{\dFourDothers}[1]{#1}
\newcommand{\dFourDmodel}[3][0pt]{\multirow[c]{#2}{*}[#1]{\rotatebox[origin=c]{90}{\scriptsize\shortstack{#3}}}}
\definecolor{d4dblue}{rgb}{0.21,0.49,0.74}

\DeclareRobustCommand{\methodname}[1]{\mbox{#1}}


\renewcommand{\dbltopfraction}{0.9}

\title{\methodname{Diff4Diff}: \underline{Diff}erential Activation Quantization for \underline{Diff}usion Transformers}

\author{Philippe Bich, Chiara Boretti, Hyun-Min Chang, Lukas Cavigelli, Lorenz K Muller}

\laborg{Huawei}
\labname{Computing Systems Lab}
\date{}

\makeatletter
\laborg{}
\renewcommand{\lp@titleblock}{%
  \lp@panelheader
  {\sffamily\bfseries\fontsize{21}{25}\selectfont\raggedright
  \methodname{Diff4Diff}: \underline{Diff}erential Activation Quantization\\
  for \underline{Diff}usion Transformers\par}%
  \vspace{10pt}
  {\sffamily\normalfont\small\raggedright\@author\par}%
}
\renewenvironment{abstract}{%
  \lp@openpanel\vspace{16pt}%
  {\sffamily\bfseries\small Abstract\par}\vspace{5pt}%
  \small\parskip=0.5\baselineskip
}{\par\lp@closepanel}

\definecolor{labpanel}{RGB}{252,231,235}
\tcbset{lp@panelstyle/.style={colback=labpanel,colframe=labpanel,boxrule=0pt,arc=5pt,boxsep=0pt,left=\lp@inset,right=\lp@inset,top=\lp@inset,bottom=15pt}}
\renewcommand{\lp@panelheader}{%
  \noindent\begin{minipage}[c]{0.75\linewidth}
  {\sffamily\bfseries\fontsize{11}{14}\selectfont\color{black}Computing Systems Lab}
  \end{minipage}\hfill
  \begin{minipage}[c]{0.2\linewidth}\raggedleft\includegraphics[height=1.15cm]{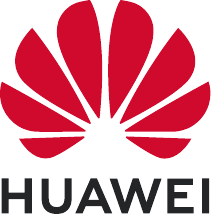}\end{minipage}
  \par\vspace{17pt}}
\makeatother

\begin{document}
\maketitle

\raggedbottom

\begin{abstract}
Diffusion transformers require repeated, computationally intensive denoising evaluations, motivating low-bit weight and activation quantization. Yet low-bit activation quantization can degrade generation quality. We identify a source of this degradation in classifier-free guidance (CFG): existing quantizers process conditional and unconditional branches independently, while CFG amplifies their difference. Quantization errors enter both branches before subtraction and are amplified by the guidance scale. We introduce \methodname{Diff4Diff} (\methodname{D4D}), a quantizer-agnostic activation transformation applied around linear projections. \methodname{D4D} maps paired branch activations to their mean and difference, allowing the quantizer to operate on these components before reconstructing the branches. The transformation leaves full-precision projections unchanged in exact arithmetic. \methodname{D4D} uses a fixed $2\times2$ transform and its inverse, introducing minimal computational overhead without requiring training, calibration data, or estimated statistics for the transformation. The method is compatible with different weight and activation quantizers and requires no high-precision correction path. Our evaluation on image, video, and 3D diffusion transformers shows that \methodname{D4D} improves fidelity to full-precision outputs when combined with existing W4A4 quantizers, reducing LPIPS by up to 11.3\%, 29.6\%, and 52.8\% for images, videos, and rendered 3D surface normals, respectively. Fused kernels introduce less than 0.3\% end-to-end latency overhead in W4A4 image and video pipelines.
Code is available at \url{https://github.com/huawei-csl/MACaQ}.
\end{abstract}
\begin{figure*}[t]
\centering
\includegraphics[width=\textwidth]{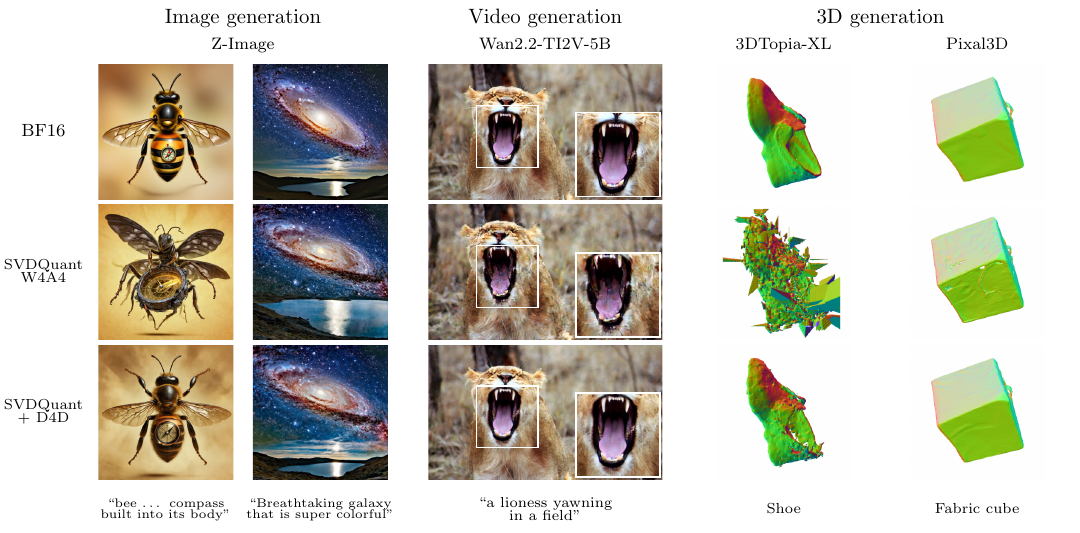}
\captionsetup{hypcap=false}
\caption{\textbf{Image, video, and 3D generation with \methodname{D4D}.} We compare \methodname{SVDQuant} W4A4 with and without \methodname{D4D} against BF16, using the same inputs and seeds for each example. Prompt excerpts are shown below the images. For video, we show the same frame and detail crop across methods; for 3D, colors represent surface normals.}
\label{fig:teaser}
\end{figure*}

\section{Introduction}
\label{sec:introduction}
Diffusion transformers have become a widely used backbone for visual generation, combining scalable architectures with strong generative performance~\cite{peebles2023dit}. They support a broad range of generative tasks, including image synthesis, video generation, and 3D generation~\cite{team2025zimage,wan2025repo,li2026pixal3d,chen2025topia}. However, generating each sample requires repeated evaluations of a large denoising network. Quantizing both weights and activations can reduce memory requirements and accelerate linear projections, but preserving generation quality at very low precision remains challenging~\cite{chen2025qdit,li2025svdquant}.

Recent post-training quantization methods have made substantial progress in preserving diffusion-model quality at low precision~\cite{chen2025qdit,li2025svdquant}. Activation quantization remains particularly sensitive because activation distributions vary across inputs, layers, and denoising steps~\cite{chen2025qdit}. Moreover, the effect of quantization error depends on how the network subsequently combines the quantized signals. Alongside improvements to the quantizer, this motivates examining whether the activation representation exposes the signals that generation needs to preserve.

Classifier-free guidance (CFG) provides a concrete reason to revisit this choice~\cite{ho2022cfg}. CFG combines conditional and unconditional predictions by amplifying their difference. When the branches share a large common component, their difference can be small relative to either branch. Quantizing them separately introduces errors at the scale of the branch activations, potentially obscuring this smaller differential signal. At the output, guidance amplifies the difference between the branch errors. Preserving each branch individually therefore need not preserve their guidance-relevant difference equally well. The relationship between the branches offers structure that can be exploited before quantization.

We introduce \textbf{\methodname{Diff4Diff} (\methodname{D4D})}, an activation transformation that maps paired CFG branches to their mean and difference before quantization. An existing quantizer operates on each component with its own quantization scale. Our analysis of guidance-weighted quantization error explains how separate scales for the mean and difference help preserve the signal amplified by guidance. \methodname{D4D} applies this change of coordinates around shared linear projections and reconstructs the original branches before nonlinear processing. Since both branches use the same projection weights, the transformation preserves the unquantized computation in exact arithmetic.

The same fixed mean--difference transformation can be used with different weight and activation quantizers, without introducing a new high-precision correction path. Its coefficients require no training, calibration data, or estimated statistics; the underlying quantizer retains its own calibration procedure. This separates the choice of activation representation from the quantization method, allowing \methodname{D4D} to complement existing approaches while retaining their weight formats, grouping rules, and low-rank components where applicable.

We evaluate \methodname{D4D} with multiple W4A4 quantizers across image, video, and 3D generation, measuring fidelity to full-precision outputs and complementary generation-quality metrics. Figure~\ref{fig:teaser} illustrates the effect of adding \methodname{D4D} to the same base quantization method. We also provide fused Triton kernels that incorporate both transformations into the surrounding computation. In the evaluated W4A4 image and video pipelines, \methodname{D4D} adds less than 0.3\% end-to-end latency relative to matched implementations without the transformation.

Our contributions are:
\begin{itemize}
 \item We introduce \methodname{D4D}, a quantizer-agnostic activation transformation whose mean--difference coordinates are motivated by the differential structure of CFG.
 \item We formulate \methodname{D4D} as a fixed, invertible transformation around shared linear projections, preserving their computation in exact arithmetic without learning or calibrating the transformation.
 \item We evaluate \methodname{D4D} across image, video, 3D, and speech generation, and provide fused Triton implementations with measured kernel and end-to-end overhead for image and video generation.
\end{itemize}

\section{Method and Theory}
\label{sec:method}
\subsection{Classifier-free guidance and paired activations}
Classifier-free guidance~\cite{ho2022cfg} combines predictions from the same denoiser under two conditioning inputs. Let $v^{\pm}=f_\theta(z_t,t,c^{\pm})$, where $f_\theta$ is the denoiser with parameters $\theta$. The latent $z_t$ and timestep $t$ are shared, while $c^+$ is the target condition and $c^-$ is an unconditional or negative condition. CFG is conventionally expressed in terms of the two branch predictions. To expose the differential signal that motivates \methodname{D4D}, we rewrite it in mean--delta coordinates, for guidance scale $s\geq1$:
\begin{equation}
 g=\underbrace{v^-+s(v^+-v^-)}_{\text{standard CFG}}
  =\underbrace{m_v+\beta d_v}_{\text{our mean--delta reformulation}},
 \label{eq:cfg}
\end{equation}
where $m_v=(v^++v^-)/2$, $d_v=v^+-v^-$, and $\beta=s-1/2$. This reformulation makes the common and guidance-weighted differential components explicit, motivating our choice of activation coordinates within the denoiser. The final guidance computation is unchanged; \methodname{D4D} applies the mean--delta representation locally around linear projections within the denoiser.

\subsection{Mean--delta projection}
\label{sec:reparameterization}
Consider corresponding activation rows $x^+,x^-\in\mathbb R^{1\times K}$ entering a linear projection with weights $\widehat W\in\mathbb R^{K\times O}$ and bias $b$. The weights are the same for both branches and may already be quantized. Let $Q$ denote the base activation quantizer, including scale selection and dequantization. Independent branch quantization produces
\begin{equation}
 \widehat y_{\rm B}^{\pm}=Q(x^\pm)\widehat W+b.
 \label{eq:baseline-projection}
\end{equation}
Each application of $Q$ can select its own scales; independent branch scales are already available to this baseline.

\begin{figure}[t]
\centering
\begingroup
\small
\setlength{\unitlength}{1pt}
\setlength{\fboxsep}{0pt}
\setlength{\fboxrule}{0.4pt}
\newcommand{\dFourDop}[1]{%
  \fcolorbox{black!35}{white}{%
    \parbox[c][22pt][c]{20pt}{\centering\ensuremath{#1}}}}
\newcommand{\dFourDtransform}[1]{%
  \fcolorbox{d4dblue!65}{d4dblue!5}{%
    \parbox[c][22pt][c]{20pt}{\centering\ensuremath{#1}}}}
\begin{picture}(236,110)(0,16)
\put(0,118){\makebox(0,0)[l]{\textbf{(a) Independent branch quantization}}}
\put(13,94){\makebox(0,0){$\begin{pmatrix}x^+\\x^-\end{pmatrix}$}}
\put(117,94){\makebox(0,0){\dFourDop{Q}}}
\put(150,94){\makebox(0,0){\dFourDop{\widehat W}}}
\put(220,94){\makebox(0,0){$\begin{pmatrix}\widehat y^+\\\widehat y^-\end{pmatrix}$}}
\put(29,94){\color{black!65}\vector(1,0){75}}
\put(130,94){\color{black!65}\vector(1,0){7}}
\put(163,94){\color{black!65}\vector(1,0){39}}
\put(0,58){\makebox(0,0)[l]{\textbf{(b) \methodname{D4D} + the same quantizer}}}
\put(13,34){\makebox(0,0){$\begin{pmatrix}x^+\\x^-\end{pmatrix}$}}
\put(45,34){\makebox(0,0){\dFourDtransform{T}}}
\put(81,34){\makebox(0,0){$\begin{pmatrix}m_x\\d_x\end{pmatrix}$}}
\put(117,34){\makebox(0,0){\dFourDop{Q}}}
\put(150,34){\makebox(0,0){\dFourDop{\widehat W}}}
\put(183,34){\makebox(0,0){\dFourDtransform{T^{-1}}}}
\put(220,34){\makebox(0,0){$\begin{pmatrix}\widehat y^+\\\widehat y^-\end{pmatrix}$}}
\put(29,34){\color{black!65}\vector(1,0){3}}
\put(58,34){\color{black!65}\vector(1,0){7}}
\put(97,34){\color{black!65}\vector(1,0){7}}
\put(130,34){\color{black!65}\vector(1,0){7}}
\put(163,34){\color{black!65}\vector(1,0){7}}
\put(196,34){\color{black!65}\vector(1,0){6}}
\end{picture}
\endgroup
\caption{\textbf{Mean--delta quantization around a linear projection.} \methodname{D4D} adds a fixed transform and its inverse (blue) to the original quantized path. The quantizer and weights are unchanged; the original bias is added to the branch outputs.}
\label{fig:implementation}
\end{figure}
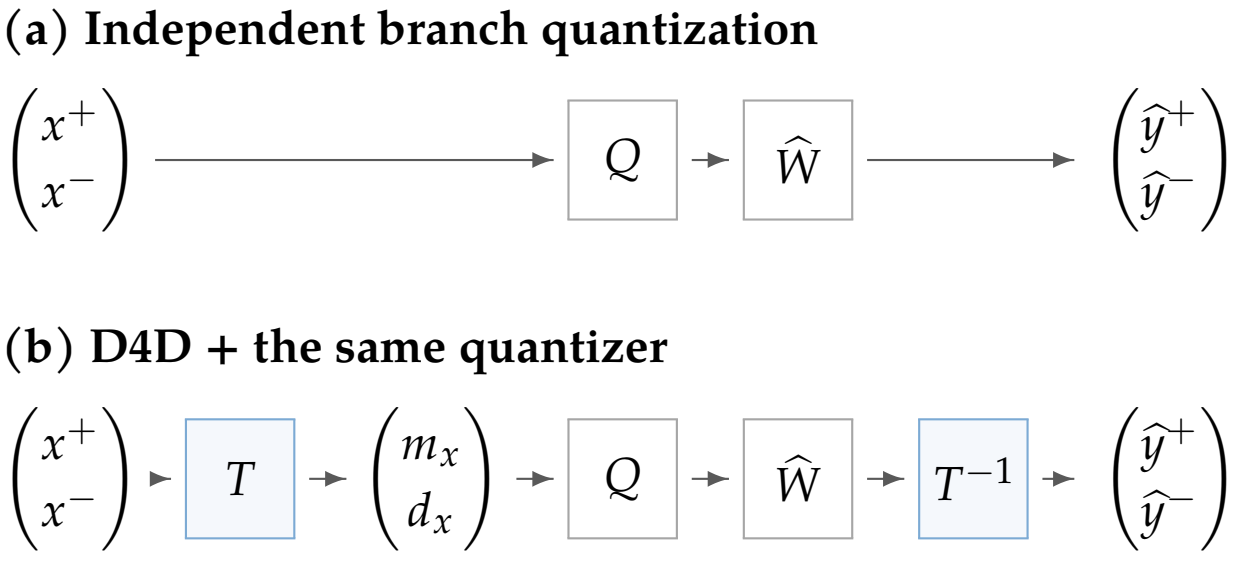

\methodname{D4D} instead forms the mean and difference before quantization:
\begin{equation}
 m_x=\tfrac12(x^++x^-),\qquad d_x=x^+-x^-.
 \label{eq:mean-delta}
\end{equation}
The same base quantizer operates on these two components, using separate scales under its existing grouping rules. After projection, \methodname{D4D} reconstructs the branch outputs as
\begin{equation}
 \widehat y_{\rm D}^{\pm}
 =Q(m_x)\widehat W\ \pm\ \tfrac12Q(d_x)\widehat W+b.
 \label{eq:decode}
\end{equation}
The change is therefore in what is quantized: the mean and delta replace the original branch activations. Their formats and bit widths, and the weight artifact $\widehat W$, remain those of the underlying quantizer. Both components are retained, and branch coordinates are restored before attention, normalization, or nonlinear processing. Figure~\ref{fig:implementation} summarizes the original and \methodname{D4D} projection paths.

\paragraph{Exact change of coordinates.}
Stacking the branch inputs as $X=[x^+;x^-]$, Eq.~\ref{eq:mean-delta} is the fixed transform
\begin{equation}
 T=\begin{pmatrix}1/2&1/2\\1&-1\end{pmatrix},
 \qquad T^{-1}=\begin{pmatrix}1&1/2\\1&-1/2\end{pmatrix}.
 \label{eq:T}
\end{equation}
Since $(TX)\widehat W=T(X\widehat W)$,
\begin{equation}
 T^{-1}\bigl((TX)\widehat W\bigr)=X\widehat W.
 \label{eq:exact}
\end{equation}
Without activation quantization, insertion and inversion leave the projection unchanged in exact arithmetic, including when its weights are quantized. The original bias is added after inversion. This follows the computational-invariance principle used by rotation-based quantization methods such as \methodname{QuaRot}~\cite{ashkboos2024quarot}, which change activation coordinates in channel space while preserving the unquantized computation. \methodname{D4D} makes a principled choice of coordinates in CFG space from the structure of CFG: one row isolates the common component and the other the difference amplified by guidance. The resulting transform acts across the branches and is fixed, invertible, and non-orthogonal; its coefficients follow from the mean--delta decomposition rather than a random rotation or a fitted change of basis.

\subsection{Computational Cost}
\methodname{D4D} adds elementwise encoding and reconstruction around the existing quantized projections. For $N$ paired activation rows with input width $K$ and output width $O$, computing the mean and difference requires $3NK$ arithmetic operations. Reconstructing both branch outputs requires another $3NO$, reusing the scaled differential output. Compared with the approximately $4NKO$ operations of the two branch projections, the relative arithmetic overhead is
\begin{equation}
 \frac{3N(K+O)}{4NKO}
 =\frac34\left(\frac1K+\frac1O\right).
 \label{eq:transform-cost}
\end{equation}
For square projections of width $D$, the relative arithmetic overhead is $3/(2D)$, approximately $0.037\%$ at $D=4096$. Actual latency also depends on memory traffic and kernel fusion.

\methodname{D4D} introduces no additional weight parameters and preserves the number of activation values passed to the quantized projections. Mean and delta can be processed together in a single batched matrix multiplication. Temporary storage and memory traffic depend on how encoding and reconstruction are fused with the surrounding operations. Section~\ref{sec:efficiency} describes our fused implementation and reports measured kernel and end-to-end latency.

\subsection{Guidance and Quantization Error}
\label{sec:bounds}
In this section we compare two linear CFG models analytically: The standard difference formulation 
\begin{equation}
 g_\pm =\bigl[sQ(x^+) +(1-s)Q(x^-)\bigr]\widehat W
 \label{eq:Q_pm}
\end{equation}
and our mean delta formulation
\begin{equation}
g_{md} = \bigl[Q(m_x)+(s-\tfrac12) Q(d_x)\bigr]\widehat W \label{eq:Q_md},
\end{equation}
where again $Q(\cdot)$ is the quantization function. We can either think of this as the minimal version of a CFG denoiser, or as a linear approximation of a larger model. 
This linear model/approximation makes the error-analysis for \methodname{D4D} tractable.

We assume a uniform absmax quantization with step sizes $\Delta_+$, $\Delta_-$, $\Delta_m$, and $\Delta_d$ respectively, and define the ratio $\rho$ between the maximal entry of $x_+$ and $d_x$
\begin{equation}
 \rho=\frac{\lVert d_x\rVert_\infty}{\lVert x^+\rVert_\infty}
      =\frac{\Delta_d}{\Delta_+}.
 \label{eq:main-range-ratio}
\end{equation}
Under the uniform-noise approximation~\cite{gray1998quantization}, each error has zero mean and variance $\Delta_i^2/12$. Assuming uncorrelated errors and comparable branch and mean step-sizes, $\Delta_-\approx\Delta_m\approx\Delta_+$, we obtain a squared-error reduction for uncorrelated rounding errors of
\begin{equation}
 \frac{E^2_{\pm}}{E^2_{md}} \approx G_{\rm uncorr}(\rho)
 =\frac{s^2+(s-1)^2}{1+(s-\tfrac12)^2\rho^2},
 \label{eq:Guncorr}
\end{equation}
where $E^2_{\pm}$is the expected squared error of the standard CFG formulation and $E_{md}^2$ of the mean-delta formulation respectively. A detailed derivation is given in Appendix Sec.~\ref{app:guided-noise}.

When $\rho$ is small (i.e. the infinity-norm of the delta $d_x$ is smaller than that of $x_\pm$) and $s$ is large, this error reduction becomes large. This is the central theoretical argument for the mean-delta basis. 

\begin{figure*}[t]
\centering
\includegraphics[width=\textwidth]{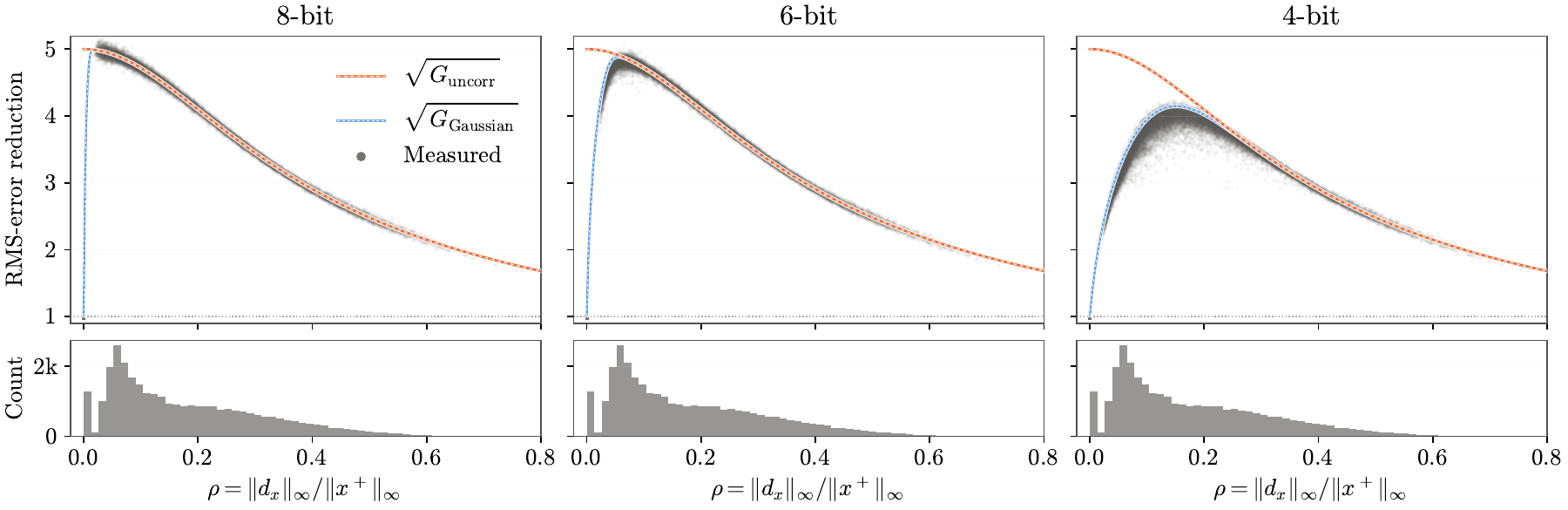}
\caption{\textbf{Mean--delta quantization reduces local guided error.} \methodname{D4D} reduces RMS error relative to quantizing the two branches separately at 8, 6, and 4 bits; values above one indicate an improvement. The curves compare the uncorrelated prediction $\sqrt{G_{\mathrm{uncorr}}}$ (Eq.~\ref{eq:Guncorr}) with the Gaussian correction $\sqrt{G_{\mathrm{Gaussian}}}$ for correlated errors (App.~\ref{app:guided-noise}). The correction follows the measurements more closely at small $\rho$, especially at four bits. Each point corresponds to one prompt, layer, and denoising step, and the histograms show the distribution of $\rho$. We evaluate 25 linear layers of FLUX.2 [klein] 4B Base~\cite{flux2026klein} over 128 prompts and 10 steps, using per-token symmetric absmax \methodname{RTN} with a 32-channel block Hadamard rotation. The empirical RMS error reduction is $\sqrt{G_\text{emp}}$ (Eq.~\ref{eq:gemp}), measured after the shared projection, branch reconstruction, and CFG with $s=4$ (Sec.~\ref{sec:bounds}). Error propagation through the denoiser is examined in Appendix~\ref{app:propagation}.}
\label{fig:guided-error-gain}
\end{figure*}

In Figure~\ref{fig:guided-error-gain} we compare the predicted RMS reduction, $\sqrt{G_{\rm uncorr}}$, with measurements from FLUX-2 Klein 4B. We take inputs to linear layers in the DiT at different depths and denoising steps and evaluate the empirical error reduction as 
\begin{equation}
G_\text{emp} = \frac{ \left\| \left.g_\pm\right|_{Q=\mathrm{RTN}} - \left.g_\pm\right|_{Q=I} \right\|_2^2 }{ \left\| \left.g_{md}\right|_{Q=\mathrm{RTN}} - \left.g_{md}\right|_{Q=I} \right\|_2^2 } \label{eq:gemp}
\end{equation}
where $I$ denotes the identity (no activation quantization), and the subscripts $Q=\mathrm{RTN}$ and $Q=I$ specify the activation quantizer used.

In practice, rounding errors may be correlated. The Gaussian correction shown in the figure accounts for this effect; Appendix~\ref{app:guided-noise} explains this.  

Of course, in the real model the CFG mixing only happens at the end, not after each layer; in a full DiT the outputs of linear layers propagate through attention, normalization, and nonlinearities before being combined in the CFG formulation. Appendix~\ref{app:propagation} examines this further: the observed benefit is larger in later layers, consistent with a shorter propagation path to the final CFG readout. 

\section{Experiments}
\label{sec:experiments}
We test \methodname{D4D} on image, video, and 3D generation with several models and quantization methods. For each method, we compare W4A4 generation with and without \methodname{D4D} using the same inputs, seeds, and sampling settings. We measure how closely the generated outputs reproduce BF16 results and evaluate their quality using metrics appropriate to each task. We also compare alternative branch coordinates and measure the latency added by \methodname{D4D}, both for individual projections and for complete generation pipelines. Calibration settings and implementation details are given in Appendix~\ref{app:protocols}. We additionally evaluate speech generation with CosyVoice2~\cite{du2024cosyvoice2} on LibriTTS~\cite{zen2019libritts}; results and evaluation details are provided in Appendix~\ref{app:audio}.

\subsection{Image Generation}
We evaluate Z-Image and Chroma1-HD~\cite{rock2025chroma} on MJHQ-30K~\cite{li2024playground}, also used in prior work~\cite{li2025svdquant,park2026diversevar}. Our category-balanced evaluation comprises 1,000 prompts shared across all methods; sampling details are provided in Appendix~\ref{app:image-protocol}. We compare \methodname{SVDQuant}~\cite{li2025svdquant}, \methodname{Q-DiT}~\cite{chen2025qdit}, and \methodname{MACaQ}~\cite{muller2026macaq} at W4A4, each with and without \methodname{D4D}. PSNR and LPIPS~\cite{zhang2018lpips} measure fidelity to BF16 images generated with the same prompts and seeds, while ImageReward~\cite{xu2023imagereward} assesses image quality and alignment with the prompt. Quantizer configurations are detailed in Appendix~\ref{app:protocols}.

\begin{figure}[t]
\centering
\includegraphics[width=\columnwidth]{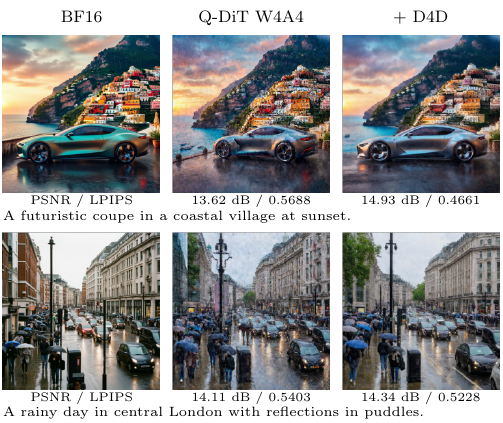}
\caption{\textbf{Image generation with \methodname{D4D}.} Each row compares Z-Image outputs with BF16, \methodname{Q-DiT} W4A4, and \methodname{Q-DiT} + \methodname{D4D}, using the same prompt and seed. PSNR (dB, higher is better) and LPIPS (lower is better) are measured relative to BF16.}
\label{fig:qualitative-main}
\label{fig:qualitative-images-main}
\end{figure}

Table~\ref{tab:images} and Figure~\ref{fig:qualitative-images-main} show improvements in PSNR, LPIPS, and ImageReward across all six model--quantizer pairs. On Z-Image, adding \methodname{D4D} reduces LPIPS by 6.2\%, 11.3\%, and 10.5\% with \methodname{SVDQuant}, \methodname{Q-DiT}, and \methodname{MACaQ}, respectively. ImageReward also increases for all three methods, with \methodname{Q-DiT} improving from 0.5153 to 0.7547. On Chroma1-HD, \methodname{D4D} improves all three metrics with each quantizer. These results show that the benefits of mean--difference coordinates extend across different quantization methods, improving fidelity to BF16 outputs alongside image preference scores.

\begin{table}[!htbp]
\centering\small
\setlength{\tabcolsep}{2pt}
\caption{\textbf{Image generation.} Each model uses MJHQ-1K for all methods. IR: ImageReward; LPIPS and PSNR (dB) compare with BF16. Bold indicates the better score within each without/with-D4D pair. Entries are mean scores; paired uncertainty is reported in Appendix~\ref{app:paired-quality}. All quantized configurations use W4A4.}
\label{tab:images}
\begin{tabular}{clcccc}
\toprule
\textbf{Model} & \textbf{Base method} & \textbf{W/A} & \textbf{PSNR $\uparrow$} & \textbf{IR $\uparrow$} & \textbf{LPIPS $\downarrow$}\\
\midrule

\dFourDmodel[\dimexpr-3\aboverulesep/2-3\belowrulesep/2-3\cmidrulewidth/2\relax]{7}{Z-Image} & \textit{BF16} & \textit{16/16} & -- & \textit{0.8945} & --\\
\cmidrule(l){2-6}
 & \dFourDothers{\methodname{SVDQuant}} & \dFourDothers{4/4} & \dFourDothers{15.165} & \dFourDothers{0.8435} & \dFourDothers{0.4958}\\
 & \dFourDours{\textbf{+ \methodname{D4D}}} & \dFourDours{4/4} & \dFourDours{\textbf{15.390}} & \dFourDours{\textbf{0.8703}} & \dFourDours{\textbf{0.4650}}\\
\cmidrule(l){2-6}
 & \dFourDothers{\methodname{Q-DiT}} & \dFourDothers{4/4} & \dFourDothers{14.175} & \dFourDothers{0.5153} & \dFourDothers{0.6341}\\
 & \dFourDours{\textbf{+ \methodname{D4D}}} & \dFourDours{4/4} & \dFourDours{\textbf{14.640}} & \dFourDours{\textbf{0.7547}} & \dFourDours{\textbf{0.5626}}\\
\cmidrule(l){2-6}
 & \dFourDothers{\methodname{MACaQ}} & \dFourDothers{4/4} & \dFourDothers{15.694} & \dFourDothers{0.8044} & \dFourDothers{0.4958}\\
 & \dFourDours{\textbf{+ \methodname{D4D}}} & \dFourDours{4/4} & \dFourDours{\textbf{16.095}} & \dFourDours{\textbf{0.8298}} & \dFourDours{\textbf{0.4437}}\\
\midrule

\dFourDmodel[\dimexpr-3\aboverulesep/2-3\belowrulesep/2-3\cmidrulewidth/2\relax]{7}{Chroma1-HD} & \textit{BF16} & \textit{16/16} & -- & \textit{0.7090} & --\\
\cmidrule(l){2-6}
 & \dFourDothers{\methodname{SVDQuant}} & \dFourDothers{4/4} & \dFourDothers{21.086} & \dFourDothers{0.6606} & \dFourDothers{0.2638}\\
 & \dFourDours{\textbf{+ \methodname{D4D}}} & \dFourDours{4/4} & \dFourDours{\textbf{21.269}} & \dFourDours{\textbf{0.6705}} & \dFourDours{\textbf{0.2566}}\\
\cmidrule(l){2-6}
 & \dFourDothers{\methodname{Q-DiT}} & \dFourDothers{4/4} & \dFourDothers{18.665} & \dFourDothers{0.6243} & \dFourDothers{0.3457}\\
 & \dFourDours{\textbf{+ \methodname{D4D}}} & \dFourDours{4/4} & \dFourDours{\textbf{18.920}} & \dFourDours{\textbf{0.6424}} & \dFourDours{\textbf{0.3401}}\\
\cmidrule(l){2-6}
 & \dFourDothers{\methodname{MACaQ}} & \dFourDothers{4/4} & \dFourDothers{20.514} & \dFourDothers{0.6179} & \dFourDothers{0.2948}\\
 & \dFourDours{\textbf{+ \methodname{D4D}}} & \dFourDours{4/4} & \dFourDours{\textbf{20.629}} & \dFourDours{\textbf{0.6301}} & \dFourDours{\textbf{0.2912}}\\
\bottomrule
\specialrule{\heavyrulewidth}{1.5pt}{0pt}
\end{tabular}

\end{table}

\subsection{Video Generation}
\begin{figure*}[t]
\centering
\includegraphics[width=\textwidth]{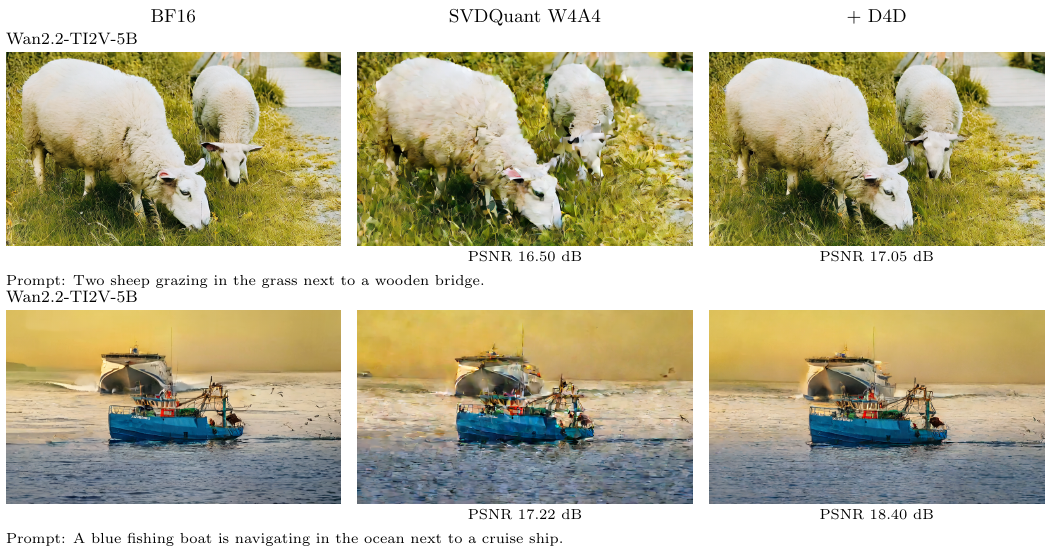}
\caption{\textbf{Video generation with \methodname{SVDQuant} and \methodname{D4D}.} Two Wan2.2-TI2V-5B examples comparing BF16, \methodname{SVDQuant} W4A4, and \methodname{SVDQuant} W4A4 + \methodname{D4D}. Each example shows one frame, with PSNR averaged over the full generated sequence relative to BF16. Prompts appear below each example. Without D4D frames from the quantized model have a granular texture artifact that D4D removes. }
\label{fig:qualitative-video-main}
\end{figure*}

We evaluate Wan2.2-TI2V-5B and Wan2.2-I2V-A14B on image-to-video generation using a subset of VBench-I2V~\cite{huang2024vbench,huang2024vbenchpp}, conditioning each video on an input image and its accompanying text prompt. All methods use the same image--prompt pairs. Subset selection and sampling details are provided in Appendix~\ref{app:protocols}. Video baselines are \methodname{SVDQuant}, \methodname{Q-DiT}, and \methodname{MACaQ}, each evaluated with and without \methodname{D4D}. On Wan TI2V-5B, \methodname{D4D} reduces LPIPS by 29.6\%, 26.4\%, and 19.0\% with \methodname{SVDQuant}, \methodname{Q-DiT}, and \methodname{MACaQ}, respectively, while improving PSNR and I2V subject preservation for all three baselines (Table~\ref{tab:video}, Figure~\ref{fig:qualitative-video-main}).

\begin{table}[!htbp]
\centering\small
\setlength{\tabcolsep}{2pt}
\caption{\textbf{Image-to-video generation.} Both Wan models use the same 128 VBench-I2V conditions. I2V-S measures subject preservation; PSNR (dB) and LPIPS compare with BF16. Bold marks the better score within each without/with-D4D pair. All quantized configurations use W4A4. Full metrics and paired uncertainty appear in Appendix~\ref{app:extended-results}.}
\label{tab:video}
\begin{tabular}{clcccc}
\toprule
\textbf{Model} & \textbf{Base method} & \textbf{W/A} & \textbf{PSNR $\uparrow$} & \textbf{I2V-S $\uparrow$} & \textbf{LPIPS $\downarrow$}\\
\midrule
\dFourDmodel[\dimexpr-3\aboverulesep/2-3\belowrulesep/2-3\cmidrulewidth/2\relax]{7}{Wan2.2\\TI2V-5B} & \textit{BF16} & \textit{16/16} & \textit{--} & \textit{0.9770} & \textit{--}\\
\cmidrule(l){2-6}
 & \dFourDothers{\methodname{SVDQuant}} & \dFourDothers{4/4} & \dFourDothers{19.013} & \dFourDothers{0.9638} & \dFourDothers{0.2616}\\
 & \dFourDours{\textbf{+ \methodname{D4D}}} & \dFourDours{4/4} & \dFourDours{\textbf{20.715}} & \dFourDours{\textbf{0.9751}} & \dFourDours{\textbf{0.1843}}\\
\cmidrule(l){2-6}
 & \dFourDothers{\methodname{Q-DiT}} & \dFourDothers{4/4} & \dFourDothers{18.635} & \dFourDothers{0.9575} & \dFourDothers{0.2972}\\
 & \dFourDours{\textbf{+ \methodname{D4D}}} & \dFourDours{4/4} & \dFourDours{\textbf{19.914}} & \dFourDours{\textbf{0.9723}} & \dFourDours{\textbf{0.2187}}\\
\cmidrule(l){2-6}
 & \dFourDothers{\methodname{MACaQ}} & \dFourDothers{4/4} & \dFourDothers{18.663} & \dFourDothers{0.9686} & \dFourDothers{0.2574}\\
 & \dFourDours{\textbf{+ \methodname{D4D}}} & \dFourDours{4/4} & \dFourDours{\textbf{19.900}} & \dFourDours{\textbf{0.9791}} & \dFourDours{\textbf{0.2085}}\\
\midrule
\dFourDmodel[\dimexpr-3\aboverulesep/2-3\belowrulesep/2-3\cmidrulewidth/2\relax]{7}{Wan2.2\\I2V-A14B} & \textit{BF16} & \textit{16/16} & \textit{--} & \textit{0.9776} & \textit{--}\\
\cmidrule(l){2-6}
 & \dFourDothers{\methodname{SVDQuant}} & \dFourDothers{4/4} & \dFourDothers{19.810} & \dFourDothers{0.9585} & \dFourDothers{0.2102}\\
 & \dFourDours{\textbf{+ \methodname{D4D}}} & \dFourDours{4/4} & \dFourDours{\textbf{20.257}} & \dFourDours{\textbf{0.9744}} & \dFourDours{\textbf{0.1928}}\\
\cmidrule(l){2-6}
 & \dFourDothers{\methodname{Q-DiT}} & \dFourDothers{4/4} & \dFourDothers{19.329} & \dFourDothers{0.9533} & \dFourDothers{0.2311}\\
 & \dFourDours{\textbf{+ \methodname{D4D}}} & \dFourDours{4/4} & \dFourDours{\textbf{19.790}} & \dFourDours{\textbf{0.9689}} & \dFourDours{\textbf{0.2071}}\\
\cmidrule(l){2-6}
 & \dFourDothers{\methodname{MACaQ}} & \dFourDothers{4/4} & \dFourDothers{19.548} & \dFourDothers{0.9617} & \dFourDothers{0.2164}\\
 & \dFourDours{\textbf{+ \methodname{D4D}}} & \dFourDours{4/4} & \dFourDours{\textbf{19.780}} & \dFourDours{\textbf{0.9737}} & \dFourDours{\textbf{0.2116}}\\
\bottomrule
\specialrule{\heavyrulewidth}{1.5pt}{0pt}
\end{tabular}
\end{table}

\subsection{3D Generation}
We evaluate Pixal3D and 3DTopia-XL on a shared subset of GSO~\cite{downs2022gso}, using \methodname{SVDQuant} and \methodname{MACaQ} at W4A4, each with and without \methodname{D4D}. We adapt \methodname{SVDQuant} to both architectures and mark this adaptation with an asterisk in Table~\ref{tab:3d}. PSNR and LPIPS compare surface-normal maps rendered from the quantized and BF16 outputs under the same viewpoints, while Chamfer distance measures differences between their surfaces. Subset size and selection, calibration, and rendering settings are detailed in Appendix~\ref{app:protocols}.

\begin{figure*}[t]
\centering
\includegraphics[width=\textwidth]{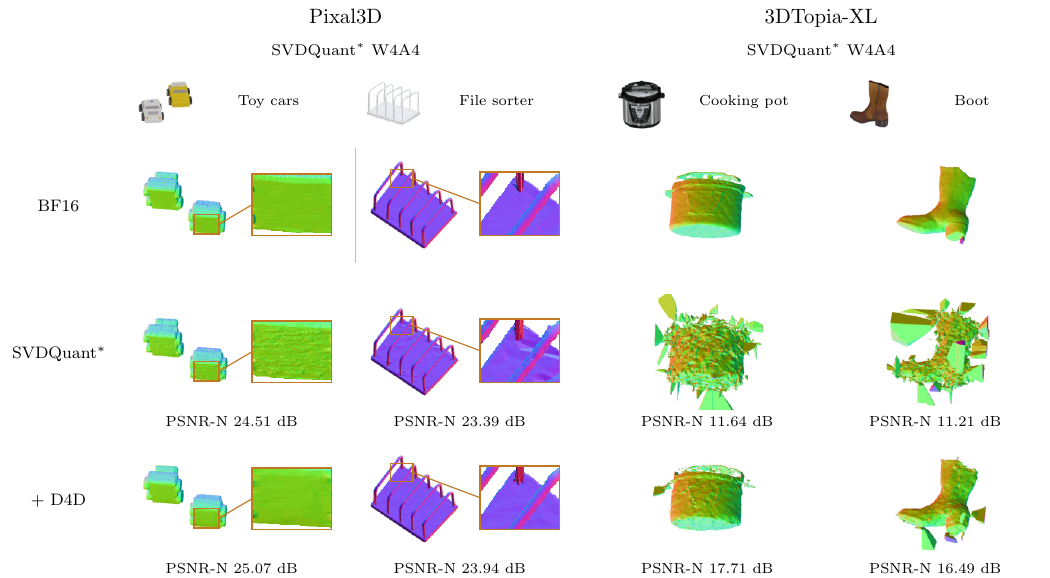}
\caption{\textbf{3D generation with \methodname{SVDQuant} and \methodname{D4D}.} Two objects per model, with conditioning images above and the same viewpoint across BF16, \methodname{SVDQuant}, and \methodname{SVDQuant} + \methodname{D4D}. Colors represent surface normals. Insets show the toy-car underside and file-sorter base, using the same crop and magnification across methods. PSNR-N is averaged over all evaluation views of each object relative to BF16; $^{*}$denotes our \methodname{SVDQuant} adaptation.}
\label{fig:qualitative-3d-main}
\end{figure*}

Table~\ref{tab:3d} and Figure~\ref{fig:qualitative-3d-main} show that \methodname{D4D} improves normal-map fidelity across both models and quantizers. On 3DTopia-XL, adding \methodname{D4D} reduces LPIPS by 52.8\% with \methodname{SVDQuant} and 36.8\% with \methodname{MACaQ}, with corresponding PSNR gains of 4.96 and 4.06 dB. Both methods also produce geometry closer to the BF16 reference. On Pixal3D, LPIPS decreases by 19.0\% and 10.4\%, respectively. These results show that the benefits of \methodname{D4D} extend to 3D generation, with the largest improvements on 3DTopia-XL. Additional geometry metrics and paired confidence intervals are reported in Appendix~\ref{app:3d-results}.

\begin{table}[!htbp]
\centering\small
\setlength{\tabcolsep}{1.5pt}
\caption{\textbf{Image-to-3D generation on GSO-150.} PSNR-N (dB) and LPIPS-N compare rendered surface normals; CD is symmetric unsquared surface distance. All metrics use matched BF16 outputs as reference. $^{*}$Our adaptation of \methodname{SVDQuant}. Quantized configurations use W4A4.}
\label{tab:3d}
\begin{tabular}{cllccc}
\toprule
\textbf{Model} & \textbf{Base method} & \textbf{W/A} & \textbf{PSNR-N$\uparrow$} & \textbf{LPIPS-N$\downarrow$} & \textbf{CD$\downarrow$}\\
\midrule
\dFourDmodel{5}{Pixal3D} & \textit{BF16} & \textit{16/16} & -- & -- & --\\
\cmidrule(l){2-6}
 & \dFourDothers{\methodname{SVDQuant}$^{*}$} & \dFourDothers{4/4} & \dFourDothers{24.25} & \dFourDothers{0.0717} & \dFourDothers{0.005412}\\
 & \dFourDours{\textbf{+ \methodname{D4D}}} & \dFourDours{4/4} & \dFourDours{\textbf{24.79}} & \dFourDours{\textbf{0.0581}} & \dFourDours{\textbf{0.005266}}\\
\cmidrule(l){2-6}
 & \dFourDothers{\methodname{MACaQ}} & \dFourDothers{4/4} & \dFourDothers{16.82} & \dFourDothers{0.2827} & \dFourDothers{0.008781}\\
 & \dFourDours{\textbf{+ \methodname{D4D}}} & \dFourDours{4/4} & \dFourDours{\textbf{17.20}} & \dFourDours{\textbf{0.2534}} & \dFourDours{\textbf{0.008780}}\\
\midrule
\dFourDmodel{5}{3DTopia-XL} & \textit{BF16} & \textit{16/16} & -- & -- & --\\
\cmidrule(l){2-6}
 & \dFourDothers{\methodname{SVDQuant}$^{*}$} & \dFourDothers{4/4} & \dFourDothers{11.30} & \dFourDothers{0.5220} & \dFourDothers{0.052781}\\
 & \dFourDours{\textbf{+ \methodname{D4D}}} & \dFourDours{4/4} & \dFourDours{\textbf{16.26}} & \dFourDours{\textbf{0.2463}} & \dFourDours{\textbf{0.040248}}\\
\cmidrule(l){2-6}
 & \dFourDothers{\methodname{MACaQ}} & \dFourDothers{4/4} & \dFourDothers{7.80} & \dFourDothers{0.7506} & \dFourDothers{0.066646}\\
 & \dFourDours{\textbf{+ \methodname{D4D}}} & \dFourDours{4/4} & \dFourDours{\textbf{11.86}} & \dFourDours{\textbf{0.4744}} & \dFourDours{\textbf{0.050159}}\\
\bottomrule
\specialrule{\heavyrulewidth}{1.5pt}{0pt}
\end{tabular}
\end{table}

\subsection{Key Ablations}
Table~\ref{tab:coordinates} compares quantization with different branch coordinates on Z-Image and 3DTopia-XL under matched conditions. Mean--anchor replaces the difference with an original branch, while anchor--delta replaces the mean. D4D explicitly represents the branch mean and difference, using a non-orthogonal normalization rather than an orthogonal Hadamard transform. A fixed-direction ablation further shows that D4D’s normalization can improve reconstruction quality under W4A4 quantization (Appendix~\ref{app:coordinate-scaling}). All variants retaining the difference improve both metrics over the original coordinates on both models, while mean--anchor degrades fidelity. D4D achieves the highest PSNR and lowest LPIPS on Z-Image, and the highest PSNR-N and lowest LPIPS-N on 3DTopia-XL. To examine how this benefit depends on the amplification of the branch difference, we also vary CFG guidance strength with quantized weights held fixed (Appendix~\ref{app:guidance-strength}).

\begin{table}[!htbp]
\centering\small
\setlength{\tabcolsep}{2pt}
\caption{\textbf{Choice of branch coordinates.} \methodname{SVDQuant} on Z-Image and its adapted implementation on 3DTopia-XL. All rows use W4A4; Direct quantizes the two branches without a coordinate transform. PSNR and LPIPS compare with BF16 images; PSNR-N and LPIPS-N compare rendered surface normals. Z-Image uses the MJHQ-1K subset from Table~\ref{tab:images}. 3DTopia-XL uses the GSO subset from Table~\ref{tab:3d}.}
\label{tab:coordinates}
\resizebox{\linewidth}{!}{%
\begin{tabular}{lcccc}
\toprule
& \multicolumn{2}{c}{\textbf{Z-Image}} & \multicolumn{2}{c}{\textbf{3DTopia-XL}}\\
\cmidrule(lr){2-3}\cmidrule(lr){4-5}
\textbf{Coordinates} & \textbf{PSNR$\uparrow$} & \textbf{LPIPS$\downarrow$} & \textbf{PSNR-N$\uparrow$} & \textbf{LPIPS-N$\downarrow$}\\
\midrule
\dFourDothers{Direct (no transform)} & \dFourDothers{15.17} & \dFourDothers{0.4958} & \dFourDothers{11.30} & \dFourDothers{0.5220}\\
\midrule
\dFourDothers{Mean--anchor} & \dFourDothers{15.08} & \dFourDothers{0.6943} & 8.61  & 0.6978\\[3pt]
\dFourDothers{Anchor--delta} & \dFourDothers{15.29} & \dFourDothers{0.4863} & 15.98 & 0.2810\\[3pt]
\dFourDothers{Hadamard} & \dFourDothers{15.28} & \dFourDothers{0.4869} & 15.68 & 0.3043\\[3pt]
\dFourDours{\textbf{\methodname{D4D}}} & \dFourDours{\textbf{15.39}} & \dFourDours{\textbf{0.4650}} & \dFourDours{\textbf{16.26}} & \dFourDours{\textbf{0.2463}}\\[3pt]
\bottomrule
\specialrule{\heavyrulewidth}{1.5pt}{0pt}
\end{tabular}}
\end{table}

\subsection{Efficiency}
\label{sec:efficiency}
The arithmetic count in Sec.~\ref{sec:method} does not capture memory traffic or kernel-launch costs. We therefore evaluate a fused implementation, first at the projection level and then in complete generation pipelines, on a recent GPU. Hardware and measurement details are given in Appendix~\ref{app:efficiency}.

\begin{figure}[t]
\centering
\includegraphics[width=\linewidth]{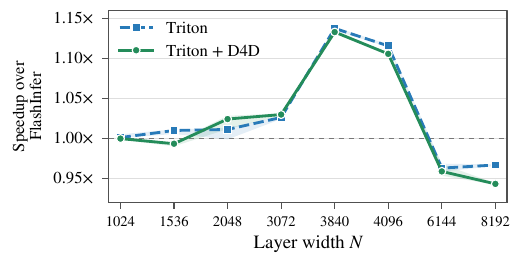}
\caption{\textbf{\methodname{D4D} projection overhead.} Matched Triton kernels without and with \methodname{D4D}, normalized to the optimized FlashInfer reference ($1\times$; higher is faster). Square NVFP4 projections with 4096 tokens per CFG branch, corresponding to a batch-one workload. Timings include packing, GEMM, and \methodname{D4D} transforms where applicable; shading shows 95\% bootstrap intervals.}
\label{fig:kernel-speedup}
\end{figure}

\paragraph{Fused projection kernels.}
We implement \methodname{D4D} in Triton~\cite{tillet2019triton}, forming mean--delta coordinates during activation packing and reconstructing the two branches in the GEMM epilogue. This avoids separate transform passes and additional matrix multiplications. Our primary comparison is between matched Triton implementations with and without \methodname{D4D}: both use the same kernel configuration, so their latency difference isolates the cost of the transformation. Figure~\ref{fig:kernel-speedup} reports this comparison for square NVFP4 projections, timing activation packing, GEMM, and, for \methodname{D4D}, encoding and reconstruction. Across the measured width sweep, \methodname{D4D} adds at most 2.51\% latency. To establish that this overhead is measured on a competitive implementation, we normalize both curves to FlashInfer's autotuned NVFP4 path~\cite{ye2025flashinfer}. Its $1\times$ reference provides an external performance check: the untransformed Triton path performs comparably to this optimized library implementation, making it a meaningful baseline for assessing \methodname{D4D}'s incremental cost. Hardware, configuration selection, and timing details are given in Appendix~\ref{app:efficiency}.

\paragraph{End-to-end efficiency.}
We also measure end-to-end image and video generation latency. Table~\ref{tab:efficiency} reports latency for Z-Image and Wan-5B using NVFP4 and BF16. Adding \methodname{D4D} to the matched Triton implementation increases generation time by 0.17\% and 0.21\% in NVFP4, and by 0.49\% and 1.40\% in BF16, respectively. These measurements include both the forward and inverse transformations. The optimized library reference contextualizes absolute latency; the matched Triton comparison isolates \methodname{D4D}'s overhead. Overhead stays below 1.5\% across both precisions. The Triton kernels and benchmark scripts are included in the supplementary material.

\begin{table}[!htbp]
\centering\small
\setlength{\tabcolsep}{2pt}
\caption{\textbf{End-to-end GPU inference latency.} Mean generation time in seconds over 20 prompts with three repetitions per prompt. Library denotes the optimized reference implementation; Triton and Triton + \methodname{D4D} use matched execution configurations. $\Delta$ is the overhead relative to Triton without \methodname{D4D}.}
\label{tab:efficiency}
\begin{tabular}{llrrrr}
\toprule
 & & & \multicolumn{2}{c}{\textbf{Triton}} & \\
\cmidrule(lr){4-5}
\textbf{Model} & \textbf{Format} & \textbf{Library} & \textbf{w/o \methodname{D4D}} & \dFourDours{\textbf{w/ \methodname{D4D}}} & \textbf{$\Delta$ (\%)}\\
\midrule
\multirow{2}{*}{Z-Image} & NVFP4 & 9.091 & 9.064 & \dFourDours{9.080} & +0.17\\
 & BF16 & 16.526 & 17.149 & \dFourDours{17.233} & +0.49\\
\midrule
\multirow{2}{*}{Wan 5B} & NVFP4 & 19.236 & 18.224 & \dFourDours{18.262} & +0.21\\
 & BF16 & 24.288 & 25.402 & \dFourDours{25.757} & +1.40\\
\bottomrule
\specialrule{\heavyrulewidth}{1.5pt}{0pt}
\end{tabular}
\end{table}

\section{Related Work}
\label{sec:related}
Post-training quantization for diffusion models must account for activation distributions that change throughout denoising. \methodname{Q-Diffusion}~\cite{li2023qdiffusion} adapts calibration and quantization to this temporal variation and the structure of diffusion networks, while \methodname{PTQD}~\cite{he2023ptqd} analyzes quantization noise within the sampling process and introduces corrections to its correlated and residual components. Subsequent methods address the channel outliers and temporal variability of diffusion transformers. \methodname{PTQ4DiT}~\cite{wu2024ptq4dit} combines channel salience balancing with timestep-aware calibration; \methodname{Q-DiT}~\cite{chen2025qdit} introduces adaptive quantization granularity and dynamic activation quantization; and \methodname{ViDiT-Q}~\cite{zhao2024viditq} develops quantization strategies for image and video transformers. \methodname{SVDQuant}~\cite{li2025svdquant} absorbs outliers into a high-precision low-rank branch, making the remaining computation more amenable to four-bit quantization. These approaches primarily address how weights and activations are quantized within individual projections. A complementary line of work changes their representation while preserving the full-precision computation. \methodname{SmoothQuant}~\cite{xiao2023smoothquant} redistributes activation magnitudes through equivalent channel rescaling. \methodname{SINQ}~\cite{muller2026sinq} normalizes weight matrices along both rows and columns using Sinkhorn-style scaling, enabling calibration-free low-precision weight quantization. \methodname{QuaRot}~\cite{ashkboos2024quarot} applies feature-space rotations to suppress outliers, and \methodname{SpinQuant}~\cite{liu2025spinquant} learns rotations that improve quantized accuracy. \methodname{D4D} operates along the conditional--unconditional branch axis, where CFG explicitly amplifies the difference between paired activations at the network output. Concurrent work, \methodname{GCBT}~\cite{jiang2026gcbt}, also investigates quantization in branch space. It calibrates layer-specific orthogonal rotations from cross-branch second moments and the guidance direction. \methodname{D4D} instead fixes the representation to the mean and difference of the branches, following the structure of CFG directly rather than fitting the coordinate transform to activation statistics. Additional comparisons with \methodname{GCBT} are provided in Appendix~\ref{app:concurrent}. Our study examines its effect on guided quantization error and its applicability across image, video, and 3D generation, together with fused implementations that limit the additional computational cost.

\section{Conclusion}
We introduced \methodname{Diff4Diff}, a quantizer-agnostic transformation that preserves the differential signal amplified by classifier-free guidance. Quantizing paired activations in mean--difference coordinates gives this signal its own scale while preserving the full-precision computation. Our analysis motivates this choice, and experiments across image, video, and 3D generation show improved fidelity to BF16 outputs with existing W4A4 quantizers. The fixed transformation requires no calibration of its own. Fused Triton kernels add less than 0.3\% end-to-end latency in the evaluated W4A4 image and video pipelines.

\FloatBarrier
\setcounter{dbltopnumber}{2}
\renewcommand{\dbltopfraction}{0.99}

\clearpage
\appendix
\twocolumn[{\centering\sffamily\Large\bfseries Supplementary Material\par\vspace{1em}}]
\nocite{radford2023whisper}
\captionsetup{font=footnotesize}
\setlength{\floatsep}{8pt}
\setlength{\textfloatsep}{10pt}
\setlength{\intextsep}{8pt}
\setlength{\dblfloatsep}{6pt}
\setlength{\dbltextfloatsep}{10pt}
\makeatletter
\setlength{\@fpsep}{8pt}
\setlength{\@dblfpsep}{6pt}
\makeatother

\paragraph{Overview.}
App.~\ref{sec:app_error} expands the error analysis and checks its premise on the evaluated models (App.~\ref{sec:app_branch_stats}). App.~\ref{app:protocols} gives the experimental protocols, including the coordinates used for calibration. App.~\ref{app:audio} reports speech generation. App.~\ref{app:extended-results} reports paired statistics, a summary over all settings (Tab.~\ref{tab:summary-gains}) and per-prompt win rates (Tab.~\ref{tab:winrates}). Apps.~\ref{app:coordinate-scaling}--\ref{app:concurrent} cover coordinate scaling, guidance strength, efficiency and GCBT, including the rotation angles GCBT selects (Tab.~\ref{tab:gcbt-angles}). App.~\ref{app:qualitative} shows qualitative examples. App.~\ref{app:limitations} states the scope and limitations of D4D.

\section{Quantization Error Analysis}
\label{sec:app_error}

This section expands the error analysis of Sec.~\ref{sec:bounds}. We derive the uncorrelated prediction $G_{\mathrm{uncorr}}$ (Sec.~\ref{sec:app_derivation}), describe how the measurements in Fig.~\ref{fig:guided-error-gain} relate to it (Sec.~\ref{sec:app_measurement}), derive the Gaussian correction for correlated rounding errors (Sec.~\ref{sec:app_gaussian}), and examine how the reduction carries through the full denoiser (Sec.~\ref{sec:app_propagation}).

\subsection{Derivation of the uncorrelated prediction}
\label{sec:app_derivation}
\label{app:guided-noise}

For an activation row $u$, let $Q(u)$ denote its quantized-and-dequantized value and $e_u=Q(u)-u$ its quantization error. We write $e_\pm$, $e_m$ and $e_d$ for the errors of $x^\pm$, $m_x$ and $d_x$. Subtracting the unquantized outputs from Eqs.~\ref{eq:Q_pm} and~\ref{eq:Q_md}, the two guided outputs differ from the exact one by $e_B\widehat{W}$ and $e_D\widehat{W}$, with
\begin{equation}
e_B = s\,e_+ + (1-s)\,e_-, \quad e_D = e_m + \beta\,e_d,\quad \beta=s-\tfrac12 .
\label{eq:app_errors}
\end{equation}
The two expressions already show the mechanism. Independent quantization multiplies both branch errors by guidance weights of magnitude $s$ and $s-1$. Mean--delta quantization amplifies only the error of the difference, while the error of the mean enters with weight one.

\paragraph{Step sizes.}
An absmax quantizer with $b$ bits maps the largest entry of a row (or group) to $q_{\max}=2^{b-1}-1$, so its step is $\Delta=\|u\|_\infty/q_{\max}$. The four components therefore have steps $\Delta_+$, $\Delta_-$, $\Delta_m$ and $\Delta_d$, and the ratio of steps equals the ratio of infinity norms used in Eq.~\ref{eq:main-range-ratio}:
\begin{equation}
\rho=\frac{\Delta_d}{\Delta_+}=\frac{\|d_x\|_\infty}{\|x^+\|_\infty}.
\label{eq:app_rho_single}
\end{equation}
The bit width cancels, so $\rho$ is a property of the activations alone.

\paragraph{Uniform noise.}
Under the uniform-noise approximation~\cite{gray1998quantization}, each component of $e_i$ has zero mean and variance $v_i=\Delta_i^2/12$, for $i\in\{+,-,m,d\}$. If the errors of the paired components are uncorrelated, the per-coordinate mean squared errors of Eq.~\ref{eq:app_errors} are
\begin{equation}
D_B = s^2 v_+ + (s-1)^2 v_-, \qquad D_D = v_m + \beta^2 v_d .
\label{eq:app_D}
\end{equation}

\paragraph{Expressing all steps through $\Delta_+$.}
The two branches share a large common component, so their ranges are similar and the mean lies between them: $\Delta_-\approx\Delta_m\approx\Delta_+$, i.e.\ $v_-\approx v_m\approx v_+=\Delta_+^2/12$. The difference has the step $\Delta_d=\rho\Delta_+$, i.e.\ $v_d=\rho^2\Delta_+^2/12$. Inserting both into Eq.~\ref{eq:app_D} gives
\begin{equation}
D_B\approx\big(s^2+(s-1)^2\big)\frac{\Delta_+^2}{12},\qquad
D_D\approx\big(1+\beta^2\rho^2\big)\frac{\Delta_+^2}{12}.
\label{eq:app_D_approx}
\end{equation}
The common factor $\Delta_+^2/12$ cancels in the ratio, which is Eq.~\ref{eq:Guncorr}:
\begin{equation}
\frac{D_B}{D_D}\approx G_{\mathrm{uncorr}}(\rho)=\frac{s^2+(s-1)^2}{1+\beta^2\rho^2}.
\label{eq:app_Guncorr}
\end{equation}
The figures show the RMS ratio $\sqrt{G_{\mathrm{uncorr}}}$.

\paragraph{Reading the prediction.}
For $\rho\to0$ the reduction approaches $s^2+(s-1)^2$, i.e.\ 25 in squared error ($5\times$ in RMS) at $s=4$. Mean--delta quantization is better whenever $G_{\mathrm{uncorr}}>1$, i.e.\ for
\begin{equation}
\rho<\frac{\sqrt{2s(s-1)}}{s-\tfrac12},
\label{eq:app_breakeven}
\end{equation}
which is $\rho<1.40$ at $s=4$. All measured values in Fig.~\ref{fig:guided-error-gain} lie well below this (the largest is $\rho\approx0.9$). The benefit is a consequence of guidance: without guidance ($s=1$), the prediction becomes $1/(1+\rho^2/4)\le1$ and mean--delta quantization offers no advantage.

Section~\ref{sec:app_gaussian} extends this analysis to correlated rounding errors and derives the Gaussian correction shown in Figure~\ref{fig:guided-error-gain}.

\subsection{Measured reduction}
\label{sec:app_measurement}

\paragraph{Shared projection and reconstruction.}
Eq.~\ref{eq:gemp} measures the guided error after the projection $\widehat{W}$. For row errors, let $S_P=\mathbb{E}[e_P^\top e_P]$ with $P\in\{B,D\}$. The projected error energy is $\mathbb{E}\|e_P\widehat{W}\|_2^2=\mathrm{tr}(\widehat{W}^\top S_P\widehat{W})$. If the error components are isotropic and the paired cross-covariances vanish, $S_P=D_P I$ and the factor $\|\widehat{W}\|_F^2$ cancels in the ratio: the prediction holds after the projection unchanged. Reconstructing the branch outputs with Eq.~\ref{eq:decode} introduces no further error, because the reconstruction is linear and its guided combination equals $e_D\widehat{W}$ exactly.

\paragraph{Grouped scales.}
With per-token or per-group scales, each group $g$ has its own steps $\Delta_{+,g}$ and $\Delta_{d,g}$. Error energies add over groups, so the prediction applies with the energy-weighted ratio
\begin{equation}
\rho=\Big(\textstyle\sum_g n_g\Delta_{d,g}^2\big/\sum_g n_g\Delta_{+,g}^2\Big)^{1/2},
\label{eq:app_rho}
\end{equation}
where $n_g$ is the number of elements in group $g$. Each group contributes its own ratio of infinity norms. This pooled $\rho$ is the horizontal axis of Figs.~\ref{fig:guided-error-gain} and~\ref{e2e-errors}, computed for each prompt, layer and denoising step.

\paragraph{Protocol for Fig.~\ref{fig:guided-error-gain}.}
\begingroup\sloppy
We generate $1024\times1024$ images with FLUX.2 [klein] 4B Base~\cite{flux2026klein} for 128 prompts, one seed each, with 20 denoising steps and CFG scale $s=4$. The BF16 reference trajectory is never modified: we record the inputs of 25 projections (the query projection of each of the 5 double-stream blocks and the fused QKV/MLP projection of each of the 20 single-stream blocks) at 10 of the 20 steps, using 32 uniformly spaced image tokens and excluding text tokens. Both branches are rotated with an online 32-channel block Hadamard transform before quantization, and the projection uses $\widehat{W}H^\top$, which leaves its unquantized output unchanged. Quantization (per-token symmetric absmax RTN), mean--delta encoding and reconstruction, and the projection are computed in FP32. The points at $\rho=0$ come from the first projection, where both branches receive identical inputs, so $d_x=0$ and the reduction is exactly one.\par\endgroup

These measurements quantify the local error reduction. Section~\ref{sec:app_propagation} examines how this reduction propagates through the remaining denoiser.

\subsection{Correlated rounding errors}
\label{sec:app_gaussian}

To keep the expressions short we write $A=s^2+(s-1)^2$ and $C=2s(s-1)$, so that $G_{\mathrm{uncorr}}=A/(1+\beta^2\rho^2)$.

Equation~\ref{eq:Guncorr} assumes that the rounding errors $e_+$ and $e_-$ are uncorrelated. This fails when $d_x$ is small compared to the quantization step: $x^+$ and $x^-$ then often fall into the same rounding cell and receive nearly the same error. Let $\lambda$ be the correlation coefficient of $e_+$ and $e_-$, with $v_+=v_-=v$. Then
\begin{equation}
\mathbb{E}\big[(s\,e_++(1-s)\,e_-)^2\big]=(A-C\lambda)\,v ,
\label{eq:app_corr_baseline}
\end{equation}
so correlated errors partly cancel in the guided combination and reduce the error of independent quantization. Keeping the mean--delta error unchanged, the reduction becomes
\begin{equation}
G_{\mathrm{Gaussian}}(\rho)=\frac{A-C\,\lambda(\rho)}{1+\beta^2\rho^2}.
\label{eq:app_Ggauss}
\end{equation}

\paragraph{Computing $\lambda$.}
If both branches are rounded on the same grid of step $\Delta$, the error difference of two values that differ by $d$ depends only on $d \bmod \Delta$. Averaging over Gaussian entries of $d_x$ with standard deviation $\sigma$ gives
\begin{equation}
\lambda(\rho)=\frac{6}{\pi^2}\sum_{k\ge1}\frac{1}{k^2}\,e^{-2\pi^2k^2a^2},
\qquad a=\frac{\sigma}{\Delta}.
\label{eq:app_lambda}
\end{equation}
For $a=0$ (identical branches), $\lambda=1$; for large $a$, $\lambda\to0$ and $G_{\mathrm{Gaussian}}\to G_{\mathrm{uncorr}}$.

To express $a$ through $\rho$, note that the absmax quantizer sets the largest of the $g$ entries of a group to $q_{\max}\Delta_d$. For Gaussian entries this maximum is on average $c_g\sigma$, with $c_g$ the expected maximum of $g$ standard normal magnitudes ($c_g\approx3.73$ for the per-token groups of $g=3072$ in Fig.~\ref{fig:guided-error-gain}). Hence
\begin{equation}
a=\frac{q_{\max}\,\rho}{c_g}.
\label{eq:app_a}
\end{equation}

\paragraph{When the correction matters.}
$\lambda$ is negligible once $a\gtrsim\tfrac12$, i.e.\ $q_{\max}\rho\gtrsim1.9$. This is $\rho\gtrsim0.27$ at 4 bits, $\rho\gtrsim0.06$ at 6 bits and $\rho\gtrsim0.015$ at 8 bits, which is why the measurements in Fig.~\ref{fig:guided-error-gain} deviate from $G_{\mathrm{uncorr}}$ mainly at small $\rho$ and 4 bits.

 For heavy-tailed activations (e.g.\ without rotation) the two branches can have different group maxima and therefore different grids; the formula then overestimates the reduction. The correction is used only for analysis, not during D4D inference.

\subsection{Propagation through the denoiser}
\label{sec:app_propagation}
\label{app:propagation}

In the full model, CFG is applied at the denoiser output, not after each projection. An error introduced at one projection passes through the remaining layers separately in each branch before the two predictions are combined. The local reduction survives only if the remaining network treats both branches similarly.

\paragraph{Model.}
Let $J^\pm$ be the Jacobians of the remaining denoiser for the two branches, and $\epsilon$ an error at the quantized layer. Assuming $\|J^+\epsilon\|\approx\|J^-\epsilon\|$, we define the branch correlation
\begin{equation}
r=\frac{\mathbb{E}\langle J^+\epsilon,\,J^-\epsilon\rangle}{\mathbb{E}\|J^+\epsilon\|^2}.
\label{eq:app_r}
\end{equation}
Independent quantization gives uncorrelated branch errors, so its output error does not depend on $r$. The mean error $e_m$, however, enters both branches and only cancels to $e_m$ in the guided output if $r=1$. Repeating the derivation of Sec.~\ref{sec:app_derivation} with this correlation gives
\begin{equation}
G_{\mathrm{e2e}}(\rho,r)=\frac{A}{(A-Cr)+(A+Cr)\,\rho^2/4}.
\label{eq:app_Ge2e}
\end{equation}
For $r=1$ this is $G_{\mathrm{uncorr}}$; for $r=0$ it becomes $1/(1+\rho^2/4)\le1$, i.e.\ a small loss. The reduction is sensitive to $r$ near one: at small $\rho$ and $s=4$, the RMS reduction drops from $5\times$ at $r=1$ to $2.7\times$ at $r=0.9$.

\paragraph{Measurement.}
We quantize the input of exactly one of the 25 projections (as in Fig.~\ref{fig:guided-error-gain}) and run the rest of the denoiser unquantized. These passes use an FP32 copy of the denoiser fed with the inputs of the BF16 reference trajectory: in BF16, rounding in the remaining layers is as large as the effect being measured. We compare the guided predictions of the quantized and unquantized passes for 54 prompts at denoising steps 2, 9 and 16.

\begin{figure*}[t]
\centering
\includegraphics[width=\textwidth]{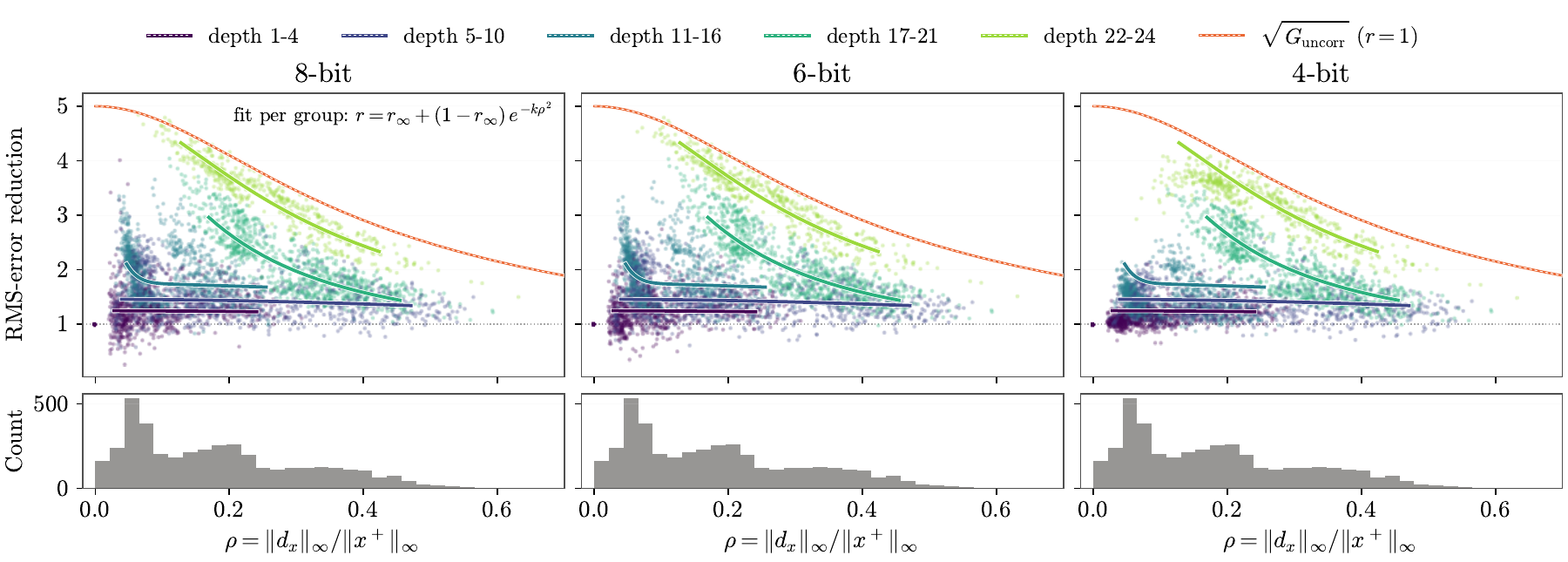}

\caption{\textbf{Depth dependence of propagated quantization error.} RMS-error reduction at 8, 6, and 4 bits on FLUX.2 [klein] 4B Base~\cite{flux2026klein}, grouped by layer depth. The dashed curve is the local prediction, colored curves are empirical fits, and histograms show the distribution of $\rho$. Later layers show larger gains and closer agreement with the prediction.}
\label{e2e-errors}
\end{figure*}

\paragraph{Results.}
For a single quantized projection, the median RMS reduction is $1.6\times$ at 8 and 6 bits and $1.4\times$ at 4 bits, compared with $4.3\times$ and $3.5\times$ locally. The reduction grows with depth, see Fig.~\ref{e2e-errors}. For depths 1--16, the fitted $r$ is roughly constant in $\rho$, rising from about 0.4 to 0.7 with depth. For depths 17--24, $r$ is close to one and decreases slowly with $\rho$, so the reduction approaches $G_{\mathrm{uncorr}}$ for the last layers. Fits at 8 bits predict 6 bits within 4\% without refitting; at 4 bits they overestimate by 4--28\%, because Eq.~\ref{eq:app_Ge2e} omits the rounding correlation of Sec.~\ref{sec:app_gaussian}.

With all 25 projections quantized jointly, the reduction is $2.7\times$ at every bit width. The joint error energy equals the sum of the single-projection energies (within 0.2\% at 8 and 6 bits, 3\% at 4 bits), so errors from different layers add independently. Individual prompts can still lose: in the double-stream blocks at steps 9 and 16, up to a third of prompts show a reduction below one for a single projection. This analysis measures denoiser error; sample quality is evaluated in Sec.~\ref{sec:experiments}.

\subsection{Choice of coordinates}
\label{sec:app_anchor}

\paragraph{Rescaled coordinates.}
An absmax quantizer is scale-equivariant: its step follows the row maximum, so $Q(c\,u)=c\,Q(u)$ for any $c>0$. Rescaling the mean and difference coordinates therefore leaves the reconstructed branches unchanged. In particular, the normalized Hadamard transform, whose coordinates are $\sqrt2\,m_x$ and $d_x/\sqrt2$, produces exactly the same errors as D4D under this model. Differences between the two arise only from quantizer details that break scale-equivariance, such as power-of-two shared scales (see Appendix~\ref{app:coordinate-scaling}).

\paragraph{Relation to the branch Hadamard.}
D4D and normalized Hadamard share the same directions and differ only in the scale of each coordinate. In exact arithmetic and with ideal per-group absmax scales they give the same errors; anchor--delta belongs to the same family, and all three improve over the original coordinates (Tab.~\ref{tab:coordinates}). The family is defined by one property: the guidance-amplified difference receives its own quantization scale. Among its members, D4D is the one with power-of-two coefficients $\{\tfrac12,1\}$. Its encoding and decoding are exact in binary floating point, whereas the $1/\sqrt2$ factors of the Hadamard are not. Since the ideal model predicts equal errors for the two, the differences observed between them in Tab.~\ref{tab:coordinates} reflect quantizer and implementation details outside the model. With power-of-two shared scales (MX formats) the normalization matters by construction, and D4D outperforms normalized Hadamard by 7.1\% in LPIPS (App.~\ref{app:coordinate-scaling}). The orthogonal rotations that GCBT fits from activation statistics also land close to the D4D directions (App.~\ref{app:concurrent}, Tab.~\ref{tab:gcbt-angles}).

\paragraph{Relation to sum--difference coding.}
The pair $(m_x,d_x)$ is the sum--difference transform of classical joint coding of two correlated channels, e.g.\ mid/side stereo coding (Johnston and Ferreira, ICASSP 1992). What is specific to CFG is that the output weights the two coordinates asymmetrically: the error of $d_x$ is amplified by $\beta=s-\tfrac12$ and that of $m_x$ is not (Eq.~\ref{eq:app_errors}). This asymmetry, rather than the correlation alone, determines when the transform pays off (Eq.~\ref{eq:app_breakeven}) and why the benefit grows with guidance (App.~\ref{app:guidance-strength}).

\paragraph{Anchor coordinate.}
More generally, any coordinate on the line through $x^-$ and $x^+$ can take the role of the mean. Let the anchor be $a=m_x+t\,d_x$, so that $t=0$ is D4D and $t=\pm\tfrac12$ are the original branches $x^\pm$. The guided output is then
\begin{equation}
g=a+(\beta-t)\,d_x .
\end{equation}
The anchor error enters $g$ with weight one, while the error of $d_x$ is multiplied by $\beta-t$. Since CFG extrapolates beyond $x^+$, an anchor closer to $x^+$ shortens this factor, and the local prediction becomes $A/(1+(\beta-t)^2\rho^2)$. Locally, anchoring on $x^+$ ($t=\tfrac12$) is therefore slightly better than D4D: at $s=4$ the difference error is amplified by 3 instead of 3.5.

\paragraph{Propagation favors the mean.}
The branches, however, pass through the remaining denoiser separately. The anchor distributes the difference error unevenly: $x^+=a+(\tfrac12-t)\,d_x$ and $x^-=a-(\tfrac12+t)\,d_x$. With the branch correlation $r$ of Sec.~\ref{sec:app_propagation}, the difference error contributes $D(t,r)\,\rho^2$ to the denominator of Eq.~\ref{eq:app_Ge2e}, with
\begin{equation}
\begin{aligned}
D(t,r)={}&s^2(\tfrac12-t)^2+(s-1)^2(\tfrac12+t)^2\\
&+C\,r\,(\tfrac14-t^2).
\end{aligned}
\end{equation}
For $r=1$ this reduces to $(\beta-t)^2$ and favors $x^+$. For smaller $r$ the optimum moves towards the mean:
\begin{equation}
t^\ast(r)=\frac{2s-1}{2(A-Cr)},
\end{equation}
which is $t^\ast=0.14$ for $r=0$ and $s=4$. Comparing the two candidates directly, D4D beats the $x^+$ anchor whenever $r<(4(s-1)^2-A)/C$, i.e.\ $r<0.46$ at $s=4$. The measured $r$ lies between about 0.4 and 0.7 for the first 16 projections and approaches one only for the last few (Sec.~\ref{sec:app_propagation}). For the measured range of $r$, the locally optimal anchor $t^\ast(r)$ lies on the $x^+$ side of the mean ($t^\ast>0$), reaches $x^+$ at $r=0.75$ and moves beyond it as $r$ approaches one; it depends on $r$, which varies across layers, steps and prompts and is not known before inference; the model also omits the correlated rounding of Sec.~\ref{sec:app_gaussian}. A fixed anchor therefore has to trade off layers with small and large $r$. We use the symmetric choice $t=0$, which needs no estimate of $r$, favors neither branch, and keeps power-of-two coefficients. In Tab.~\ref{tab:coordinates}, D4D and anchor--delta ($t=\tfrac12$) perform similarly, and both are far better than mean--anchor. Note that mean--anchor in Tab.~\ref{tab:coordinates} stores $(m_x,x^-)$, i.e.\ it replaces the difference by a branch, and is not a member of the anchor family above; anchor--delta stores $(x^+,d_x)$ and is the member $t=\tfrac12$. An $r$-dependent anchor per layer is a possible refinement that would require calibration statistics, which D4D avoids.

\subsection{Branch statistics of the evaluated models}
\label{sec:app_branch_stats}
Figs.~\ref{fig:guided-error-gain} and~\ref{e2e-errors} use FLUX.2 [klein]. To check the premise of the analysis on the evaluated models, we compute for every D4D-eligible projection the energy ratio $\rho_2=(\mathbb{E}\|d_x\|_2^2/\mathbb{E}\|x^+\|_2^2)^{1/2}$ from second moments of BF16 branch activations collected on the four calibration examples of each model (App.~\ref{app:protocols}). $\rho_2$ is an energy-based analogue of the range ratio $\rho$ of Eq.~\ref{eq:main-range-ratio}; it does not depend on the base quantizer. Tab.~\ref{tab:branch-stats} reports it for the five models with archived paired BF16 activations, together with the break-even ratio of Eq.~\ref{eq:app_breakeven} at each model's guidance scale.

On every eligible projection of every model, $\rho_2$ is below the break-even ratio; the largest value is 1.01, against a break-even of about 1.4. The premise of the analysis, a branch difference that is small relative to the branches, therefore holds on the evaluated models as on FLUX.2 [klein]. These statistics describe the local regime only; the size of the end-to-end gain also depends on propagation (App.~\ref{sec:app_propagation}) and on the base quantizer, and is measured in Sec.~\ref{sec:experiments}.

\begin{table}[!htbp]
\centering\small
\setlength{\tabcolsep}{3pt}
\caption{\textbf{Branch statistics of the evaluated models.} $\rho_2$ over D4D-eligible projections: median, 10--90\% range and maximum. $\rho^\ast$: break-even ratio of Eq.~\ref{eq:app_breakeven} at the model's guidance scale $s$.}
\label{tab:branch-stats}
\resizebox{\columnwidth}{!}{%
\begin{tabular}{lcccccc}
\toprule
Model & $s$ & Proj. & $\rho_2$ median & 10--90\% & max & $\rho^\ast$\\
\midrule
Chroma1-HD & 3 & 304 & 0.56 & 0.20--0.70 & 0.78 & 1.39\\
Wan2.2-TI2V-5B & 5 & 90 & 0.30 & 0.15--0.54 & 0.59 & 1.41\\
Wan2.2-I2V-A14B & 3.5 & 240 & 0.42 & 0.14--0.92 & 0.98 & 1.39\\
Pixal3D & 7 & 384 & 0.32 & 0.10--0.71 & 1.01 & 1.41\\
3DTopia-XL & 6 & 112 & 0.50 & 0.19--0.76 & 0.96 & 1.41\\
\bottomrule
\end{tabular}}
\end{table}

\section{Experimental and Implementation Details}
\label{app:protocols}
All comparisons use the same evaluation inputs, generation seeds, and sampling settings with and without \methodname{D4D}. Calibration data are excluded from evaluation. Tables~\ref{tab:checkpoints} and~\ref{tab:sampling-protocol} list the models and sampling settings. The D4D transform itself requires no calibration; the base quantizers retain their calibration procedures.

\subsection{Image generation}
\label{app:image-protocol}
\label{app:chroma-protocol}
Z-Image and Chroma1-HD use the same 1,000 MJHQ-30K prompts, with 100 per category and one seed per prompt. Z-Image uses CFG scale $s=5$ in our notation, while Chroma uses guidance scale 3 and an empty negative prompt.

SVDQuant uses NVFP4 with groups of 16 and FP8 (E4M3) scales, rank 128, and \methodname{GPTQ}~\cite{frantar2023gptq} residual quantization. Our Q-DiT adaptation uses asymmetric INT4 groups of 128, dynamic activation quantization, and GPTQ damping 0.01. MACaQ uses MXFP4~\cite{ocp2023mx} groups of 32, a width-32 block-Hadamard transform, balancing exponent 0.75, ridge factor 1, exponent offsets $\{1,0,-1\}$, and up to 512 refinement iterations.

For Z-Image, Q-DiT and MACaQ each calibrate on 20 disjoint prompts and 20,480 BF16 activation rows per target. Q-DiT samples eight timesteps without progressive recapture of quantized blocks. Both quantize 238 attention and feed-forward projections, retaining embeddings, modulation, normalization, and the output head in BF16. For Chroma, Q-DiT and MACaQ use four calibration prompts at timesteps 0, 6, 11, 17, 22, 28, 33, and 39, with 128 rows per branch and timestep. All three Chroma methods quantize 418 projections, of which 304 are eligible for D4D.

Z-Image baseline/D4D pairs share quantized weights. Text tokens retain branch coordinates.

\paragraph{Coordinates during calibration.}
In all settings, the base quantizers are calibrated once on activations in the original branch coordinates (smoothing factors, GPTQ statistics, low-rank decompositions, MACaQ balancing), and the resulting quantized weights are shared by the runs with and without D4D. D4D is then applied at inference without recalibrating the base quantizer. This isolates the effect of the coordinate change: baseline and D4D runs differ only in the coordinates passed to the same quantizer. In SVDQuant, D4D acts on the 4-bit residual path; the BF16 low-rank path receives the original branch activations. All W4A4 settings quantize linear projections; attention score and value products remain in BF16 for every method.

\subsection{Video generation}
Both Wan models use the same 128 category-stratified VBench-I2V image--prompt pairs, selected with seed 20261101 after excluding 64 calibration and 16 development conditions. Each pair has one generation seed shared across methods. The evaluation manifest specifies the conditioning images, guidance settings, and negative prompts.

Calibration uses four separate conditions and eight timesteps. SVDQuant uses groups of 64, rank 128, and GPTQ; its low-rank path remains BF16. Q-DiT uses asymmetric INT4 groups of 128 and GPTQ damping 0.01. MACaQ uses MXFP4 groups of 32 and shares quantized weights across the pair. D4D acts on feed-forward and self-attention output projections.

PSNR averages the 60 frames after the conditioning frame. AlexNet LPIPS averages frames 4, 12, 20, 28, 36, 44, 52, and 60 against matched lossless BF16 references. We then average scores over videos. VBench-I2V metrics use their official dimension-specific subsets.

\subsection{Image-to-3D generation}
\label{app:3d-protocol}
Pixal3D and 3DTopia-XL use normalized renders of the same 150 GSO objects, with one shared seed per object ($20262102+i$ for manifest index $i$). Four separate objects provide calibration data. Pixal3D-D uses guidance 7 at each stage; 3DTopia-XL uses guidance 6.

Our SVDQuant uses NVFP4 with groups of 16 and FP8 (E4M3) scales, activation/weight smoothing, GPTQ, and rank $\min(128,d_{\rm in},d_{\rm out})$
MACaQ uses groups of 32, 512 refinement iterations, ridge factor 1, and balancing exponent 0.75. Cross-attention retains branch coordinates; encoders, decoders, conditioning modulation, and excluded projections retain their original precision.

We render world-space vertex normals from 16 fixed Fibonacci-sphere cameras at $512\times512$ (radius 2, orthographic scale 0.65). PSNR and AlexNet LPIPS are averaged over views, then objects. Chamfer distance is the mean of the two directed, unsquared nearest-neighbor distances between 100,000 area-uniform surface samples. Both surfaces use the BF16 maximum extent for normalization, with no independent alignment. References are generated BF16 geometry, not ground-truth GSO surfaces. Results in Table~\ref{tab:3d-full}.

\begin{table*}[!t]
\centering\small
\setlength{\tabcolsep}{4pt}
\caption{\textbf{Evaluated models.} Checkpoint sources, tasks, and declared model licenses.}
\label{tab:checkpoints}
\begin{tabular}{p{.22\textwidth}p{.49\textwidth}p{.21\textwidth}}
\toprule
\textbf{Model} & \textbf{Checkpoint or source} & \textbf{Task / license}\\
\midrule
Z-Image & \path{Tongyi-MAI/Z-Image} & T2I / Apache-2.0\\
Chroma1-HD & \path{lodestones/Chroma1-HD} & T2I / Apache-2.0\\
Wan 5B & \path{Wan-AI/Wan2.2-TI2V-5B-Diffusers} & I2V / Apache-2.0\\
Wan A14B & \path{Wan-AI/Wan2.2-I2V-A14B-Diffusers} & I2V / Apache-2.0\\
Pixal3D & \path{TencentARC/Pixal3D-D} & I2-3D / MIT\\
3DTopia-XL & \path{3DTopia/3DTopia-XL} & I2-3D / Apache-2.0\\
\bottomrule
\specialrule{\heavyrulewidth}{1.5pt}{0pt}
\end{tabular}

\end{table*}

\begin{table*}[!t]
\centering\small
\setlength{\tabcolsep}{4pt}
\caption{\textbf{Sampling protocol.} Each condition uses one shared seed across BF16 and quantized variants.
50/30/15 denotes the steps in the three Pixal3D stages.}
\label{tab:sampling-protocol}
\begin{tabular}{llcc}
\toprule
\textbf{Model / benchmark} & \textbf{Conditions} & \textbf{Steps} & \textbf{Output setting}\\
\midrule
Z-Image / MJHQ & 1,000 & 28 & $1024\times1024$\\
Chroma1-HD / MJHQ & 1,000 & 40 & $1024\times1024$\\
Wan TI2V-5B / VBench-I2V & 128 & 40 & $480\times832$, 61 frames\\
Wan I2V-A14B / VBench-I2V & 128 & 40 & $480\times832$, 61 frames\\
Pixal3D / GSO & 150 objects & 50/30/15 & Geometry\\
3DTopia-XL / GSO & 150 objects & 25 & Geometry\\
\bottomrule
\specialrule{\heavyrulewidth}{1.5pt}{0pt}
\end{tabular}

\end{table*}

\section{Speech Generation}
\label{app:audio}
We evaluate CosyVoice2~\cite{du2024cosyvoice2} on 1,000 LibriTTS~\cite{zen2019libritts} test-clean utterances from 38 speakers, using a separate conditioning recording per speaker. All methods share language-model tokens, conditioning, and synthesis seeds, with ten flow steps and CFG scale $s=1.7$. We quantize 352 denoiser linear projections and apply D4D to the 336 paired token-wise projections. RTN uses INT4 groups of 64; MACaQ~\cite{muller2026macaq} uses MXFP4 groups of 32, calibrated on 32 dev-clean utterances. Each quantizer uses the same quantized weights with and without D4D.

We report CAMPPlus~\cite{wang2023campp} speaker similarity to the conditioning audio (SIM) and spectral distance to matched BF16 outputs (SD). SD averages magnitude-STFT spectral convergence over FFT sizes 256, 512, and 1024, with quarter-window hops.

\begin{table}[!htbp]
\centering\small
\setlength{\tabcolsep}{4pt}
\caption{\textbf{CosyVoice2 on 1,000 LibriTTS test-clean utterances.} Bold marks the better value within each quantized pair.}
\label{tab:audio}
\begin{tabular}{>{\raggedright\arraybackslash}p{\dimexpr0.4\linewidth-2\tabcolsep\relax}>{\raggedleft\arraybackslash}p{\dimexpr0.3\linewidth-2\tabcolsep\relax}>{\raggedleft\arraybackslash}p{\dimexpr0.3\linewidth-2\tabcolsep\relax}}
\toprule
\textbf{Method} & \textbf{SIM $\uparrow$} & \textbf{SD $\downarrow$}\\
\midrule
BF16 & 0.8670 & --\\
\midrule
RTN W4A4 & 0.8267 & 0.6286\\
\dFourDours{\textbf{+ D4D}} & \dFourDours{\textbf{0.8541}} & \dFourDours{\textbf{0.5538}}\\
\midrule
MACaQ W4A4 & 0.8650 & 0.3302\\
\dFourDours{\textbf{+ D4D}} & \dFourDours{\textbf{0.8659}} & \dFourDours{\textbf{0.3058}}\\
\bottomrule
\specialrule{\heavyrulewidth}{1.5pt}{0pt}
\end{tabular}
\end{table}

D4D reduces spectral distance to BF16 by 11.9\% with RTN and 7.4\% with MACaQ. Speaker similarity improves from 0.8267 to 0.8541 with RTN and changes only slightly with MACaQ (0.8650 to 0.8659). The benefit is clearest in fidelity to BF16 and, for RTN, speaker similarity.

\section{Extended Quantitative Results}
\label{app:extended-results}
\label{app:paired-quality}
\label{app:3d-results}
Tables~\ref{tab:paired-quality}--\ref{tab:3d-ci} provide additional metrics and paired differences (D4D minus baseline) with 95\% confidence intervals.

\paragraph{Typical versus largest gains.}
Tab.~\ref{tab:summary-gains} collects the relative LPIPS reduction of every model--quantizer pair in Tabs.~\ref{tab:images}--\ref{tab:3d}. D4D lowers mean LPIPS in all 16 settings. The median reduction is 10.4\%; the abstract reports the largest value per modality. The smallest reductions occur on Chroma1-HD (1.2--2.7\%), which uses the lowest guidance scale among the image models, consistent with the dependence on $s$ in Eq.~\ref{eq:Guncorr} and App.~\ref{app:guidance-strength}. The largest occur on 3DTopia-XL, where both W4A4 baselines degrade most, and on Wan2.2-TI2V-5B.

\begin{table}[!htbp]
\centering\small
\setlength{\tabcolsep}{4pt}
\caption{\textbf{Relative LPIPS reduction from D4D in every setting} (\%, computed from Tabs.~\ref{tab:images}--\ref{tab:3d}; LPIPS-N for 3D). Positive values favor D4D. Q-DiT was not evaluated on 3D.}
\label{tab:summary-gains}
\begin{tabular}{lrrr}
\toprule
Model & SVDQuant & Q-DiT & MACaQ\\
\midrule
Z-Image & 6.2 & 11.3 & 10.5\\
Chroma1-HD & 2.7 & 1.6 & 1.2\\
Wan2.2-TI2V-5B & 29.6 & 26.4 & 19.0\\
Wan2.2-I2V-A14B & 8.3 & 10.4 & 2.2\\
Pixal3D & 19.0 & -- & 10.4\\
3DTopia-XL & 52.8 & -- & 36.8\\
\midrule
\multicolumn{4}{l}{Median over 16 settings: 10.4; range 1.2--52.8}\\
\bottomrule
\end{tabular}
\end{table}

\paragraph{Per-prompt wins.}
Tab.~\ref{tab:winrates} reports, for Z-Image on all 1,000 MJHQ prompts, how often D4D lowers LPIPS to BF16 for an individual prompt. D4D wins on 78--94\% of prompts with every base quantizer, so the mean gains of Tab.~\ref{tab:images} are systematic across prompts and not driven by a few outliers.

\begin{table}[!htbp]
\centering\small
\setlength{\tabcolsep}{4pt}
\caption{\textbf{Per-prompt wins on Z-Image} (1,000 MJHQ prompts). Fraction of prompts where D4D has lower LPIPS to BF16 than the same quantizer without D4D.}
\label{tab:winrates}
\begin{tabular}{lc}
\toprule
Base & LPIPS win rate\\
\midrule
SVDQuant & 78.3\%\\
Q-DiT & 93.9\%\\
MACaQ & 86.3\%\\
\bottomrule
\end{tabular}
\end{table}

\begin{table}[!htbp]
\centering\small
\setlength{\tabcolsep}{4pt}
\caption{\textbf{Z-Image on MJHQ-1K.} Mean paired differences with 95\% bootstrap confidence intervals.}
\label{tab:paired-quality}
\begin{tabular}{lrr}
\toprule
\textbf{Metric} & \textbf{Mean difference} & \textbf{95\% interval}\\
\midrule
\multicolumn{3}{l}{\textbf{\methodname{SVDQuant}}}\\
LPIPS $\downarrow$ & $-0.03088$ & $[-0.03422,-0.02749]$\\
PSNR $\uparrow$ (dB) & $+0.2246$ & $[+0.1667,+0.2828]$\\
ImageReward $\uparrow$ & $+0.02680$ & $[+0.00611,+0.04736]$\\
\midrule
\multicolumn{3}{l}{\textbf{\methodname{Q-DiT}}}\\
LPIPS $\downarrow$ & $-0.07146$ & $[-0.07511,-0.06784]$\\
PSNR $\uparrow$ (dB) & $+0.46405$ & $[+0.41863,+0.50903]$\\
ImageReward $\uparrow$ & $+0.23939$ & $[+0.21011,+0.26900]$\\
\midrule
\multicolumn{3}{l}{\textbf{\methodname{MACaQ}}}\\
LPIPS $\downarrow$ & $-0.05212$ & $[-0.05581,-0.04846]$\\
PSNR $\uparrow$ (dB) & $+0.40106$ & $[+0.34214,+0.46076]$\\
ImageReward $\uparrow$ & $+0.02537$ & $[+0.00266,+0.04775]$\\
\bottomrule
\specialrule{\heavyrulewidth}{1.5pt}{0pt}
\end{tabular}
\end{table}

\begin{table}[!htbp]
\centering\small
\setlength{\tabcolsep}{4pt}
\caption{\textbf{Chroma1-HD on MJHQ-1K.} Mean paired differences with 95\% confidence intervals.}
\label{tab:chroma-paired-quality}
\begin{tabular}{lrr}
\toprule
\textbf{Metric} & \textbf{Mean difference} & \textbf{95\% interval}\\
\midrule
\multicolumn{3}{l}{\textbf{\methodname{SVDQuant}}}\\
LPIPS $\downarrow$ & $-0.0072$ & $[-0.0091,-0.0054]$\\
PSNR $\uparrow$ (dB) & $+0.183$ & $[+0.102,+0.261]$\\
ImageReward $\uparrow$ & $+0.0099$ & $[+0.0018,+0.0194]$\\
\midrule
\multicolumn{3}{l}{\textbf{\methodname{Q-DiT}}}\\
LPIPS $\downarrow$ & $-0.0056$ & $[-0.0078,-0.0039]$\\
PSNR $\uparrow$ (dB) & $+0.255$ & $[+0.194,+0.321]$\\
ImageReward $\uparrow$ & $+0.0181$ & $[+0.0067,+0.0289]$\\
\midrule
\multicolumn{3}{l}{\textbf{\methodname{MACaQ}}}\\
LPIPS $\downarrow$ & $-0.0036$ & $[-0.0052,-0.0017]$\\
PSNR $\uparrow$ (dB) & $+0.115$ & $[+0.038,+0.187]$\\
ImageReward $\uparrow$ & $+0.0122$ & $[+0.0031,+0.0228]$\\
\bottomrule
\specialrule{\heavyrulewidth}{1.5pt}{0pt}
\end{tabular}
\end{table}

\begin{table*}[!htbp]
\centering\footnotesize
\setlength{\tabcolsep}{4pt}
\caption{\textbf{Additional VBench-I2V metrics.} Following the benchmark annotations, I2V-S (subject preservation), MS (motion smoothness), and DD (dynamic degree) use the 78 subject examples, while I2V-B (background preservation) uses the 50 background examples. PSNR and LPIPS use all 128 videos.}
\label{tab:video-full}
\begin{tabular}{clccccccc}
\toprule
\textbf{Model} & \textbf{Base method} & \textbf{W/A} & \textbf{I2V-S $\uparrow$} & \textbf{I2V-B $\uparrow$} & \textbf{MS $\uparrow$} & \textbf{DD $\uparrow$} & \textbf{PSNR $\uparrow$} & \textbf{LPIPS $\downarrow$}\\
\midrule
\dFourDmodel[-7.5pt]{7}{Wan2.2\\TI2V-5B} & \textit{BF16} & \textit{16/16} & \textit{0.9770} & \textit{0.9858} & \textit{0.9746} & \textit{0.7564} & \textit{--} & \textit{--}\\
\cmidrule(l){2-9}
 & \dFourDothers{\methodname{SVDQuant}} & \dFourDothers{4/4} & \dFourDothers{0.9638} & \dFourDothers{0.9804} & \dFourDothers{0.9739} & \dFourDothers{0.6795} & \dFourDothers{19.013} & \dFourDothers{0.2616}\\
 & \dFourDours{\textbf{+ \methodname{D4D}}} & \dFourDours{4/4} & \dFourDours{\textbf{0.9751}} & \dFourDours{\textbf{0.9853}} & \dFourDours{\textbf{0.9770}} & \dFourDours{\textbf{0.6923}} & \dFourDours{\textbf{20.715}} & \dFourDours{\textbf{0.1843}}\\
\cmidrule(l){2-9}
 & \dFourDothers{\methodname{Q-DiT}} & \dFourDothers{4/4} & \dFourDothers{0.9575} & \dFourDothers{0.9778} & \dFourDothers{0.9757} & \dFourDothers{0.6410} & \dFourDothers{18.635} & \dFourDothers{0.2972}\\
 & \dFourDours{\textbf{+ \methodname{D4D}}} & \dFourDours{4/4} & \dFourDours{\textbf{0.9723}} & \dFourDours{\textbf{0.9839}} & \dFourDours{\textbf{0.9782}} & \dFourDours{\textbf{0.6538}} & \dFourDours{\textbf{19.914}} & \dFourDours{\textbf{0.2187}}\\
\cmidrule(l){2-9}
 & \dFourDothers{\methodname{MACaQ}} & \dFourDothers{4/4} & \dFourDothers{0.9686} & \dFourDothers{0.9835} & \dFourDothers{0.9768} & \dFourDothers{0.6282} & \dFourDothers{18.663} & \dFourDothers{0.2574}\\
 & \dFourDours{\textbf{+ \methodname{D4D}}} & \dFourDours{4/4} & \dFourDours{\textbf{0.9791}} & \dFourDours{\textbf{0.9873}} & \dFourDours{\textbf{0.9781}} & \dFourDours{\textbf{0.6410}} & \dFourDours{\textbf{19.900}} & \dFourDours{\textbf{0.2085}}\\
\midrule
\dFourDmodel[-7.5pt]{7}{Wan2.2\\I2V-A14B} & \textit{BF16} & \textit{16/16} & \textit{0.9776} & \textit{0.9917} & \textit{0.9807} & \textit{0.6795} & \textit{--} & \textit{--}\\
\cmidrule(l){2-9}
 & \dFourDothers{\methodname{SVDQuant}} & \dFourDothers{4/4} & \dFourDothers{0.9585} & \dFourDothers{0.9827} & \dFourDothers{0.9813} & \dFourDothers{0.6410} & \dFourDothers{19.810} & \dFourDothers{0.2102}\\
 & \dFourDours{\textbf{+ \methodname{D4D}}} & \dFourDours{4/4} & \dFourDours{\textbf{0.9744}} & \dFourDours{\textbf{0.9896}} & \dFourDours{\textbf{0.9813}} & \dFourDours{0.6410} & \dFourDours{\textbf{20.257}} & \dFourDours{\textbf{0.1928}}\\
\cmidrule(l){2-9}
 & \dFourDothers{\methodname{Q-DiT}} & \dFourDothers{4/4} & \dFourDothers{0.9533} & \dFourDothers{0.9777} & \dFourDothers{0.9796} & \dFourDothers{0.5897} & \dFourDothers{19.329} & \dFourDothers{0.2311}\\
 & \dFourDours{\textbf{+ \methodname{D4D}}} & \dFourDours{4/4} & \dFourDours{\textbf{0.9689}} & \dFourDours{\textbf{0.9869}} & \dFourDours{\textbf{0.9803}} & \dFourDours{\textbf{0.6410}} & \dFourDours{\textbf{19.790}} & \dFourDours{\textbf{0.2071}}\\
\cmidrule(l){2-9}
 & \dFourDothers{\methodname{MACaQ}} & \dFourDothers{4/4} & \dFourDothers{0.9617} & \dFourDothers{0.9831} & \dFourDothers{0.9826} & \dFourDothers{0.6154} & \dFourDothers{19.548} & \dFourDothers{0.2164}\\
 & \dFourDours{\textbf{+ \methodname{D4D}}} & \dFourDours{4/4} & \dFourDours{\textbf{0.9737}} & \dFourDours{\textbf{0.9893}} & \dFourDours{\textbf{0.9831}} & \dFourDours{\textbf{0.6410}} & \dFourDours{\textbf{19.780}} & \dFourDours{\textbf{0.2116}}\\
\bottomrule
\specialrule{\heavyrulewidth}{1.5pt}{0pt}
\end{tabular}
\end{table*}

\begin{table*}[!t]
\centering\footnotesize
\setlength{\tabcolsep}{4pt}
\caption{\textbf{Paired video differences.} D4D minus each matched W4A4 baseline, with 95\% bootstrap confidence intervals.}
\label{tab:i2v}
\renewcommand{\arraystretch}{1.3}
\begin{tabular}{clrrrr}
\toprule
\textbf{Model} & \textbf{Base method} & $\Delta$\textbf{PSNR (dB)} & \textbf{95\% interval} & $\Delta$\textbf{LPIPS} & \textbf{95\% interval}\\
\midrule
\dFourDmodel{3}{Wan2.2\\TI2V-5B} & \methodname{SVDQuant} & $+1.7021$ & $[+1.3812, +2.0349]$ & $-0.0773$ & $[-0.0863, -0.0684]$\\
 & \methodname{Q-DiT} & $+1.2790$ & $[+0.9221, +1.6398]$ & $-0.0785$ & $[-0.0897, -0.0667]$\\
 & \methodname{MACaQ} & $+1.2365$ & $[+0.9495, +1.5255]$ & $-0.0488$ & $[-0.0569, -0.0403]$\\
\midrule
\dFourDmodel{3}{Wan2.2\\I2V-A14B} & \methodname{SVDQuant} & $+0.4469$ & $[+0.1395, +0.7532]$ & $-0.01749$ & $[-0.02932, -0.00588]$\\
 & \methodname{Q-DiT} & $+0.4610$ & $[+0.0846, +0.8184]$ & $-0.02403$ & $[-0.03725, -0.01033]$\\
 & \methodname{MACaQ} & $+0.2315$ & $[-0.0672, +0.5258]$ & $-0.00487$ & $[-0.01352, -0.00117]$\\
\bottomrule
\specialrule{\heavyrulewidth}{1.5pt}{0pt}
\end{tabular}
\end{table*}

\begin{table*}[!htbp]
\centering\footnotesize
\setlength{\tabcolsep}{4pt}
\caption{\textbf{Additional 3D metrics on GSO-150.} Fidelity to matched BF16 outputs. $^{*}$Our adaptation of \methodname{SVDQuant}.}
\label{tab:3d-full}
\begin{tabular}{cllccc}
\toprule
\textbf{Model} & \textbf{Base method} & \textbf{W/A} & \textbf{PSNR-N$\uparrow$} & \textbf{LPIPS-N$\downarrow$} & \textbf{CD$\downarrow$}\\
\midrule
\dFourDmodel[-5pt]{5}{Pixal3D} & \textit{BF16} & \textit{16/16} & -- & -- & --\\
\cmidrule(l){2-6}
 & \dFourDothers{\methodname{SVDQuant}$^{*}$} & \dFourDothers{4/4} & \dFourDothers{24.25} & \dFourDothers{0.0717} & \dFourDothers{0.005412}\\
 & \dFourDours{\textbf{+ \methodname{D4D}}} & \dFourDours{4/4} & \dFourDours{\textbf{24.79}} & \dFourDours{\textbf{0.0581}} & \dFourDours{\textbf{0.005266}}\\
\cmidrule(l){2-6}
 & \dFourDothers{\methodname{MACaQ}} & \dFourDothers{4/4} & \dFourDothers{16.82} & \dFourDothers{0.2827} & \dFourDothers{0.008781}\\
 & \dFourDours{\textbf{+ \methodname{D4D}}} & \dFourDours{4/4} & \dFourDours{\textbf{17.20}} & \dFourDours{\textbf{0.2534}} & \dFourDours{\textbf{0.008780}}\\
\midrule
\dFourDmodel[-5pt]{5}{3DTopia-XL} & \textit{BF16} & \textit{16/16} & -- & -- & --\\
\cmidrule(l){2-6}
 & \dFourDothers{\methodname{SVDQuant}$^{*}$} & \dFourDothers{4/4} & \dFourDothers{11.30} & \dFourDothers{0.5220} & \dFourDothers{0.052781}\\
 & \dFourDours{\textbf{+ \methodname{D4D}}} & \dFourDours{4/4} & \dFourDours{\textbf{16.26}} & \dFourDours{\textbf{0.2463}} & \dFourDours{\textbf{0.040248}}\\
\cmidrule(l){2-6}
 & \dFourDothers{\methodname{MACaQ}} & \dFourDothers{4/4} & \dFourDothers{7.80} & \dFourDothers{0.7506} & \dFourDothers{0.066646}\\
 & \dFourDours{\textbf{+ \methodname{D4D}}} & \dFourDours{4/4} & \dFourDours{\textbf{11.86}} & \dFourDours{\textbf{0.4744}} & \dFourDours{\textbf{0.050159}}\\
\bottomrule
\specialrule{\heavyrulewidth}{1.5pt}{0pt}
\end{tabular}
\end{table*}

\begin{table*}[!t]
\centering\footnotesize
\setlength{\tabcolsep}{3pt}
\caption{\textbf{Paired 3D differences.} Mean differences with 95\% bootstrap confidence intervals over objects.}
\label{tab:3d-ci}
\begin{tabular}{llrrr}
\toprule
\textbf{Model} & \textbf{Base method} & $\Delta$\textbf{PSNR-N} & $\Delta$\textbf{LPIPS-N} & $\Delta$\textbf{CD}\\
\midrule
Pixal3D & \methodname{SVDQuant}$^{*}$ & +0.542 [+0.284, +0.797] & -0.0136 [-0.0161, -0.0112] & -0.000146 [-0.000375, -0.000012]\\
Pixal3D & \methodname{MACaQ} & +0.385 [+0.154, +0.617] & -0.0292 [-0.0334, -0.0250] & -0.000001 [-0.000001, -0.000001]\\
3DTopia-XL & \methodname{SVDQuant}$^{*}$ & +4.966 [+4.613, +5.316] & -0.2756 [-0.2887, -0.2624] & -0.012533 [-0.016188, -0.008900]\\
3DTopia-XL & \methodname{MACaQ} & +4.058 [+3.854, +4.269] & -0.2763 [-0.2873, -0.2655] & -0.016486 [-0.018464, -0.014508]\\
\bottomrule
\specialrule{\heavyrulewidth}{+1.5pt}{0pt}
\end{tabular}
\end{table*}
Additional qualitative comparisons are provided in Appendix~\ref{app:qualitative}.

\section{Effect of Branch-Coordinate Scaling}
\label{app:coordinate-scaling}
To isolate the effect of normalization, we keep the mean--difference directions fixed and vary their relative scaling:
\begin{equation}
T(a)=
\begin{bmatrix}2^{-a/2}&0\\0&2^{a/2}\end{bmatrix}
\frac{1}{\sqrt{2}}
\begin{bmatrix}1&1\\1&-1\end{bmatrix}.
\end{equation}
We apply $T(a)$ to paired branch activations before quantization and its inverse to the projected outputs. At $a=0$, $T(a)$ is normalized Hadamard. At $a=1$, it produces the mean and difference coordinates used by D4D. Thus, D4D shares the Hadamard directions but uses a non-orthogonal normalization.

\begin{figure}[!htbp]
\centering
\includegraphics[width=\linewidth]{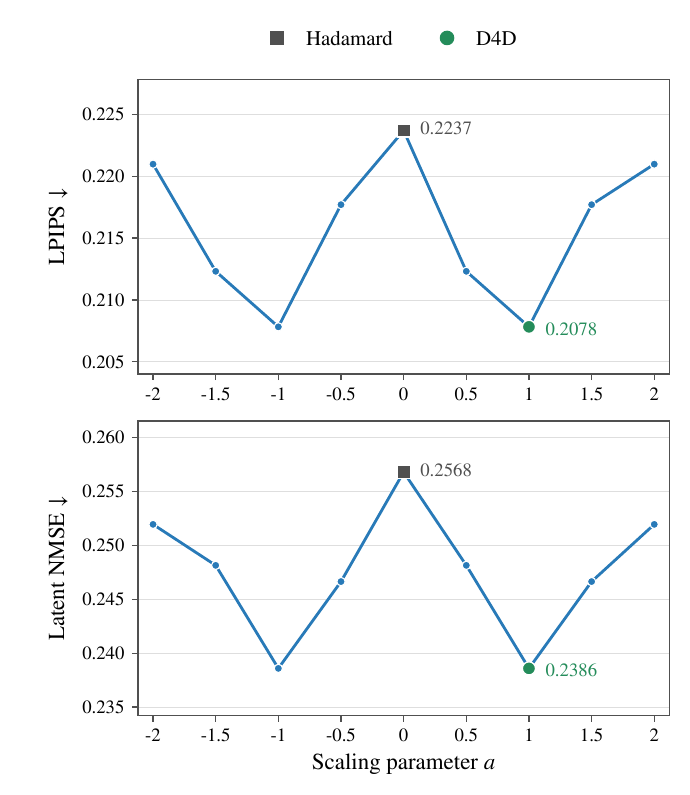}
\caption{\textbf{Effect of branch-coordinate scaling.} LPIPS and latent NMSE relative to BF16 on a subset of 16 image--prompt pairs from VBench-I2V with Wan2.2-TI2V-5B and MACaQ W4A4. We vary $a$ from $-2$ to $2$ in increments of $0.5$, keeping quantized weights, calibration parameters, evaluated inputs, and sampling settings fixed. Normalized Hadamard ($a=0$) and D4D ($a=1$) are highlighted. Lower is better.}
\label{fig:coordinate-scaling}
\end{figure}

Figure~\ref{fig:coordinate-scaling} evaluates this family with Wan2.2-TI2V-5B and MACaQ W4A4. Relative to normalized Hadamard, D4D reduces LPIPS from 0.2237 to 0.2078 and latent NMSE from 0.2568 to 0.2386, a 7.1\% reduction in both metrics. D4D attains the lowest errors among the evaluated scalings, tied with $a=-1$. These observed means favor mean--delta normalization in this MXFP4 pipeline.

MXFP4's power-of-two scales imply an expected period of 2 in $a$. Increasing $a$ by 2 halves one coordinate and doubles the other, changes that the quantization scales can absorb within their representable range. Finite-precision arithmetic can introduce deviations from exact periodicity.

\paragraph{When normalization matters.}
Write $T(a)=D(a)H$, where $H$ is normalized Hadamard. If each coordinate uses the same positively scale-equivariant quantization rule, $Q(cu)=cQ(u)$ for $c>0$, then
\begin{equation}
T(a)^{-1}Q(T(a)x)=H^{-1}Q(Hx).
\end{equation}
This identity assumes matched grouping and independent scales, and holds in exact arithmetic. It also holds after a shared linear projection. Thus, unequal scaling is not an intrinsic advantage for ideal absmax INT4. MXFP4's restricted scale grid and finite-precision arithmetic can break this equivalence.

\section{Sensitivity to CFG Guidance Strength}
\label{app:guidance-strength}
CFG amplifies the difference between conditional and unconditional predictions. We evaluate how changing this amplification affects D4D's benefit on Z-Image~\cite{team2025zimage} with SVDQuant~\cite{li2025svdquant} W4A4.

We use the same 100 MJHQ-30K~\cite{li2024playground} prompts, balanced across categories, with matched seeds at CFG scales $s\in\{3,5,7\}$. Quantized weights and calibration parameters remain fixed. Each setting has its own BF16 reference for LPIPS and PSNR. ImageReward evaluates image preference and prompt alignment. At $s=5$, we reuse the existing outputs for the same prompts.

\begin{figure}[!htbp]
\centering
\includegraphics[width=\linewidth]{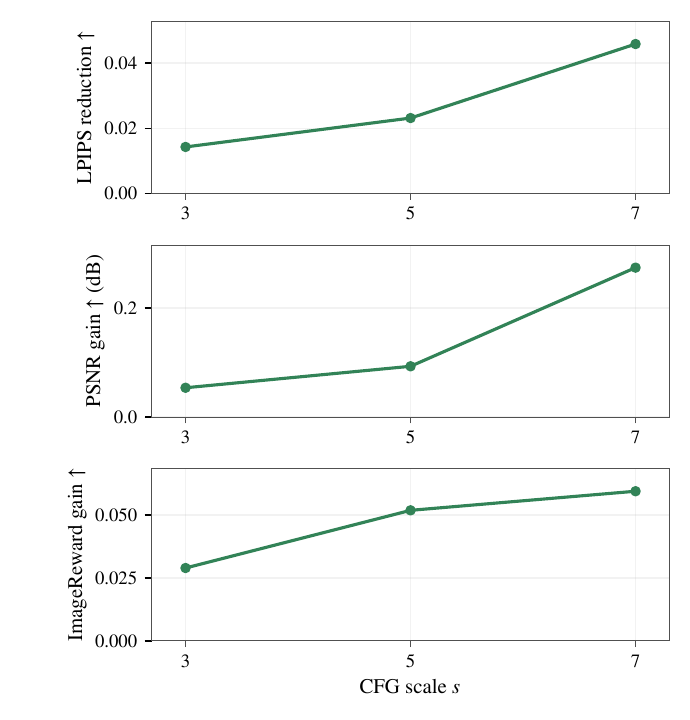}
\caption{\textbf{Sensitivity to CFG guidance strength.} Improvements over SVDQuant W4A4 on Z-Image, using the same 100 prompts at each guidance setting. We show the LPIPS reduction and the PSNR and ImageReward gains. Higher values favor D4D.}
\label{fig:guidance-strength}
\end{figure}

Table~\ref{tab:guidance-strength} reports the metrics, and Figure~\ref{fig:guidance-strength} compares D4D with its quantized baseline. The mean gains from D4D increase across the tested guidance settings, consistent with stronger amplification of the branch difference at higher guidance scales.

\begin{table}[!htbp]
\centering\small
\setlength{\tabcolsep}{3pt}
\caption{\textbf{Effect of CFG guidance strength on Z-Image.} Results on a category-balanced subset of 100 MJHQ-30K prompts. Each quantized output is compared with BF16 at the same guidance setting, prompt, and seed. The CFG scale $s=5$ is used in the main experiments.}
\label{tab:guidance-strength}
\begin{tabular}{clccc}
\toprule
\textbf{$s$} & \textbf{Method} & \textbf{LPIPS$\downarrow$} & \textbf{PSNR$\uparrow$} & \textbf{IR$\uparrow$}\\
\midrule
\multirow{3}{*}{3} & \textit{BF16} & -- & -- & 0.9473\\
 & \methodname{SVDQuant} W4A4 & 0.4263 & 17.29 & 0.7474\\
 & \dFourDours{+ D4D} & \dFourDours{\textbf{0.4120}} & \dFourDours{\textbf{17.34}} & \dFourDours{\textbf{0.7763}}\\
\midrule
\multirow{3}{*}{5} & \textit{BF16} & -- & -- & 1.0753\\
 & \methodname{SVDQuant} W4A4 & 0.4952 & 15.38 & 0.9809\\
 & \dFourDours{+ D4D} & \dFourDours{\textbf{0.4720}} & \dFourDours{\textbf{15.47}} & \dFourDours{\textbf{1.0328}}\\
\midrule
\multirow{3}{*}{7} & \textit{BF16} & -- & -- & 1.0317\\
 & \methodname{SVDQuant} W4A4 & 0.5464 & 14.13 & 1.0197\\
 & \dFourDours{+ D4D} & \dFourDours{\textbf{0.5006}} & \dFourDours{\textbf{14.41}} & \dFourDours{\textbf{1.0791}}\\
\bottomrule
\specialrule{\heavyrulewidth}{1.5pt}{0pt}
\end{tabular}
\end{table}

\section{Efficiency and Numerical Validation}
\label{app:efficiency}
All latency measurements use a GPU with 500 TFLOPS at FP16 and 1.8 TB/s memory bandwidth. All measured tensors and model components remain on the device. We compare matched Triton~\cite{tillet2019triton} paths with and without \methodname{D4D}; FlashInfer~\cite{ye2025flashinfer} and PyTorch provide external NVFP4 and BF16 references, respectively.

\paragraph{Projection benchmark.}
Square projections use $M=8192$ activation rows (4096 tokens per CFG branch, batch size one) and input/output feature width $N\in\{1024,\allowbreak1536,\allowbreak2048,\allowbreak3072,\allowbreak3840,\allowbreak4096,\allowbreak6144,\allowbreak8192\}$. Activations are quantized from BF16 to NVFP4 using \methodname{RTN} with groups of 16. Projection outputs are stored in BF16. Timings include activation packing, GEMM, and \methodname{D4D} reconstruction where applicable; weights are prepared offline. Configuration selection and timing use separate inputs, and Triton configurations remain fixed between \methodname{D4D} variants. After warmup, 31 paired repetitions each measure 50 executions per method in randomized order with synchronized device timing. Figure~\ref{fig:kernel-speedup} reports ratios of median latencies with paired bootstrap intervals.

\paragraph{End-to-end benchmark.}
Z-Image~\cite{team2025zimage} generates $1024\times1024$ images with 28 steps and CFG scale $s=5$; Wan2.2-TI2V-5B~\cite{wan2025repo} generates frames at the same $480\times832$ resolution and guidance scale 5. Both use jointly evaluated CFG branches. Each model uses 20 fixed prompts and seeds, three repetitions, and randomized method order. Timings include prompt encoding, denoising, decoding, and output conversion, excluding model loading, compilation, tuning, and static weight preparation. In BF16, encoding is fused into preceding operations where possible and reconstruction into the GEMM epilogue. All paths share common pipeline optimizations; differences from external library references therefore include integration effects. Table~\ref{tab:runtime-detail} reports paired overhead uncertainty.

\begin{table}[!htbp]
\centering\small
\setlength{\tabcolsep}{4pt}
\caption{\textbf{Uncertainty of end-to-end \methodname{D4D} overhead.} Paired 95\% bootstrap intervals over the 20 prompt clusters (20,000 bootstrap resamples), relative to matched Triton without \methodname{D4D}.}
\label{tab:runtime-detail}
\begin{tabular}{llrr}
\toprule
Model & Format & Overhead (\%) & 95\% interval\\
\midrule
Z-Image & NVFP4 & +0.17 & [+0.15, +0.19]\\
 & BF16 & +0.49 & [+0.46, +0.52]\\
Wan 5B & NVFP4 & +0.21 & [+0.20, +0.22]\\
 & BF16 & +1.40 & [+1.39, +1.40]\\
\bottomrule
\end{tabular}
\end{table}

\paragraph{Numerical validation.}
Packed values and scales are checked against the reference quantizer, and projection outputs against independently decoded FP32 products. Fused implementations are checked against their unfused counterparts before timing. These checks do not imply equality between \methodname{D4D} and untransformed BF16 in finite precision. The BF16 end-to-end overhead benefits from fusion with preceding operations; the NVFP4 projection overhead bound does not extend to standalone BF16 kernels.

\section{Comparison with GCBT}
\label{app:concurrent}
D4D and concurrent GCBT both operate in CFG branch space. GCBT~\cite{jiang2026gcbt} selects layer-specific orthogonal branch transforms from uncentered moments $\Sigma_\ell=\mathbb{E}[xx^\top]$ and guidance $g_s=(s,1-s)^\top$. For orthonormal rows $r_i^\top$, its surrogate is
\begin{equation}
 D_{\mathrm{sur}}(R)=\sum_{i=1}^2(r_i^\top g_s)^2(r_i^\top\Sigma_\ell r_i).
\end{equation}
It assumes error variances proportional to coordinate second moments and neglects cross-coordinate errors. Its optimum is $\theta_\ell^\star=(\phi_s+\psi_\ell)/2-\pi/4$ modulo $\pi/2$, where $\psi_\ell$ is the dominant-eigenvector angle and $\phi_s=\operatorname{atan2}(1-s,s)$. Optimality concerns this surrogate, not generation quality.

D4D explicitly exposes the branch difference amplified by CFG, using fixed mean--difference directions without transform statistics or guidance-dependent coefficients. Orthogonal transforms can also expose the difference. D4D's unequal normalization, $TT^\top=\operatorname{diag}(1/2,2)$, has quantizer-dependent effects. Both methods retain host-quantizer calibration.

\paragraph{Matched image comparison.}
Table~\ref{tab:zimage-gcbt} compares our GCBT implementation with D4D on 100 MJHQ prompts using Z-Image and frozen SVDQuant W4A4 weights. Prompts, seeds, BF16 references, $1024\times1024$ resolution, 28 steps, and CFG $s=5$ match. GCBT moments use all aligned image-token values after host smoothing across all 28 native-quantized trajectory steps on 64 calibration prompts. These and the 20 host-weight calibration prompts exclude evaluation. Both transforms act on 224 projections. Text tokens retain branch coordinates and the BF16 low-rank path is unchanged. The analytic solution matches dense-grid optimization over 20,001 angles on all 224 saved statistics. D4D has lower LPIPS on this subset. The table reports paired intervals for both metrics. These 100 prompts are a different subset from the one used in App.~\ref{app:guidance-strength}, so absolute values differ between the two tables. This supports the fixed transform in this configuration, without establishing general superiority.

\paragraph{Angles selected by GCBT.}
For each projection, GCBT selects a rotation angle $\theta_\ell^\star$ from activation statistics. Tab.~\ref{tab:gcbt-angles} summarizes these angles on the D4D-eligible projections of all six evaluated models, shown modulo $90^\circ$ so that the sum--difference directions of D4D and normalized Hadamard are at $45^\circ$. For Z-Image the angles come from the GCBT statistics of the matched comparison above; for the other models they are computed with GCBT's formula on BF16 activations of the calibration examples (App.~\ref{app:protocols}). On every model, 90\% of the selected angles lie within $6^\circ$ of the sum--difference directions, and the median offset is between $1.9^\circ$ and $5.6^\circ$. The statistics-driven rotation thus lands close to the fixed directions that D4D uses without any statistics.

\begin{table}[!htbp]
\centering\small
\setlength{\tabcolsep}{4pt}
\caption{\textbf{Branch-rotation angles selected by GCBT} on D4D-eligible projections (degrees, modulo $90^\circ$; sum--difference directions at $45^\circ$). Z-Image: statistics of the matched comparison in Tab.~\ref{tab:zimage-gcbt}; other models: diagnostic fit on BF16 calibration activations.}
\label{tab:gcbt-angles}
\begin{tabular}{lrccc}
\toprule
Model & Proj. & Median & 5\% & 95\%\\
\midrule
Z-Image & 224 & 48.2 & 47.1 & 48.8\\
Chroma1-HD & 304 & 50.6 & 49.9 & 50.9\\
Wan2.2-TI2V-5B & 90 & 48.2 & 47.8 & 48.3\\
Wan2.2-I2V-A14B & 240 & 49.7 & 48.5 & 50.1\\
Pixal3D & 384 & 47.1 & 44.7 & 47.4\\
3DTopia-XL & 112 & 46.9 & 45.1 & 47.8\\
\bottomrule
\end{tabular}
\end{table}

\begin{table}[!htbp]
\centering\small
\setlength{\tabcolsep}{3pt}
\caption{\textbf{Branch-space transforms on Z-Image with SVDQuant W4A4.} Means over 100 matched prompts. The final rows report D4D minus GCBT with paired 95\% percentile bootstrap intervals over prompts using 20,000 resamples. $^{*}$Our implementation of GCBT's published formulation.}
\label{tab:zimage-gcbt}
\begin{tabular}{lcc}
\toprule
\textbf{Method} & \textbf{LPIPS$\downarrow$} & \textbf{PSNR (dB)$\uparrow$}\\
\midrule
\methodname{SVDQuant} & 0.4893 & 15.429\\
\dFourDothers{+ GCBT$^{*}$} & \dFourDothers{0.4658} & \dFourDothers{15.642}\\
\dFourDours{+ D4D} & \dFourDours{0.4529} & \dFourDours{15.709}\\
\midrule
$\Delta$ & $-0.01287$ & $+0.06698$\\
95\% interval & $[-0.02241,-0.00338]$ & $[-0.07452,+0.21870]$\\
\bottomrule
\specialrule{\heavyrulewidth}{1.5pt}{0pt}
\end{tabular}
\end{table}

\paragraph{Video comparison.}
We compare GCBT and D4D on a subset of 16 VBench-I2V image--prompt pairs using Wan2.2-TI2V-5B with MACaQ W4A4 (Table~\ref{tab:wan-gcbt}). Weights, inputs, sampling, and transformed projections match. GCBT statistics use 64 training conditions at steps 4, 20, and 36, passing cached BF16 trajectory states through the native quantized model. LPIPS averages eight frames per clip; latent NMSE compares final latents with BF16.

\begin{table}[!htbp]
\centering\small
\setlength{\tabcolsep}{18.06763pt}
\caption{\textbf{Branch-space comparison on Wan2.2-TI2V-5B.} Mean fidelity to BF16 on a subset of 16 VBench-I2V image--prompt pairs. $^{*}$Our implementation of GCBT's published formulation.}
\label{tab:wan-gcbt}
\begin{tabular}{lcc}
\toprule
\textbf{Method} & \textbf{LPIPS$\downarrow$} & \textbf{Latent NMSE$\downarrow$}\\
\midrule
\methodname{MACaQ} & 0.2693 & 0.3126\\
\dFourDothers{+ GCBT$^{*}$} & \dFourDothers{0.2173} & \dFourDothers{0.2674}\\
\dFourDours{+ D4D} & \dFourDours{0.2078} & \dFourDours{0.2386}\\
\bottomrule
\specialrule{\heavyrulewidth}{1.5pt}{0pt}
\end{tabular}
\end{table}

\clearpage
\onecolumn

\section{Additional Qualitative Results}
\label{app:qualitative}
We provide additional qualitative comparisons of BF16 outputs and quantized outputs with and without D4D. How often D4D helps across all prompts is quantified by the paired statistics and per-prompt win rates of App.~\ref{app:extended-results}.

\par\addvspace{10pt}
\noindent\begin{minipage}{\textwidth}
\noindent
\begin{minipage}[t]{0.48\textwidth}
\centering\textbf{Example 1}\par\smallskip
\begingroup\scriptsize\setlength{\tabcolsep}{2pt}
\begin{tabular}{@{}ccc@{}}
\textbf{BF16} & \shortstack{\textbf{W4A4}\\\textbf{without D4D}} & \shortstack{\textbf{W4A4}\\\textbf{with D4D}}\\[2pt]
 & \multicolumn{2}{c}{SVDQuant}\\[1pt]
\includegraphics[width=0.30\linewidth]{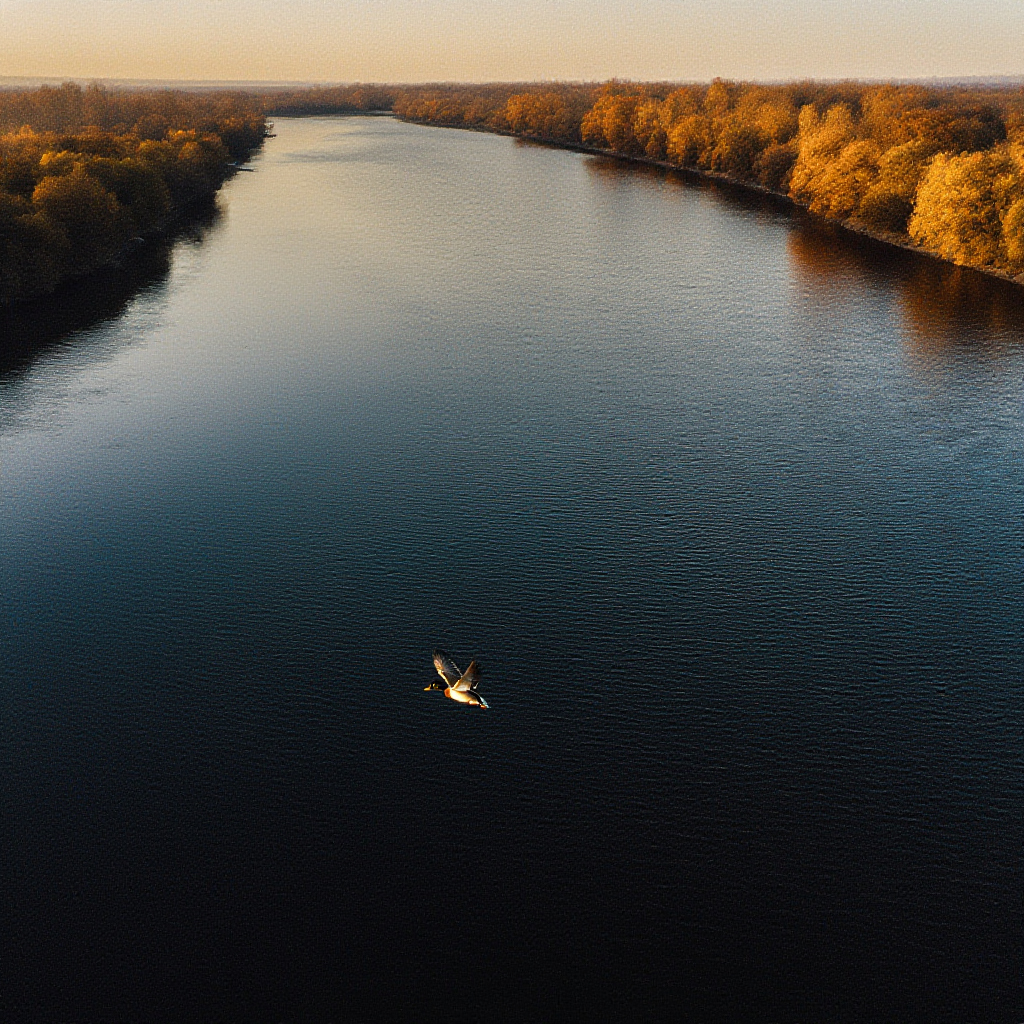} & \includegraphics[width=0.30\linewidth]{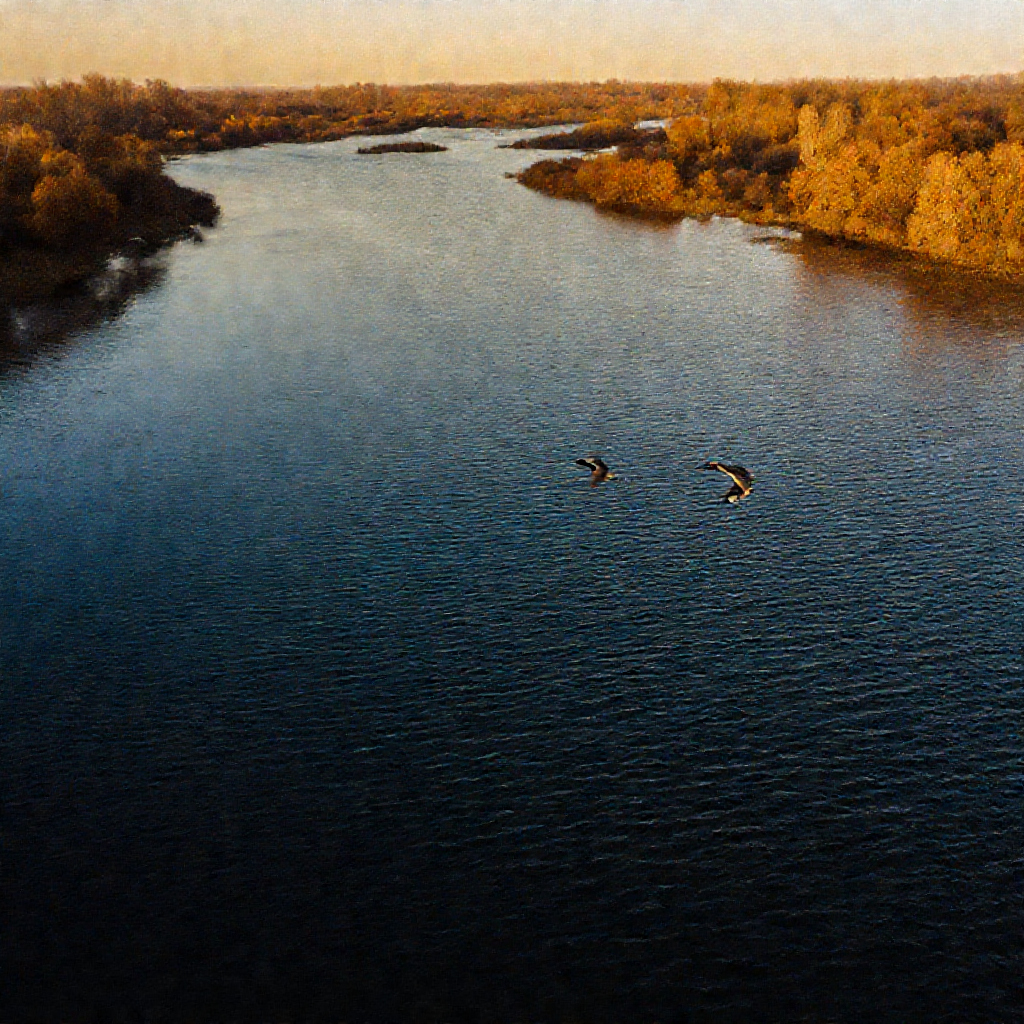} & \includegraphics[width=0.30\linewidth]{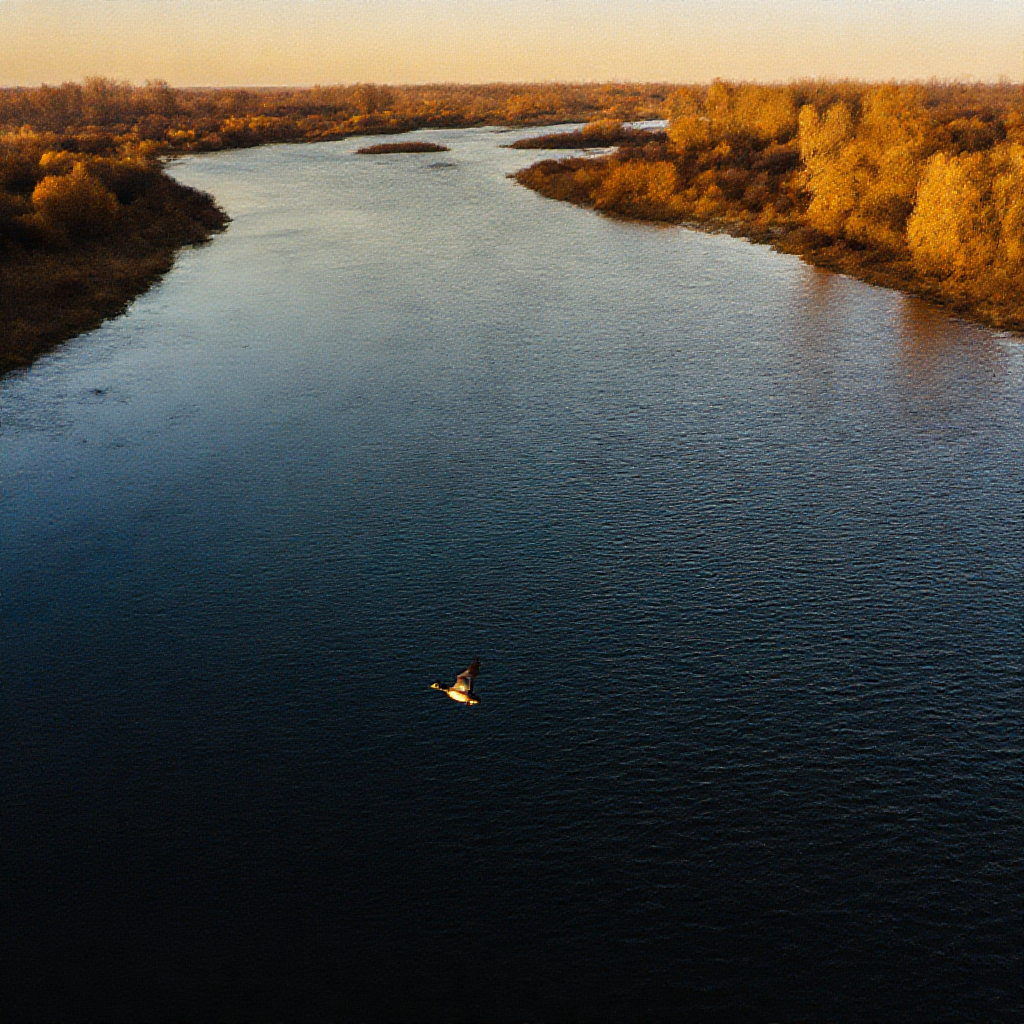}\\[2pt]
 & \multicolumn{2}{c}{Q-DiT}\\[1pt]
 & \includegraphics[width=0.30\linewidth]{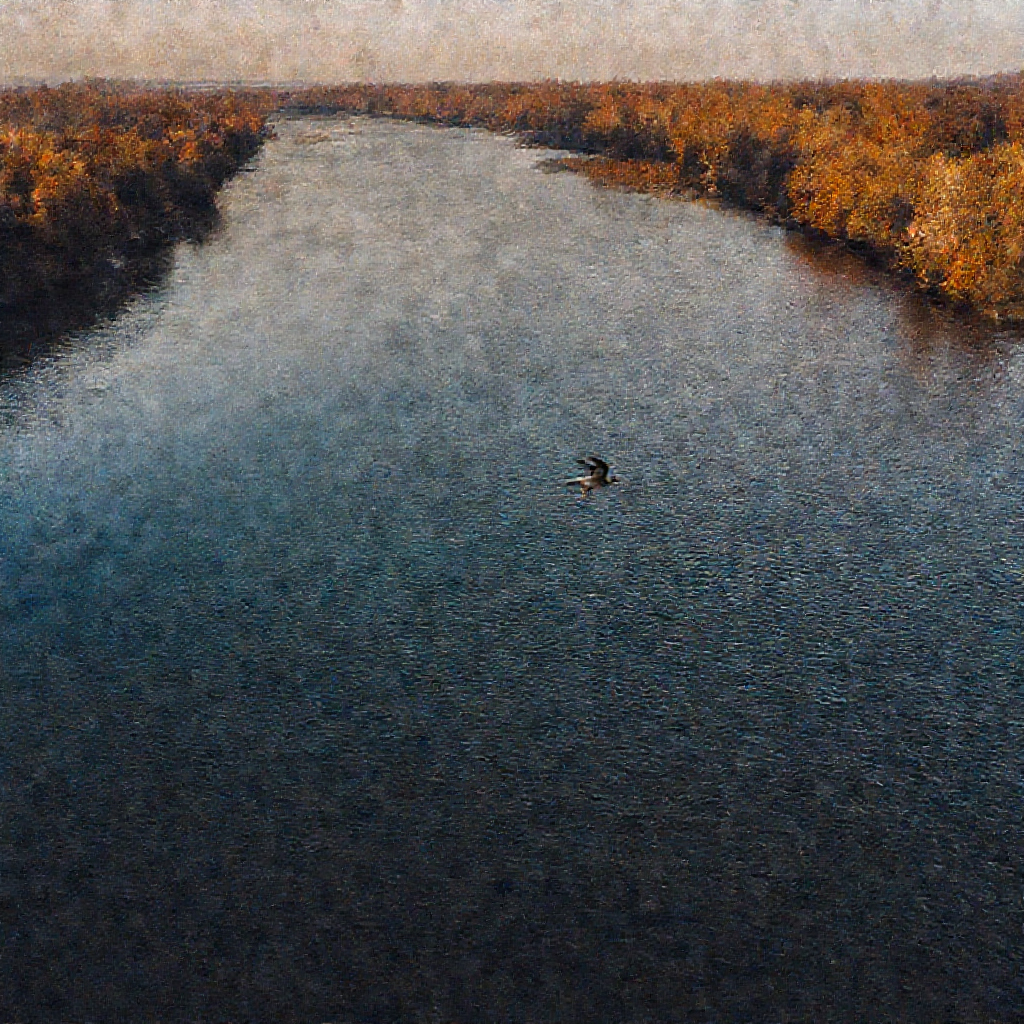} & \includegraphics[width=0.30\linewidth]{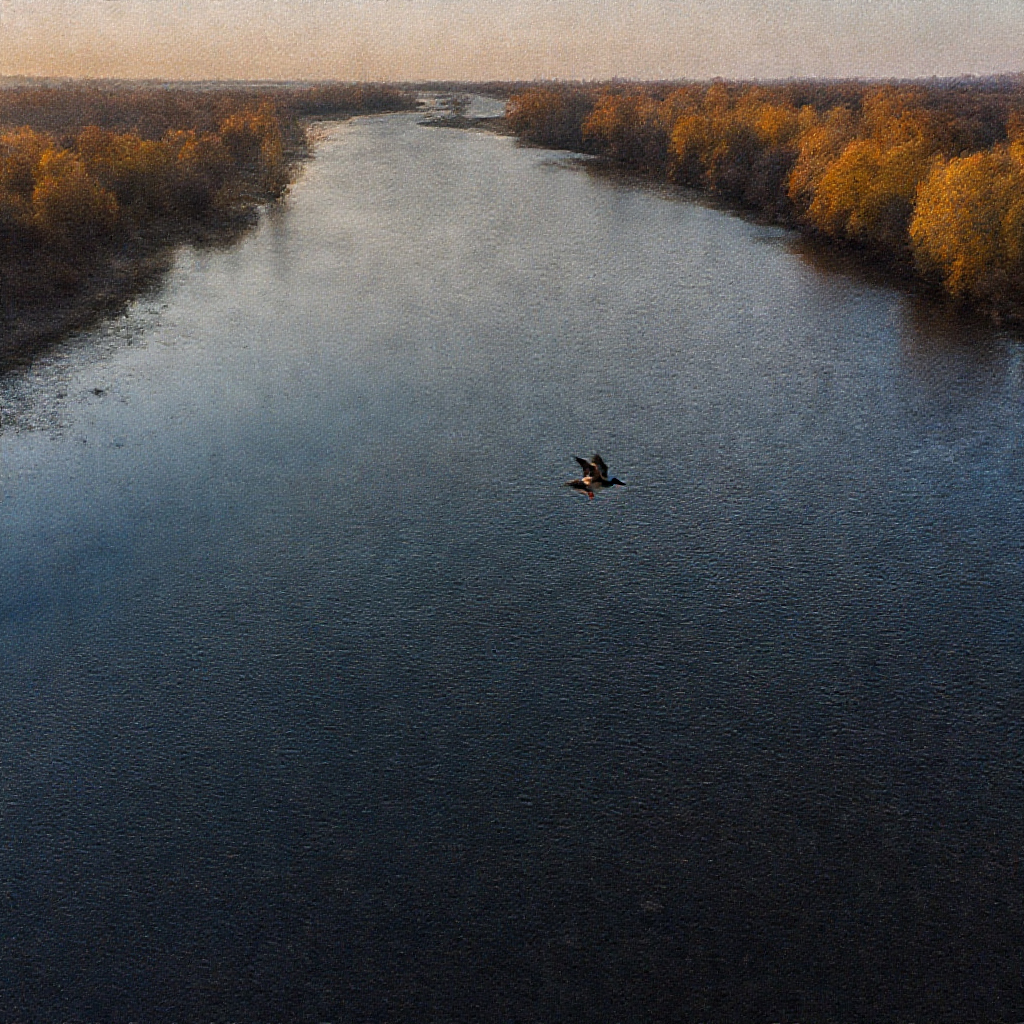}\\[2pt]
 & \multicolumn{2}{c}{MACaQ}\\[1pt]
 & \includegraphics[width=0.30\linewidth]{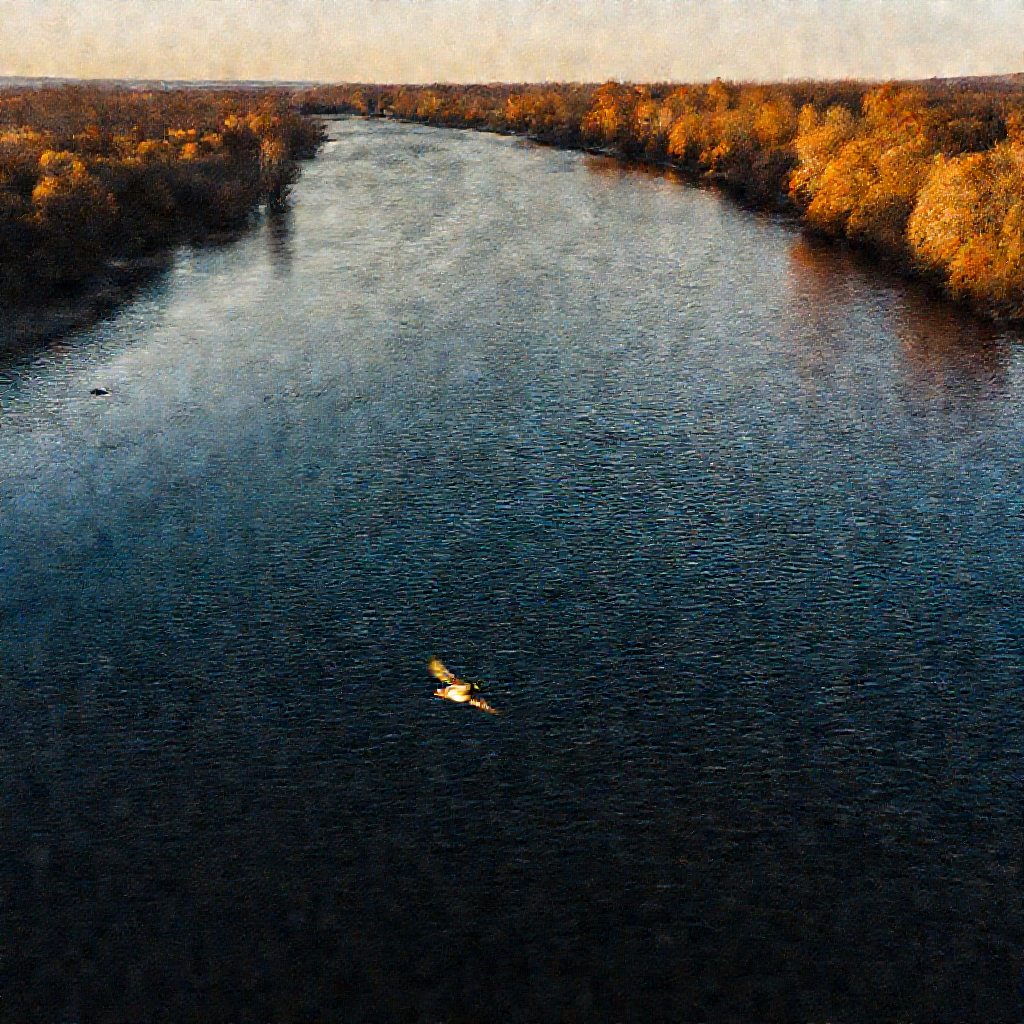} & \includegraphics[width=0.30\linewidth]{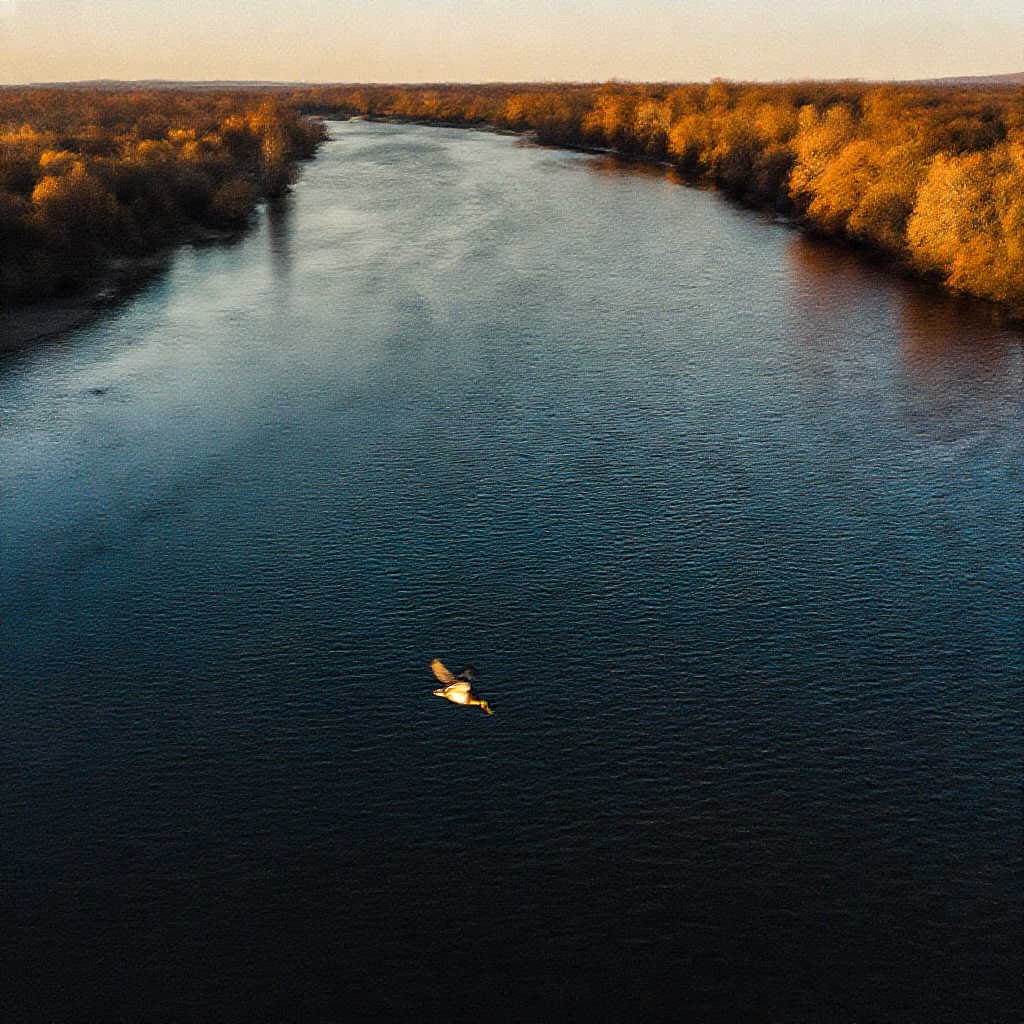}\\[2pt]
\end{tabular}
\endgroup
\end{minipage}
\hfill
\begin{minipage}[t]{0.48\textwidth}
\centering\textbf{Example 2}\par\smallskip
\begingroup\scriptsize\setlength{\tabcolsep}{2pt}
\begin{tabular}{@{}ccc@{}}
\textbf{BF16} & \shortstack{\textbf{W4A4}\\\textbf{without D4D}} & \shortstack{\textbf{W4A4}\\\textbf{with D4D}}\\[2pt]
 & \multicolumn{2}{c}{SVDQuant}\\[1pt]
\includegraphics[width=0.30\linewidth]{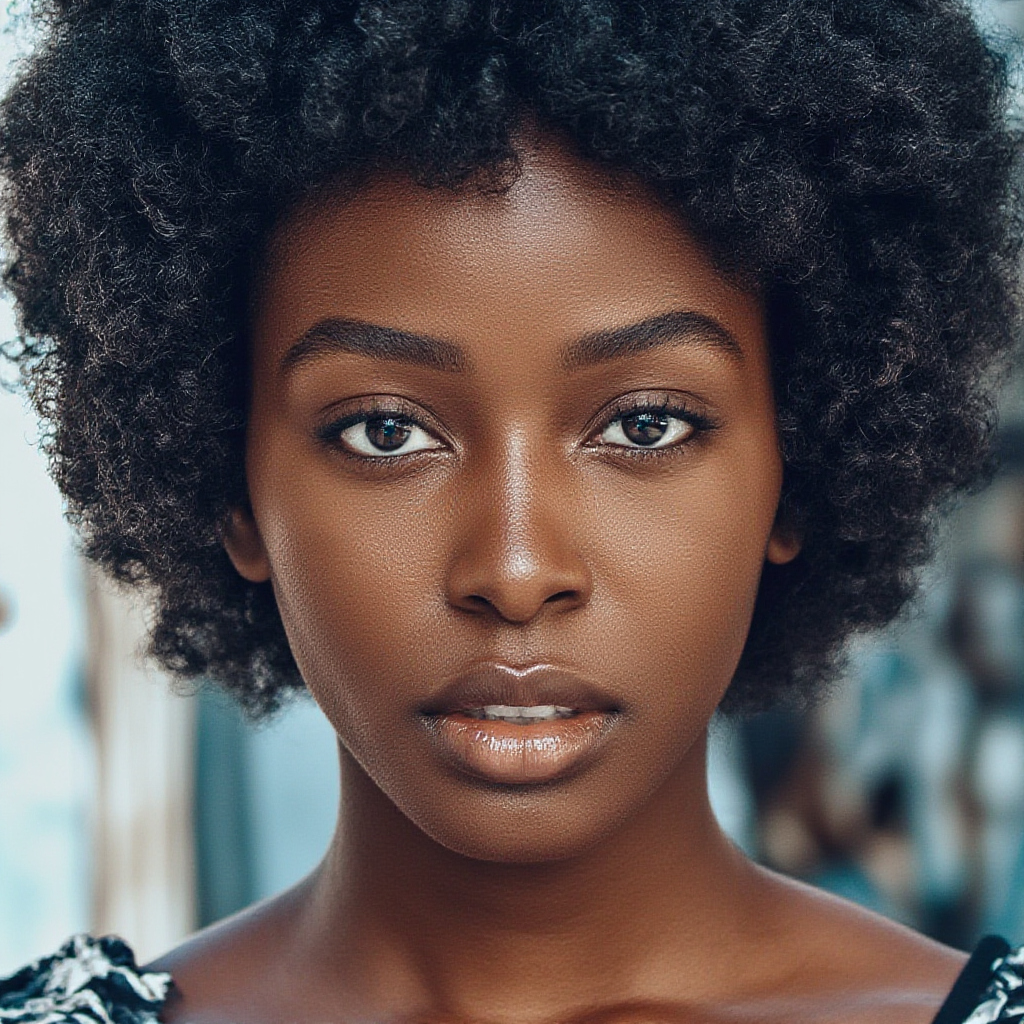} & \includegraphics[width=0.30\linewidth]{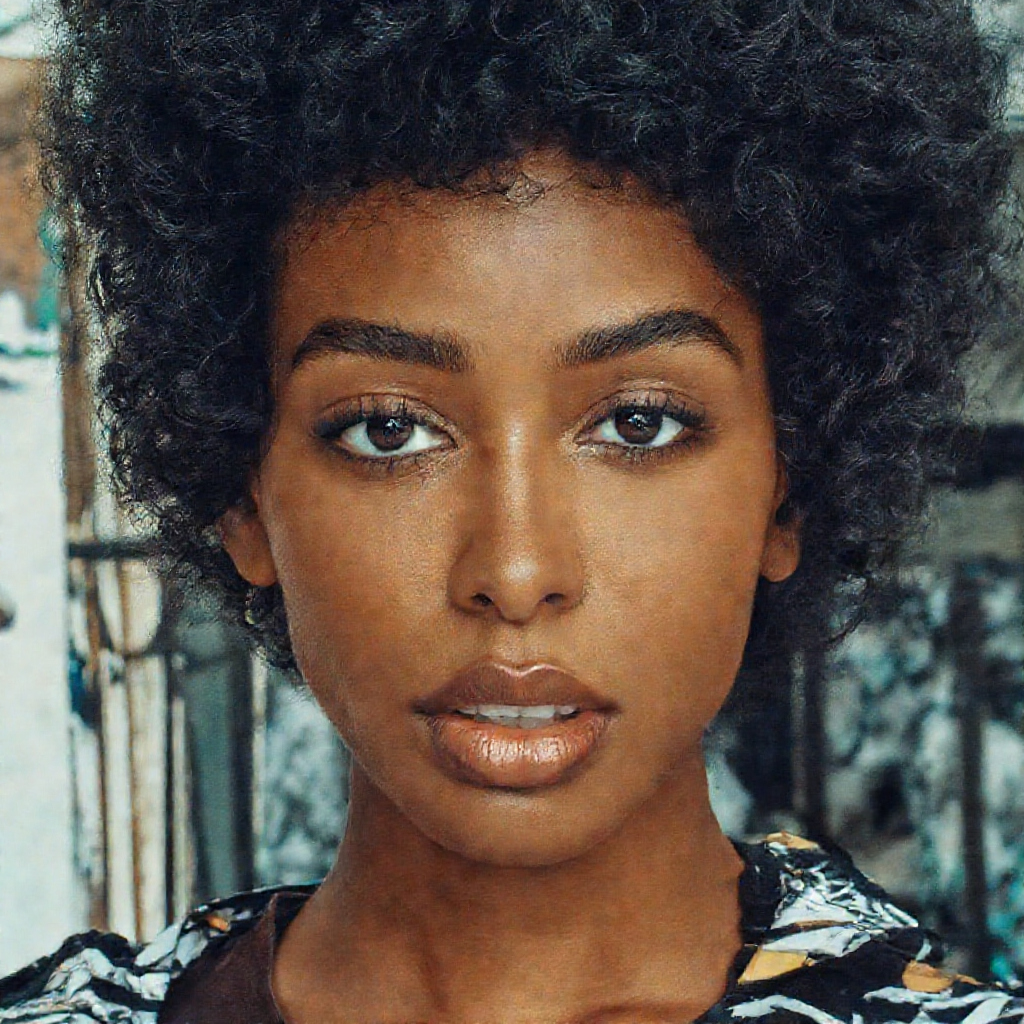} & \includegraphics[width=0.30\linewidth]{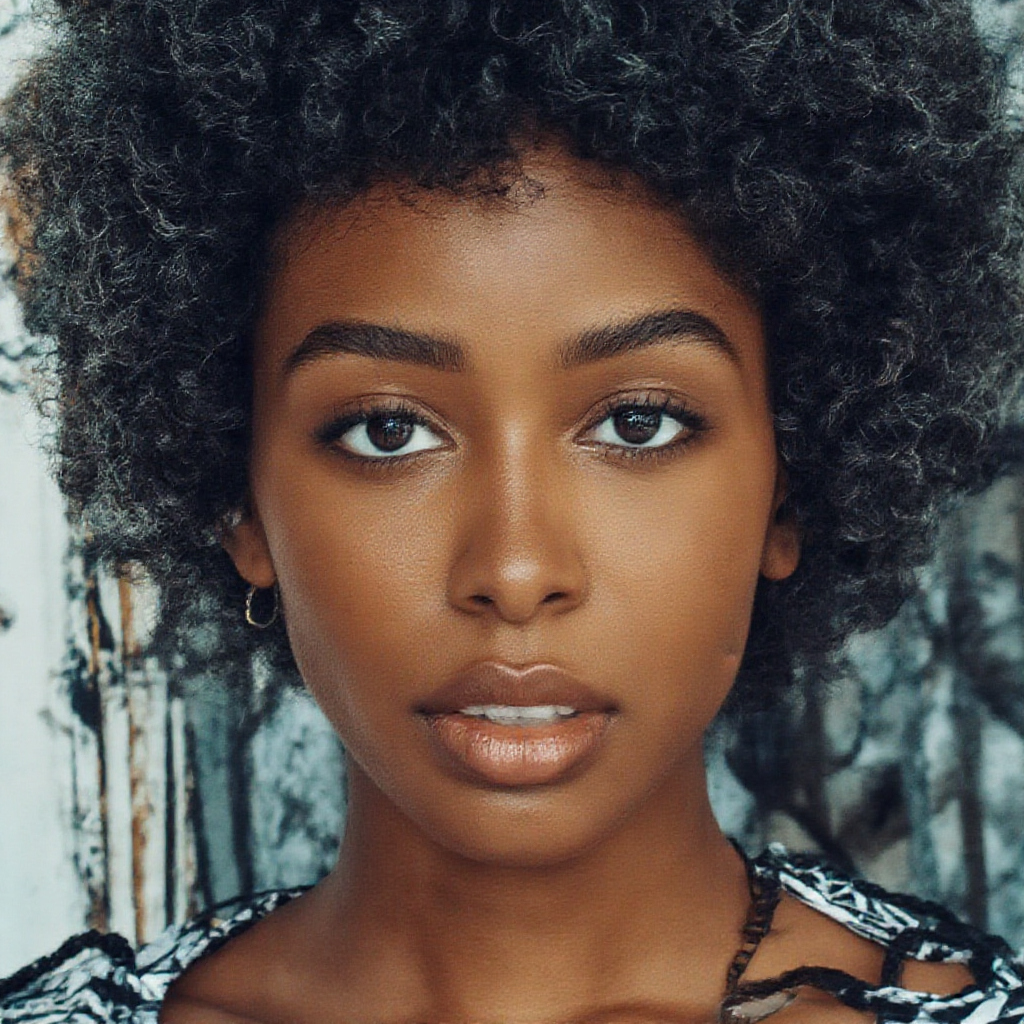}\\[2pt]
 & \multicolumn{2}{c}{Q-DiT}\\[1pt]
 & \includegraphics[width=0.30\linewidth]{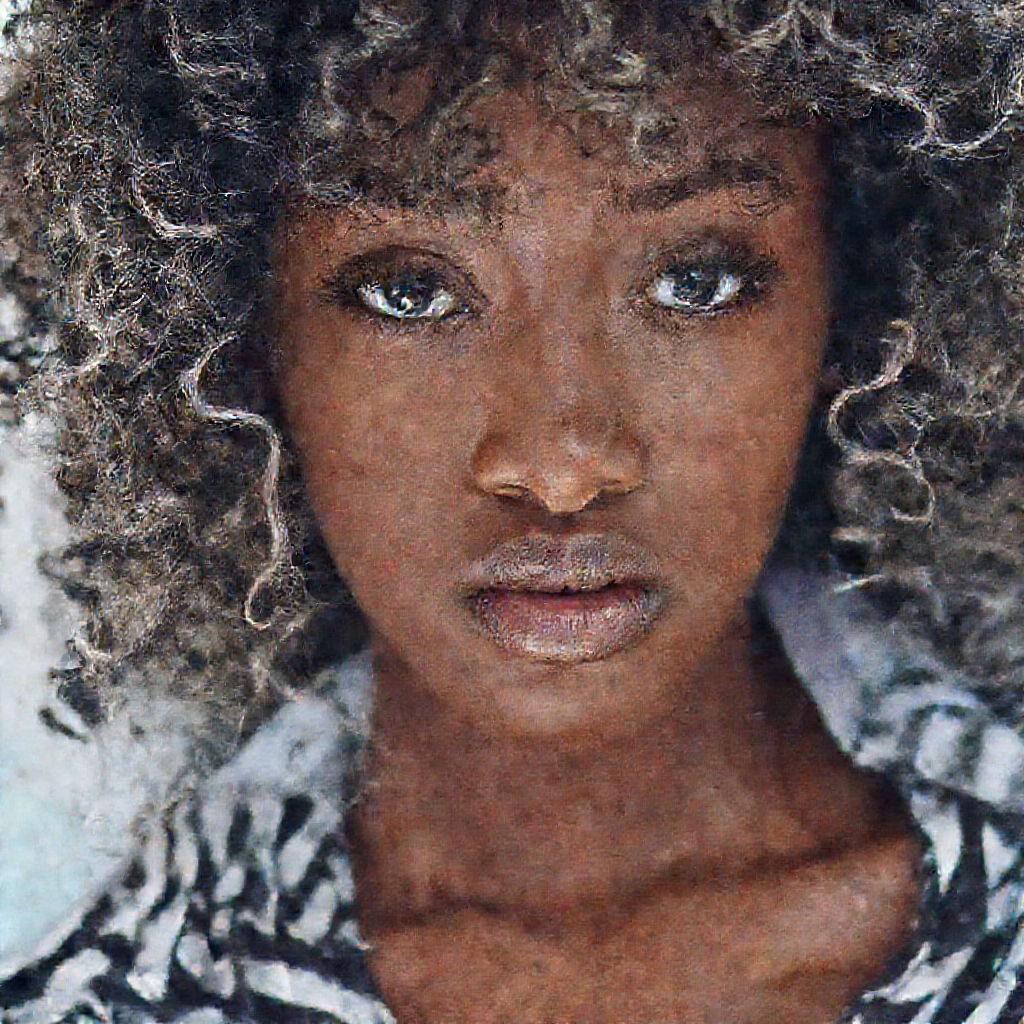} & \includegraphics[width=0.30\linewidth]{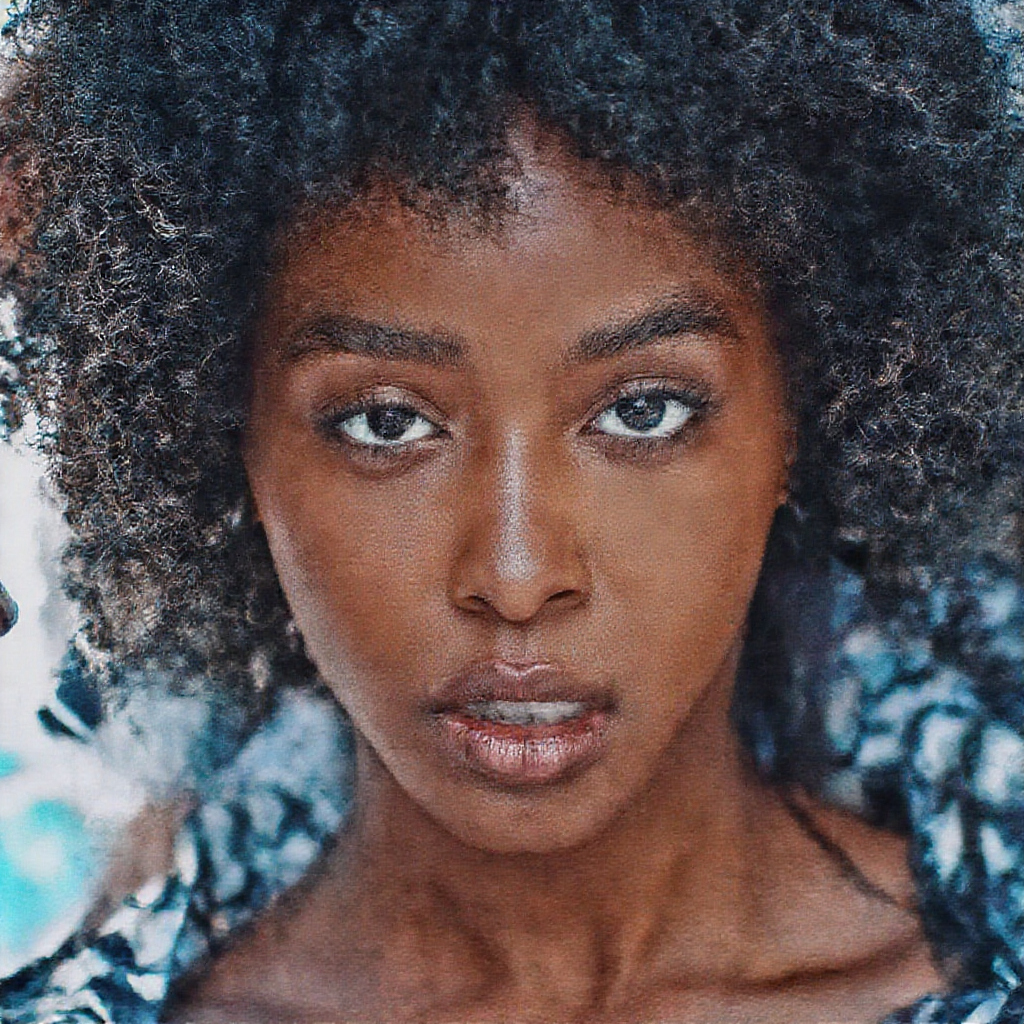}\\[2pt]
 & \multicolumn{2}{c}{MACaQ}\\[1pt]
 & \includegraphics[width=0.30\linewidth]{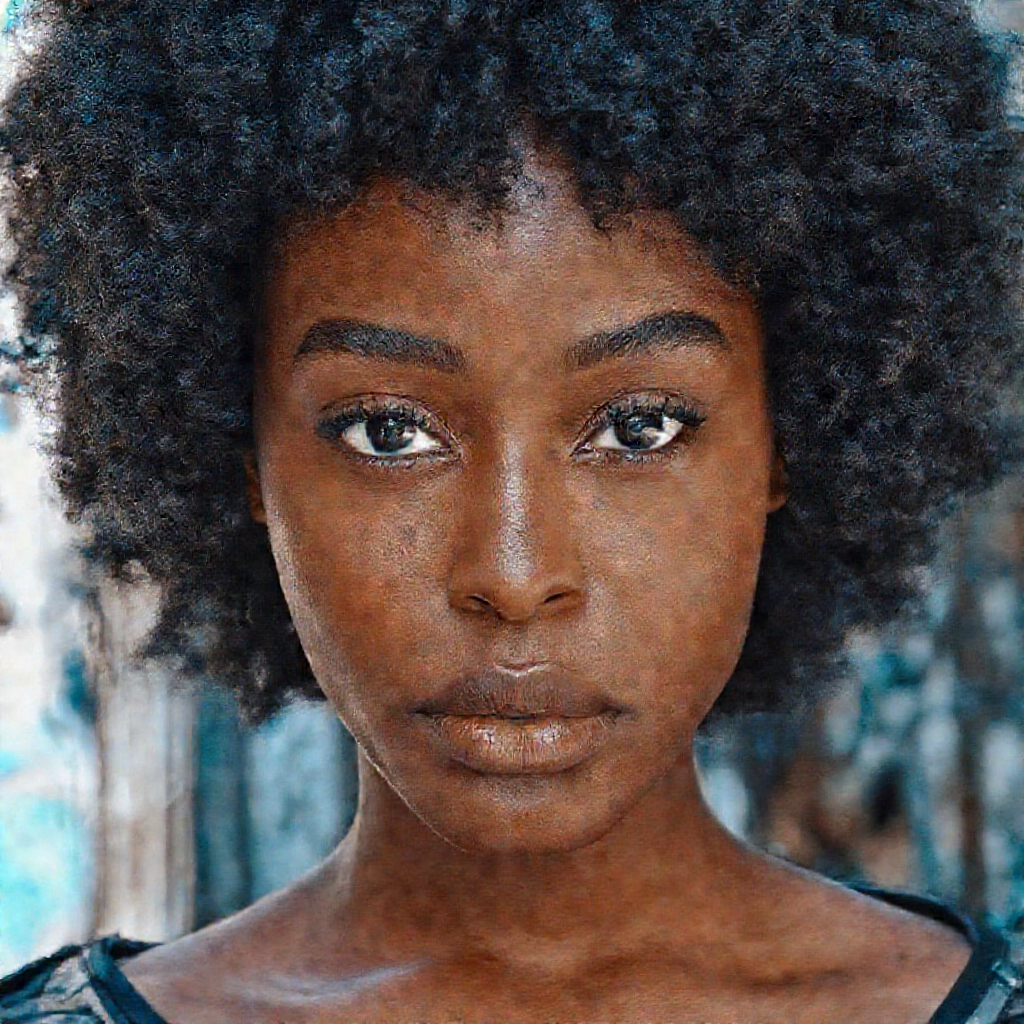} & \includegraphics[width=0.30\linewidth]{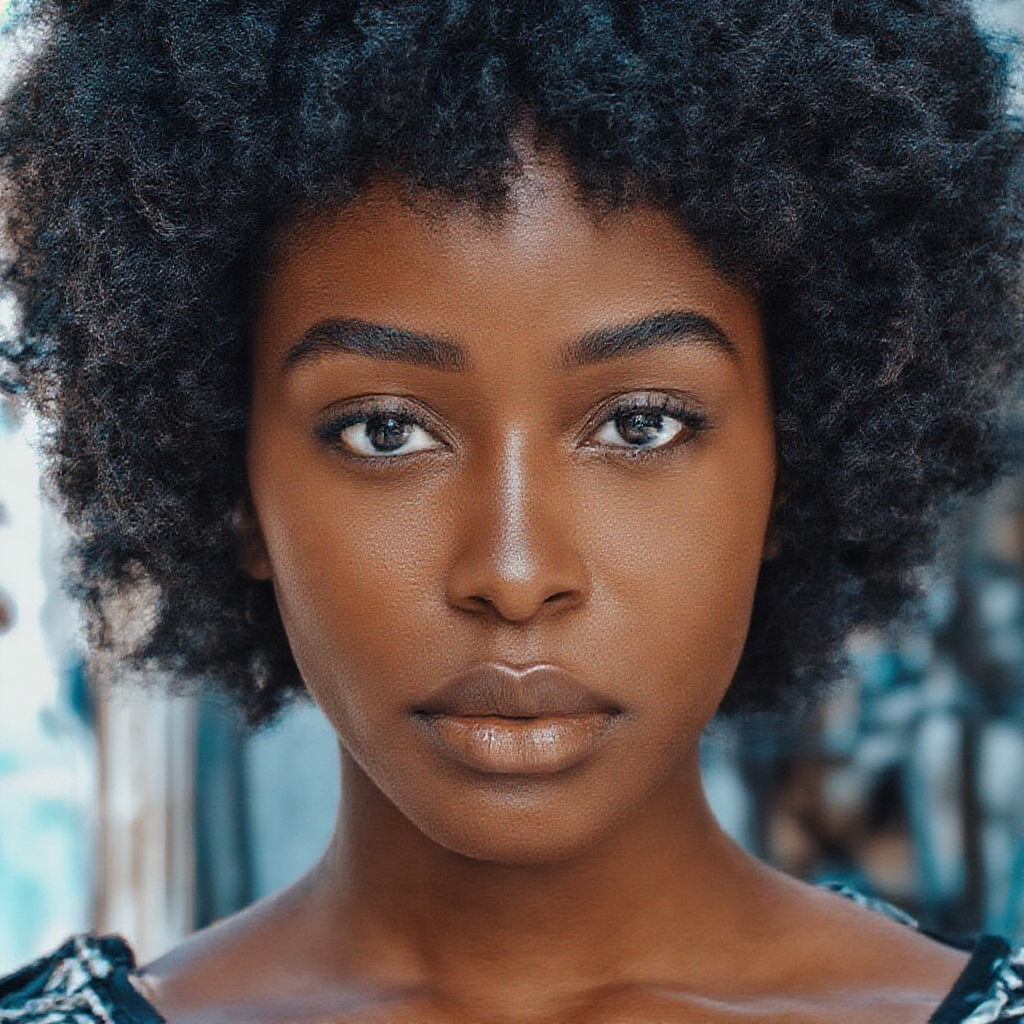}\\[2pt]
\end{tabular}
\endgroup
\end{minipage}
\par\smallskip
\captionof{figure}{\textbf{Additional Z-Image comparisons.} Two Z-Image-generated examples with BF16 references and all evaluated W4A4 quantizers, with and without D4D. Each example uses the same prompt and seed across methods.}\label{fig:qual-extra-zimage}
\end{minipage}\par

\par\addvspace{10pt}
\noindent\begin{minipage}{\textwidth}
\noindent
\begin{minipage}[t]{0.48\textwidth}
\centering\textbf{Example 1}\par\smallskip
\begingroup\scriptsize\setlength{\tabcolsep}{2pt}
\begin{tabular}{@{}ccc@{}}
\textbf{BF16} & \shortstack{\textbf{W4A4}\\\textbf{without D4D}} & \shortstack{\textbf{W4A4}\\\textbf{with D4D}}\\[2pt]
 & \multicolumn{2}{c}{SVDQuant}\\[1pt]
\includegraphics[width=0.30\linewidth]{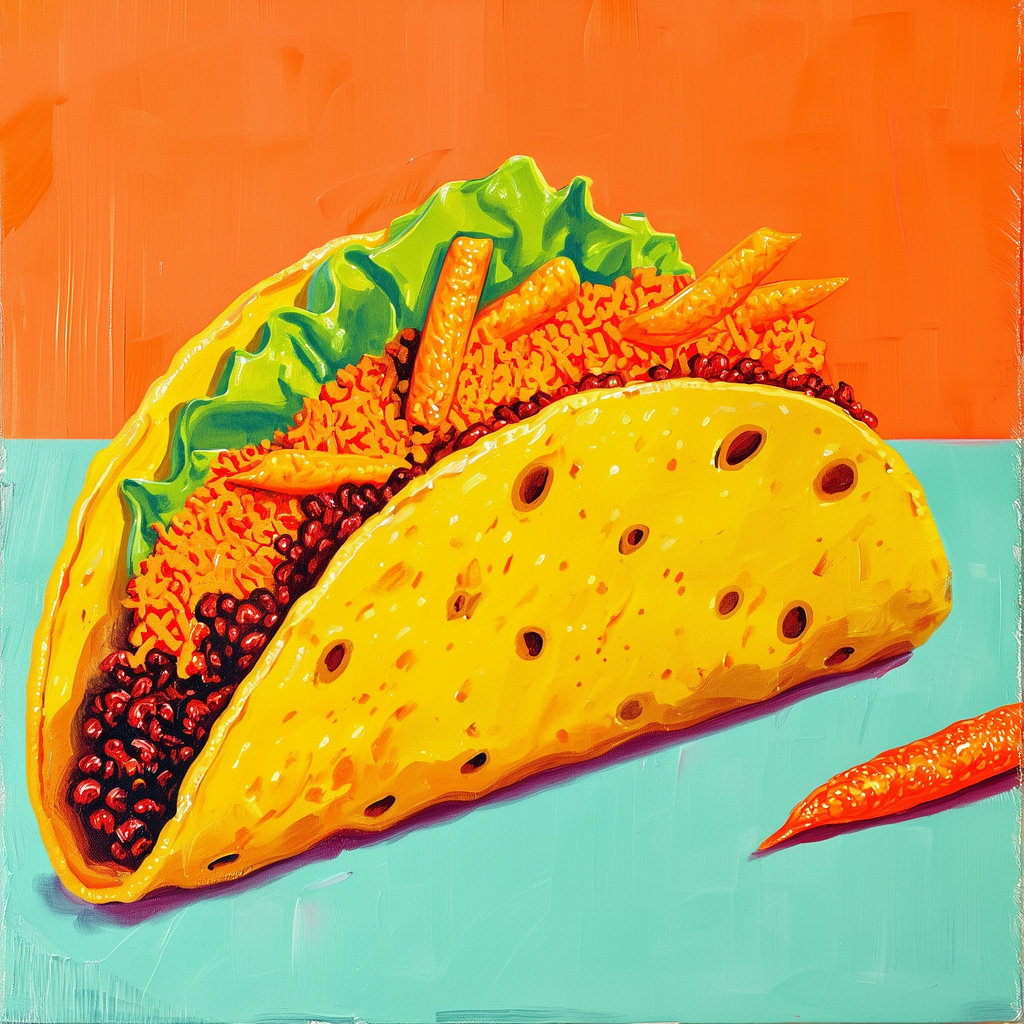} & \includegraphics[width=0.30\linewidth]{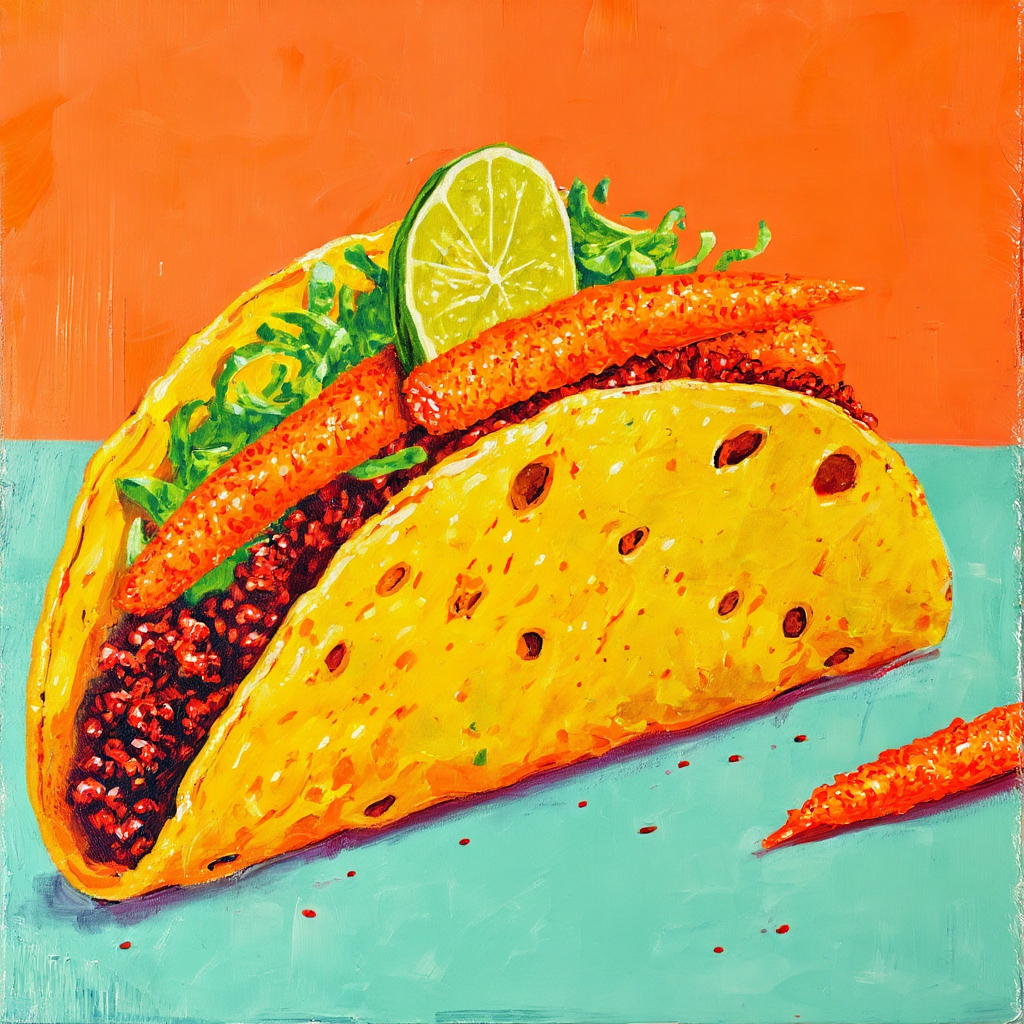} & \includegraphics[width=0.30\linewidth]{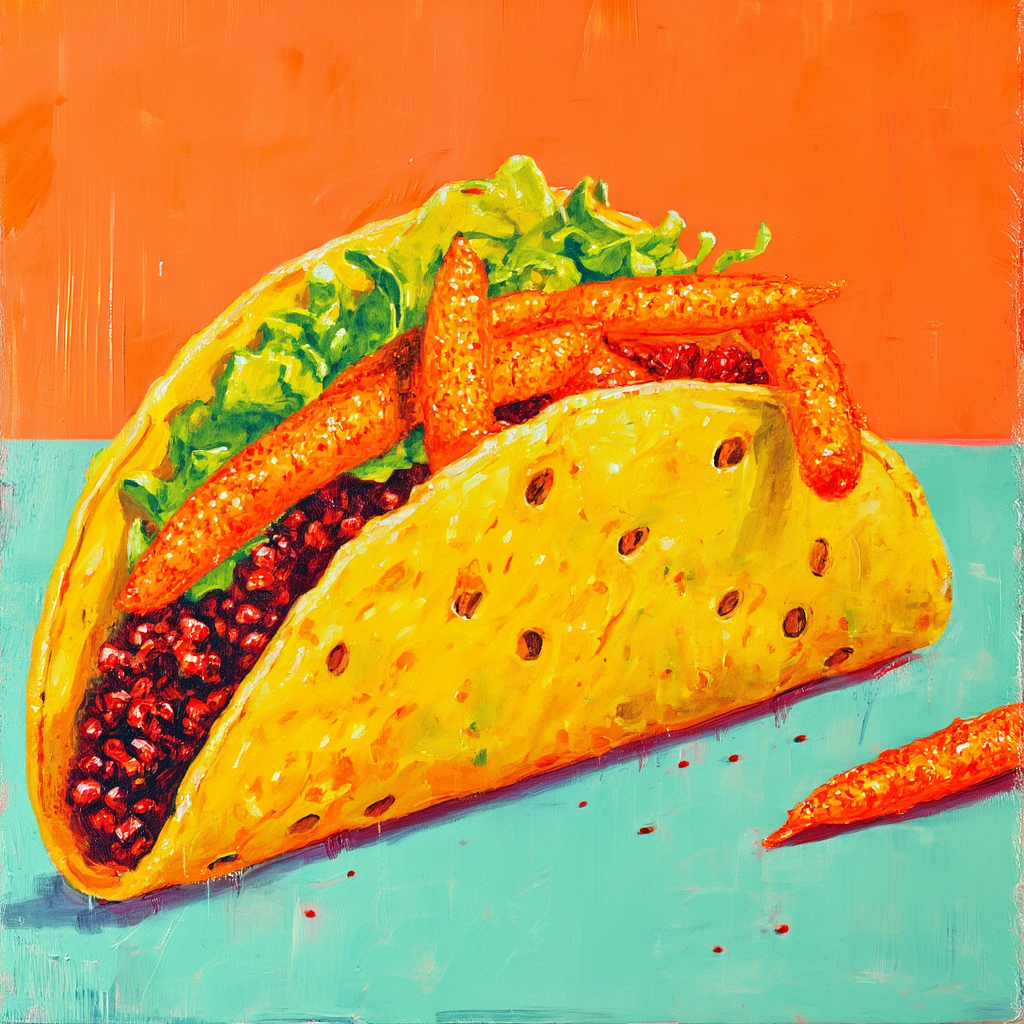}\\[2pt]
 & \multicolumn{2}{c}{Q-DiT}\\[1pt]
 & \includegraphics[width=0.30\linewidth]{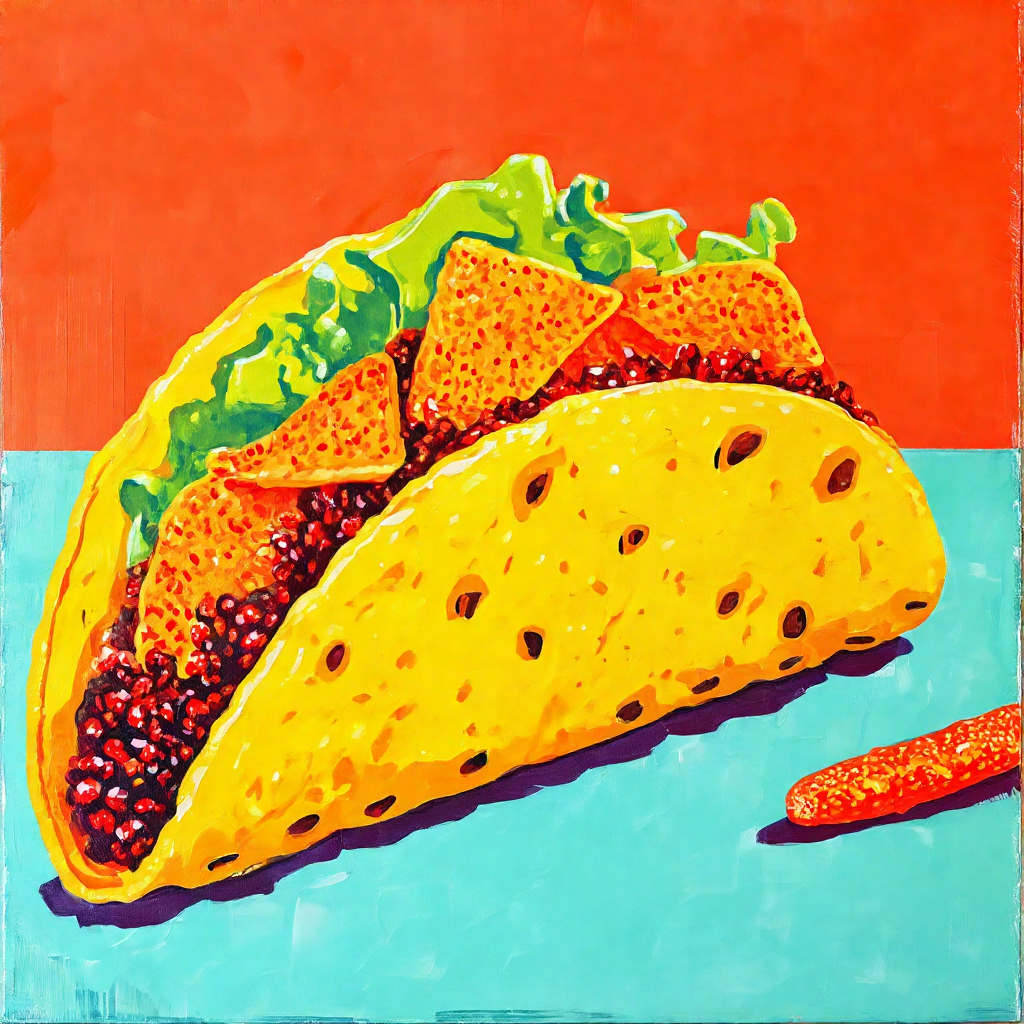} & \includegraphics[width=0.30\linewidth]{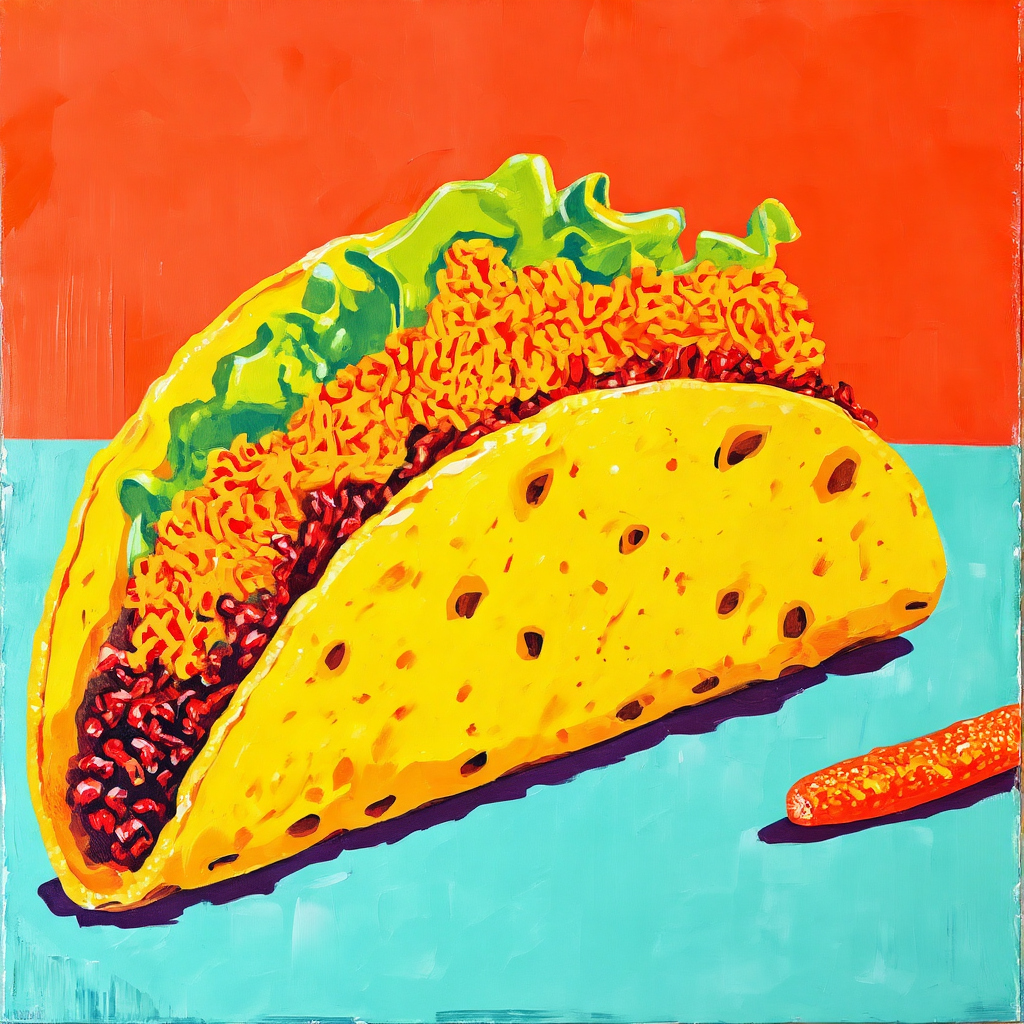}\\[2pt]
 & \multicolumn{2}{c}{MACaQ}\\[1pt]
 & \includegraphics[width=0.30\linewidth]{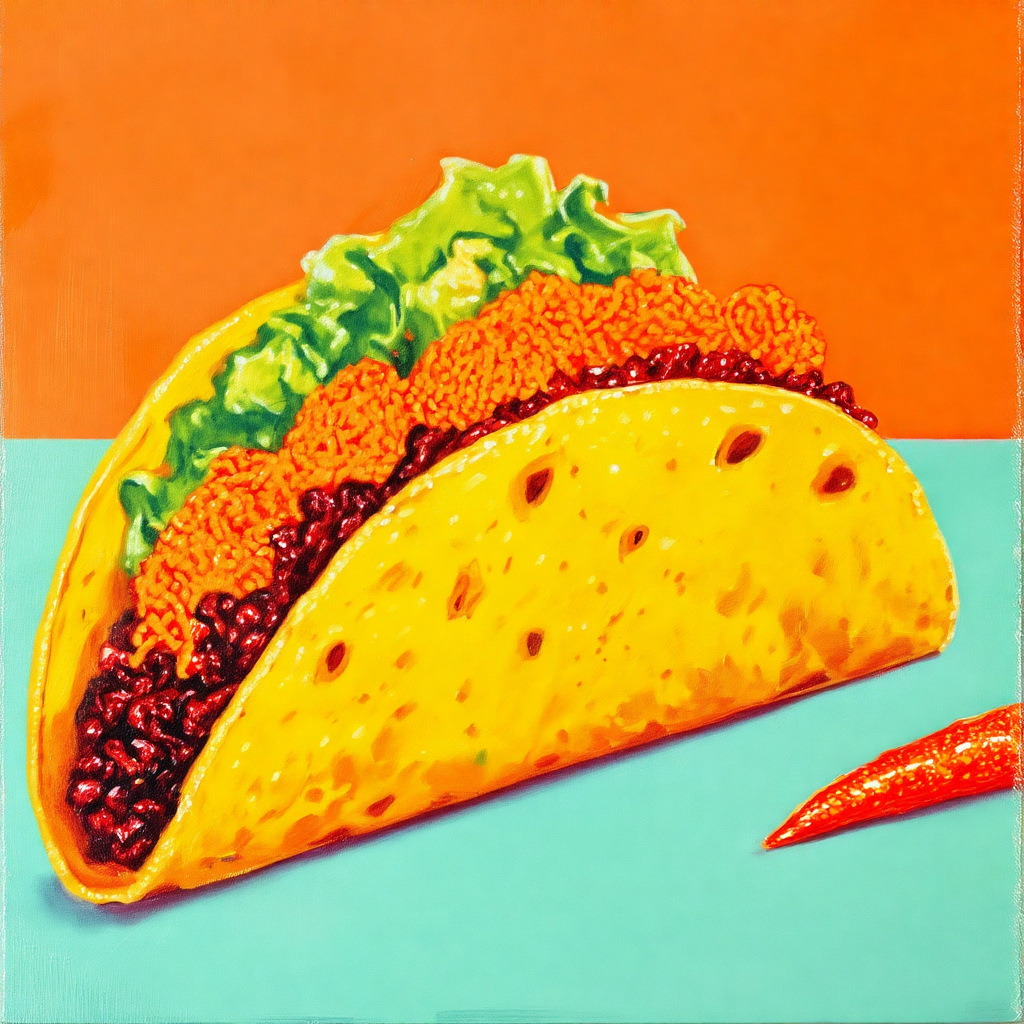} & \includegraphics[width=0.30\linewidth]{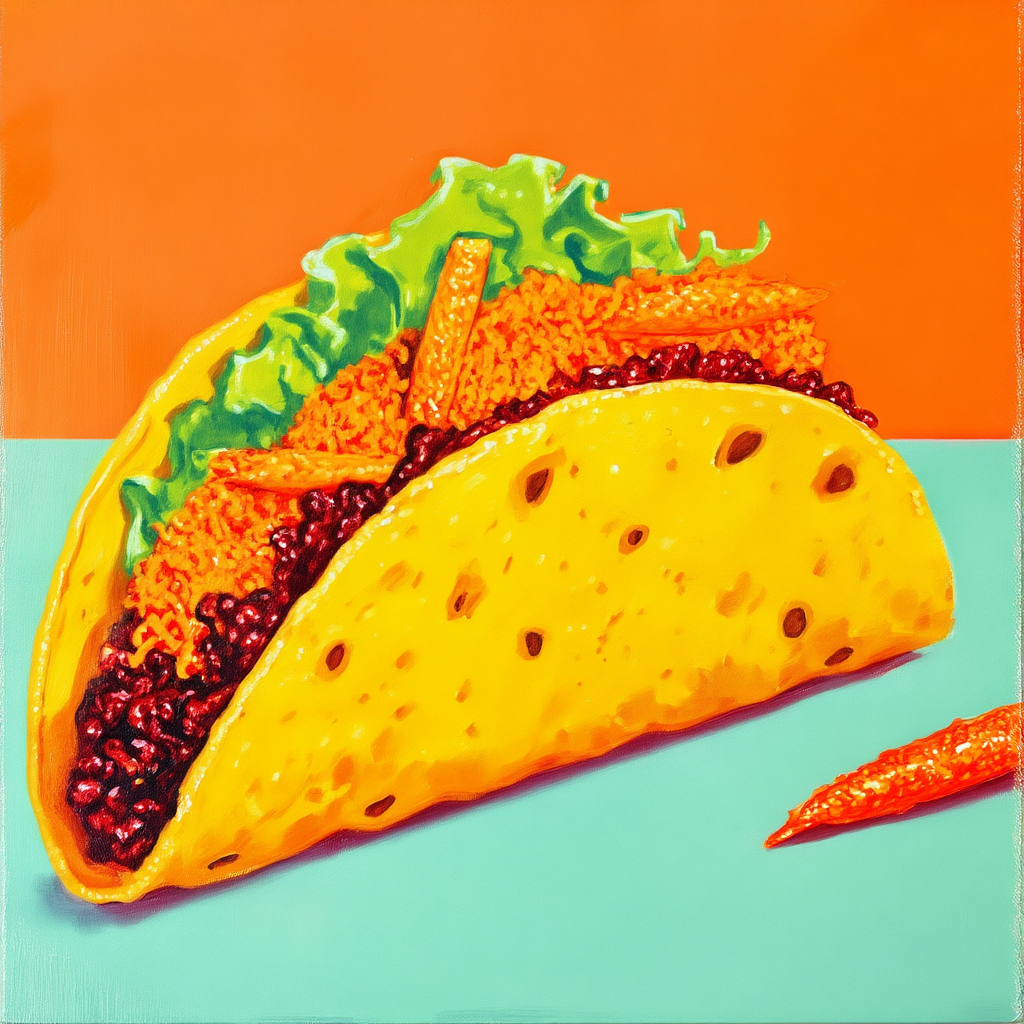}\\[2pt]
\end{tabular}
\endgroup
\end{minipage}
\hfill
\begin{minipage}[t]{0.48\textwidth}
\centering\textbf{Example 2}\par\smallskip
\begingroup\scriptsize\setlength{\tabcolsep}{2pt}
\begin{tabular}{@{}ccc@{}}
\textbf{BF16} & \shortstack{\textbf{W4A4}\\\textbf{without D4D}} & \shortstack{\textbf{W4A4}\\\textbf{with D4D}}\\[2pt]
 & \multicolumn{2}{c}{SVDQuant}\\[1pt]
\includegraphics[width=0.30\linewidth]{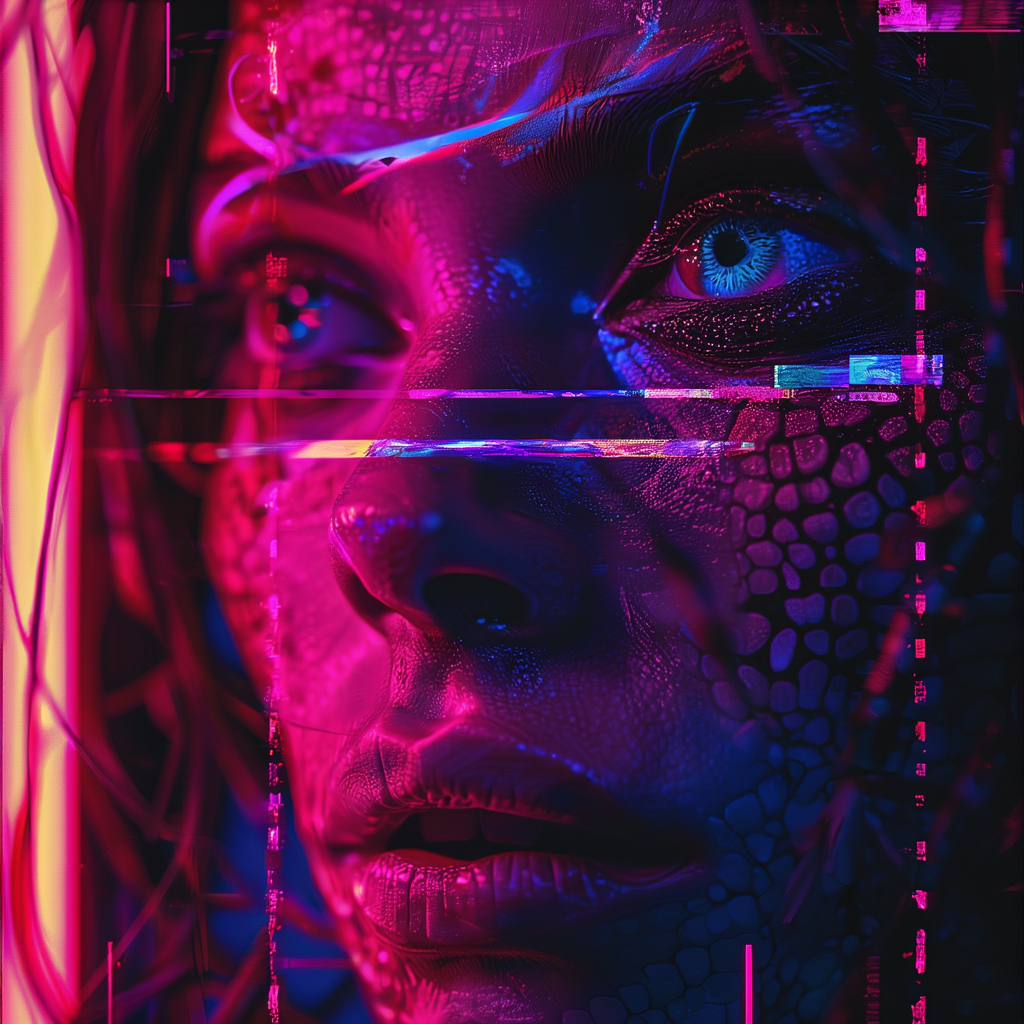} & \includegraphics[width=0.30\linewidth]{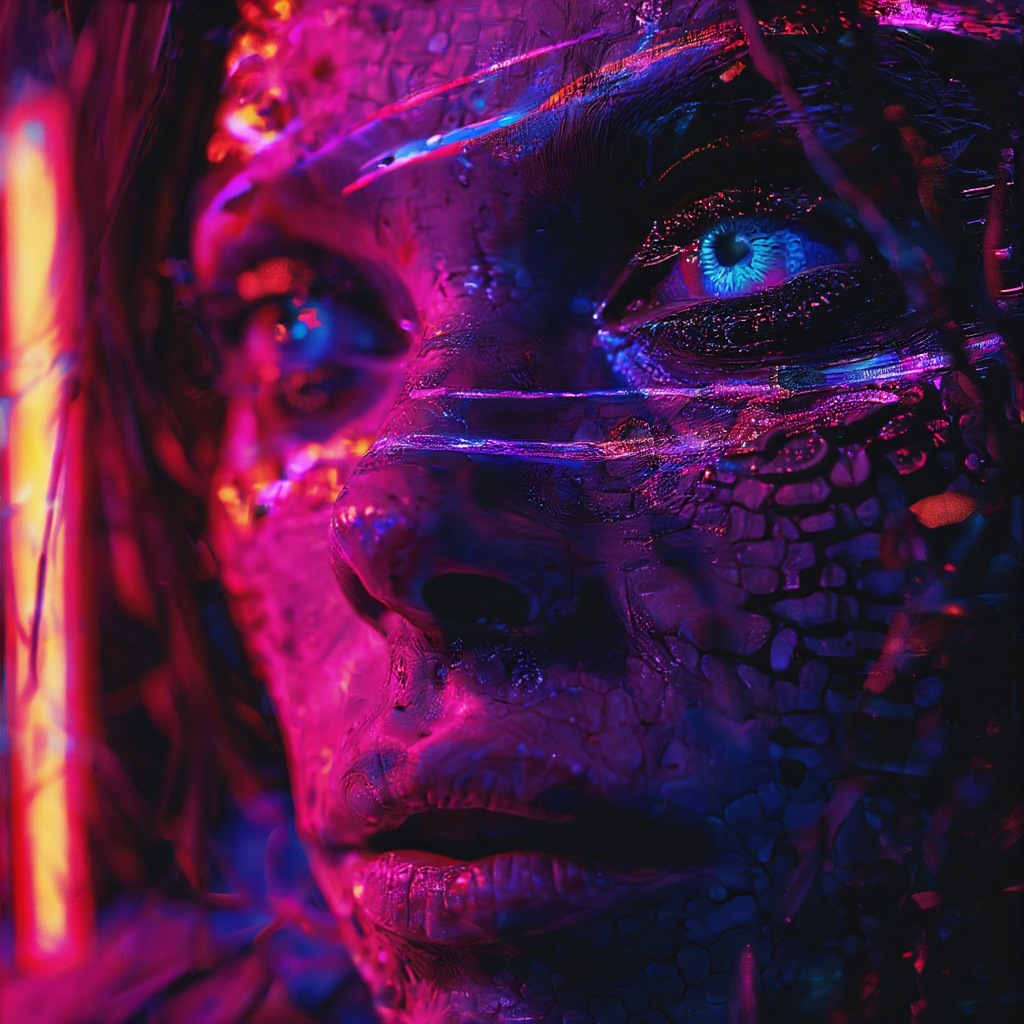} & \includegraphics[width=0.30\linewidth]{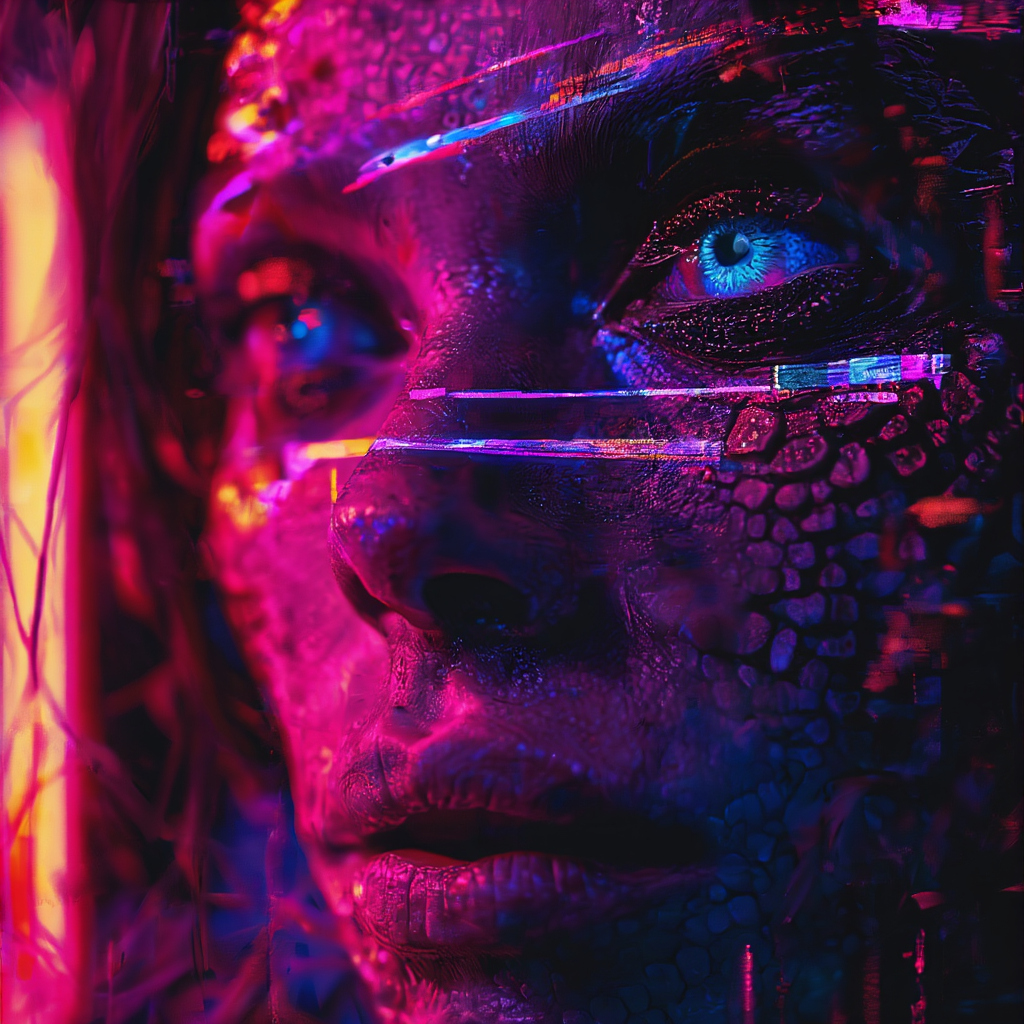}\\[2pt]
 & \multicolumn{2}{c}{Q-DiT}\\[1pt]
 & \includegraphics[width=0.30\linewidth]{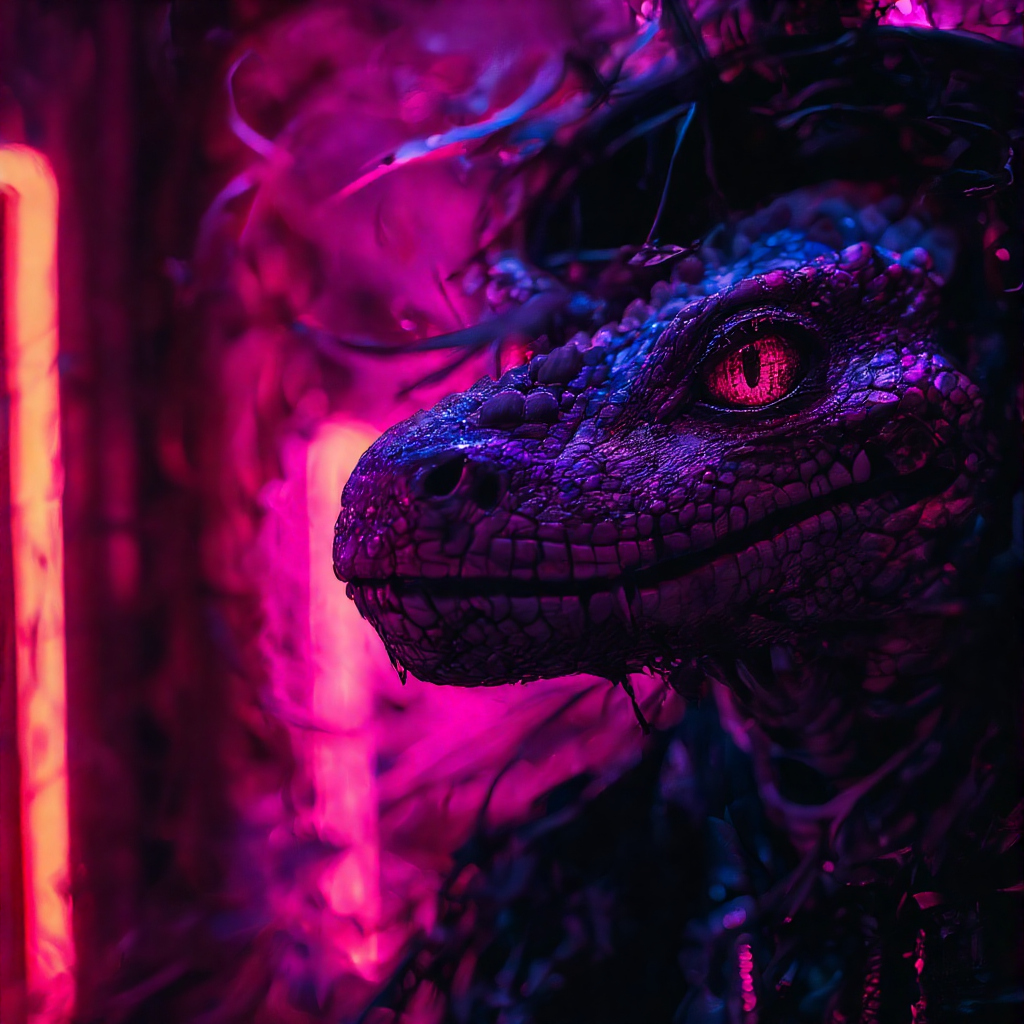} & \includegraphics[width=0.30\linewidth]{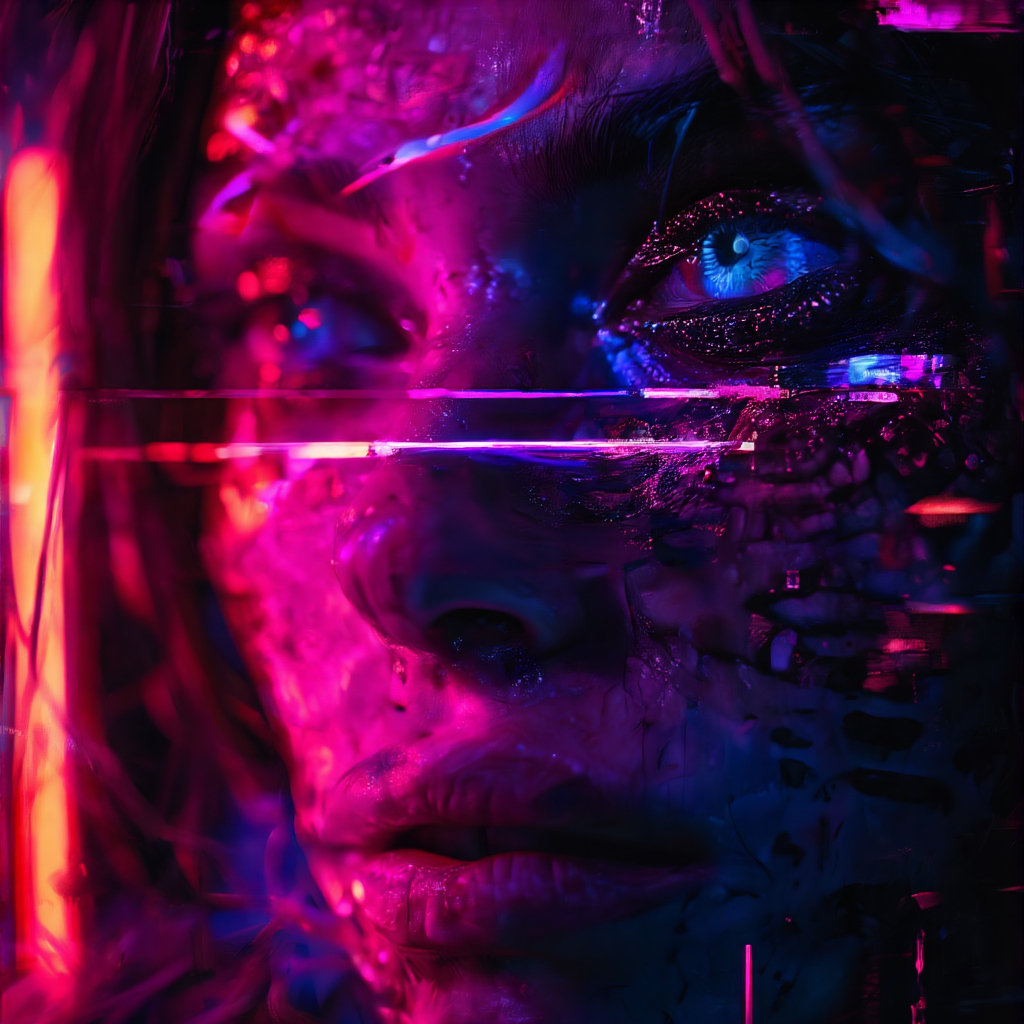}\\[2pt]
 & \multicolumn{2}{c}{MACaQ}\\[1pt]
 & \includegraphics[width=0.30\linewidth]{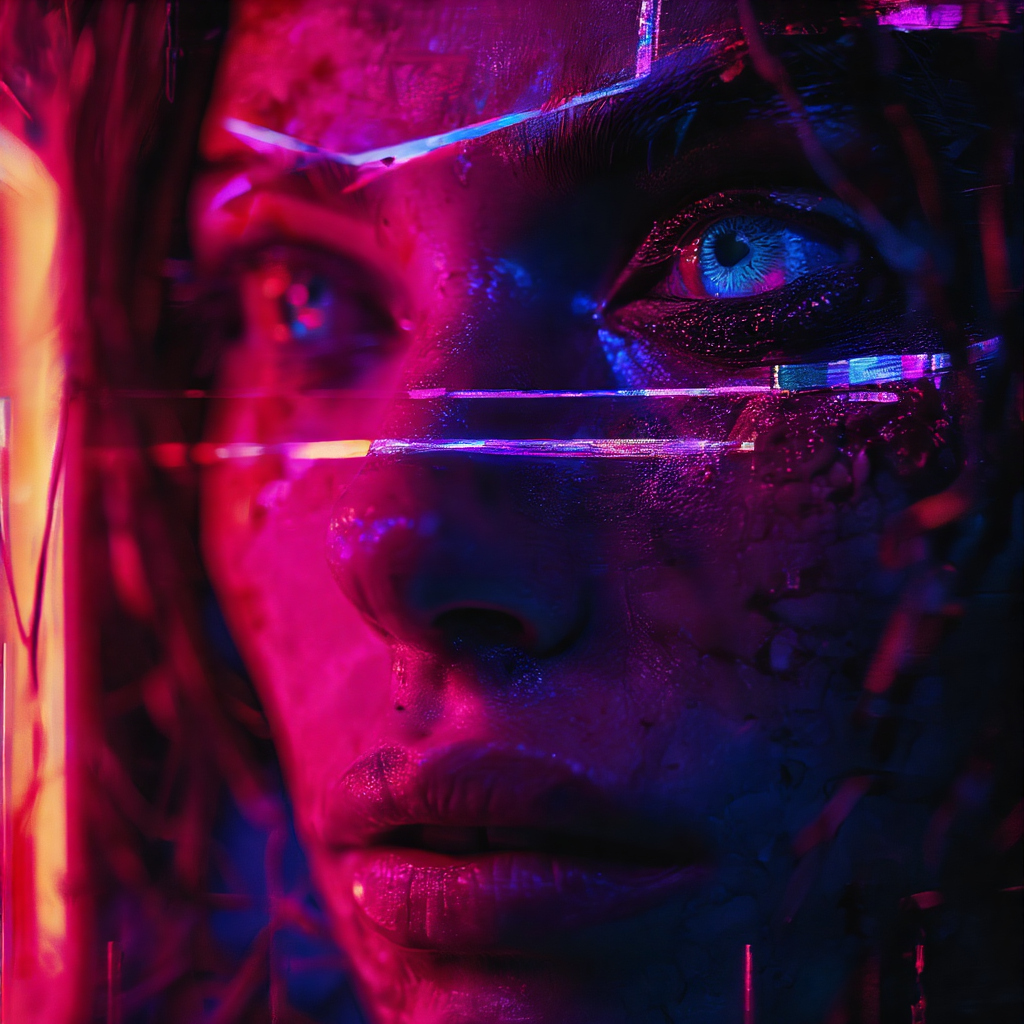} & \includegraphics[width=0.30\linewidth]{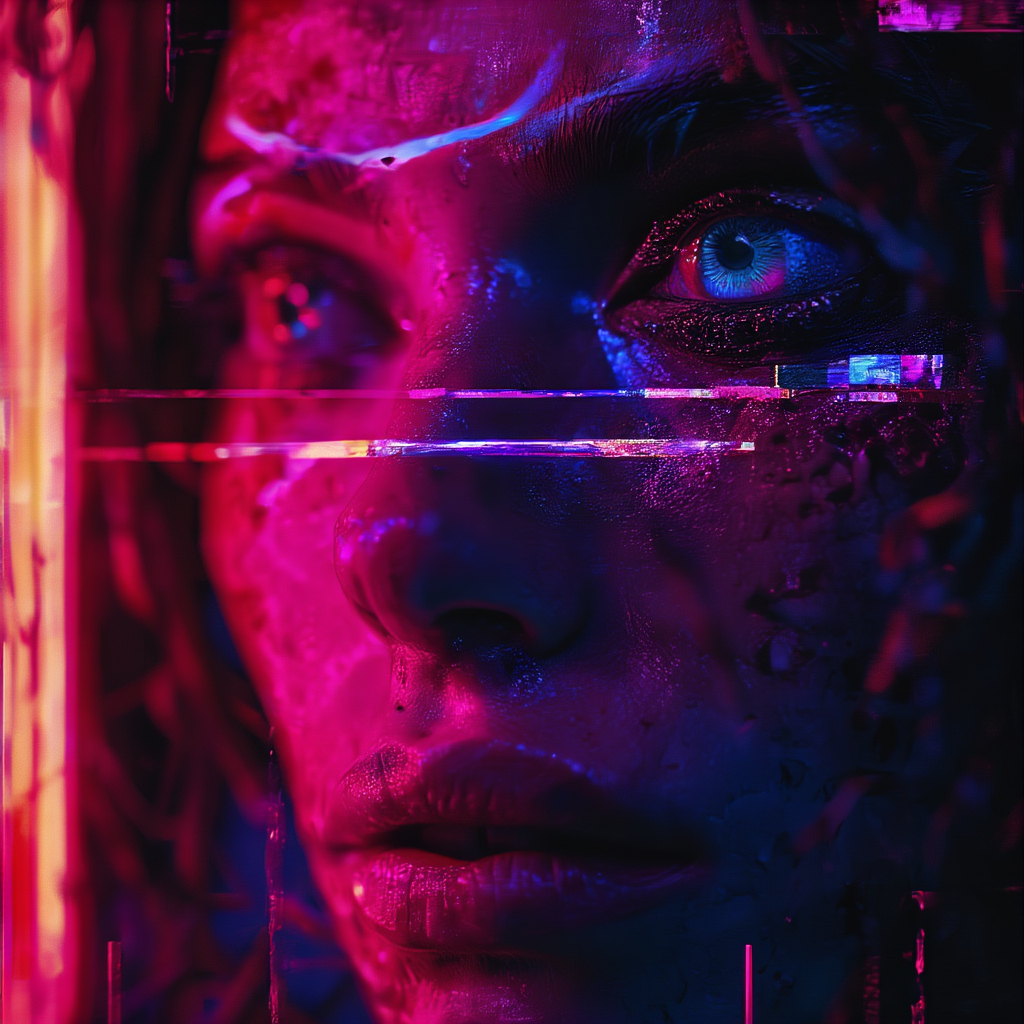}\\[2pt]
\end{tabular}
\endgroup
\end{minipage}
\par\smallskip
\captionof{figure}{\textbf{Additional Chroma1-HD comparisons.} Two matched examples with BF16 references and all evaluated W4A4 quantizers, with and without D4D.}\label{fig:qual-extra-chroma}
\end{minipage}\par

\par\addvspace{10pt}
\noindent\begin{minipage}{\textwidth}
\noindent
\begin{minipage}[t]{0.48\textwidth}
\centering\textbf{Example 1}\par\smallskip
\begingroup\scriptsize\setlength{\tabcolsep}{2pt}
\begin{tabular}{@{}ccc@{}}
\textbf{BF16} & \shortstack{\textbf{W4A4}\\\textbf{without D4D}} & \shortstack{\textbf{W4A4}\\\textbf{with D4D}}\\[2pt]
 & \multicolumn{2}{c}{SVDQuant}\\[1pt]
\includegraphics[width=0.30\linewidth]{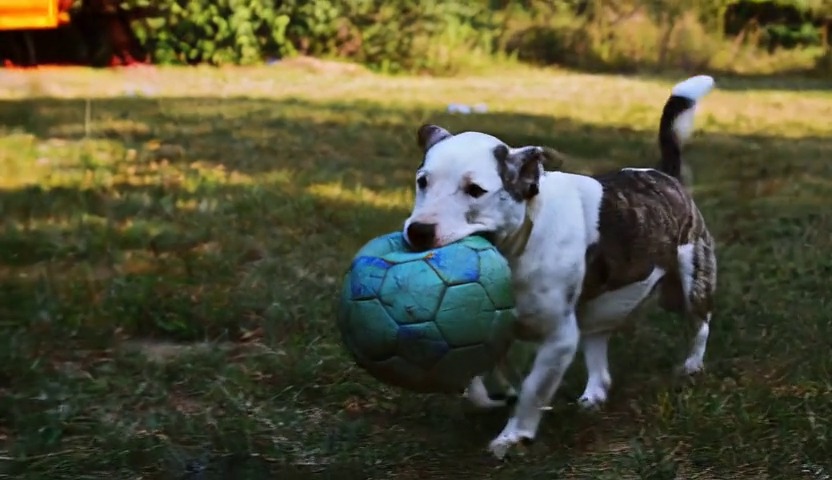} & \includegraphics[width=0.30\linewidth]{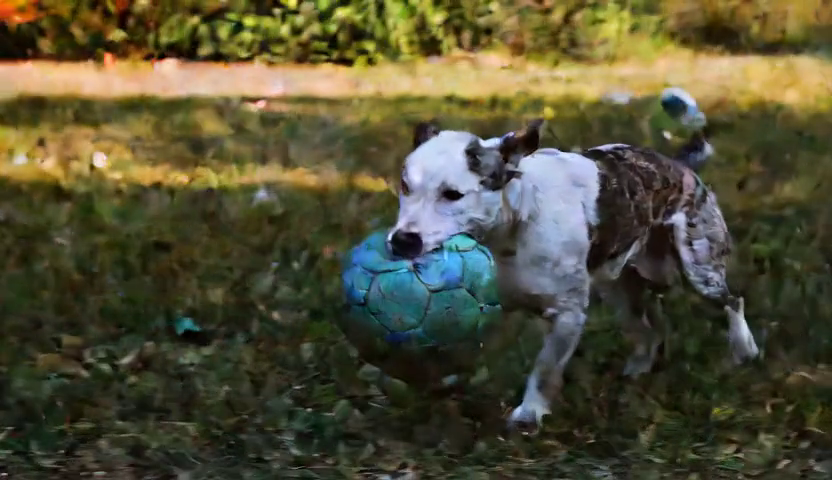} & \includegraphics[width=0.30\linewidth]{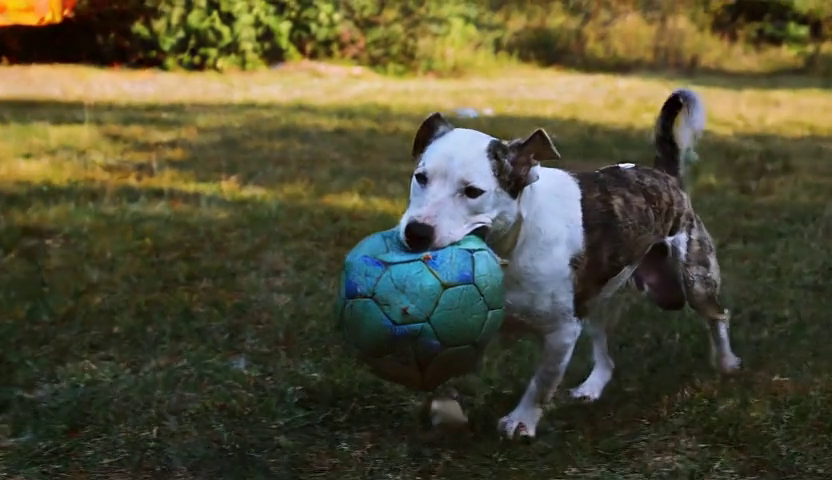}\\[2pt]
 & \multicolumn{2}{c}{Q-DiT}\\[1pt]
 & \includegraphics[width=0.30\linewidth]{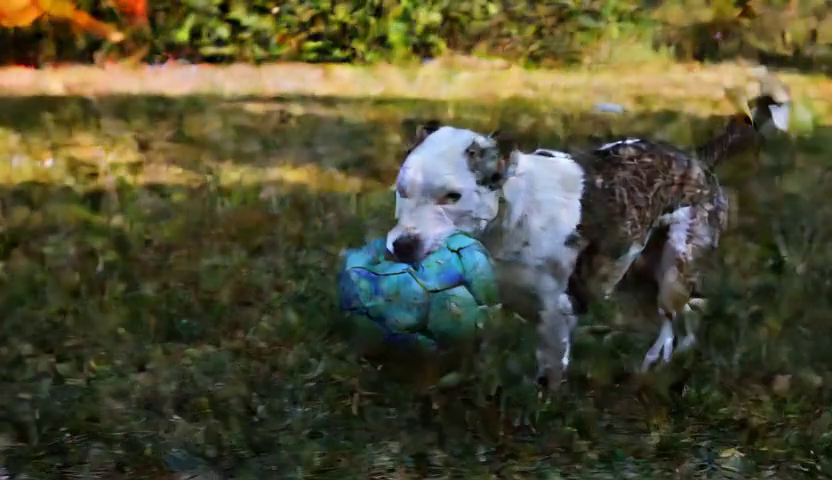} & \includegraphics[width=0.30\linewidth]{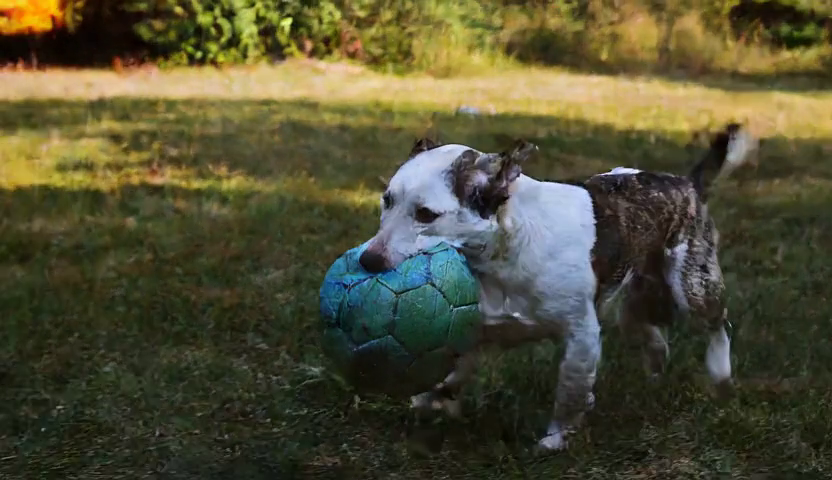}\\[2pt]
 & \multicolumn{2}{c}{MACaQ}\\[1pt]
 & \includegraphics[width=0.30\linewidth]{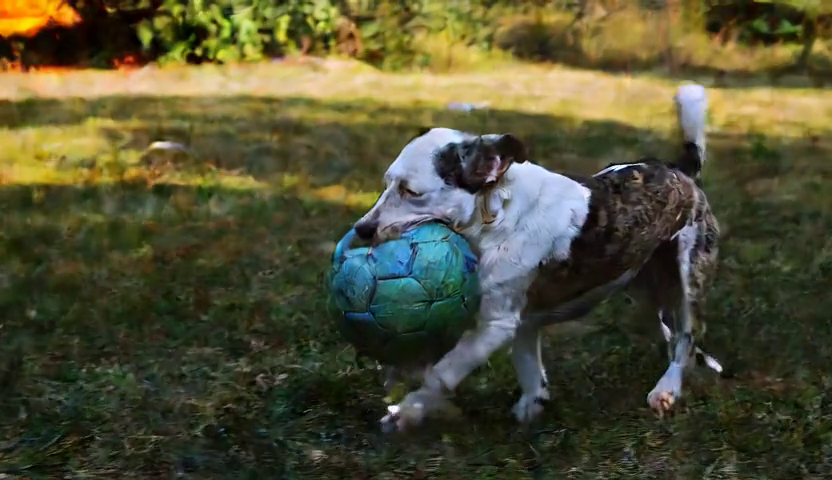} & \includegraphics[width=0.30\linewidth]{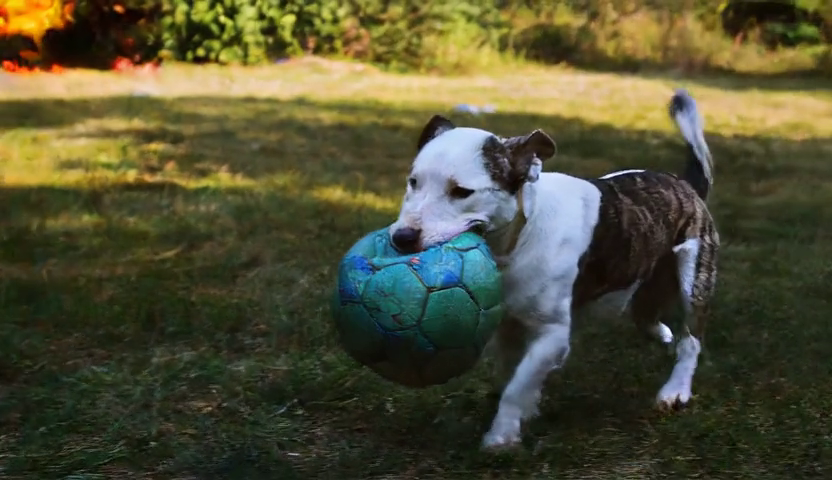}\\[2pt]
\end{tabular}
\endgroup
\end{minipage}
\hfill
\begin{minipage}[t]{0.48\textwidth}
\centering\textbf{Example 2}\par\smallskip
\begingroup\scriptsize\setlength{\tabcolsep}{2pt}
\begin{tabular}{@{}ccc@{}}
\textbf{BF16} & \shortstack{\textbf{W4A4}\\\textbf{without D4D}} & \shortstack{\textbf{W4A4}\\\textbf{with D4D}}\\[2pt]
 & \multicolumn{2}{c}{SVDQuant}\\[1pt]
\includegraphics[width=0.30\linewidth]{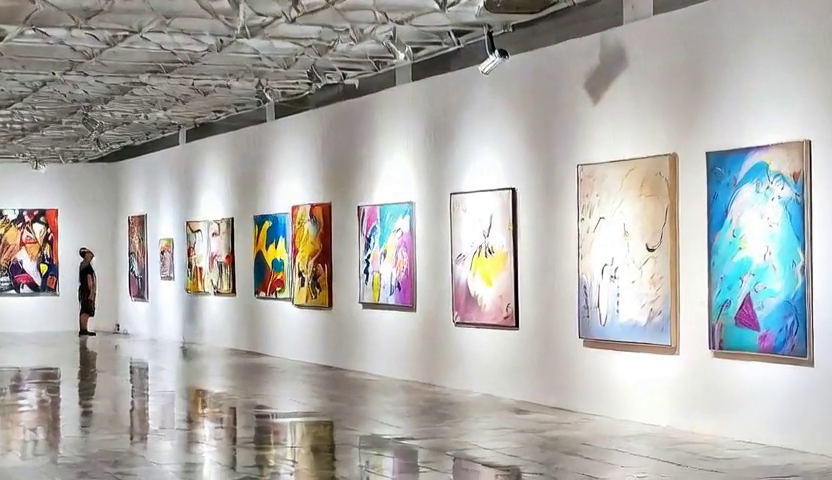} & \includegraphics[width=0.30\linewidth]{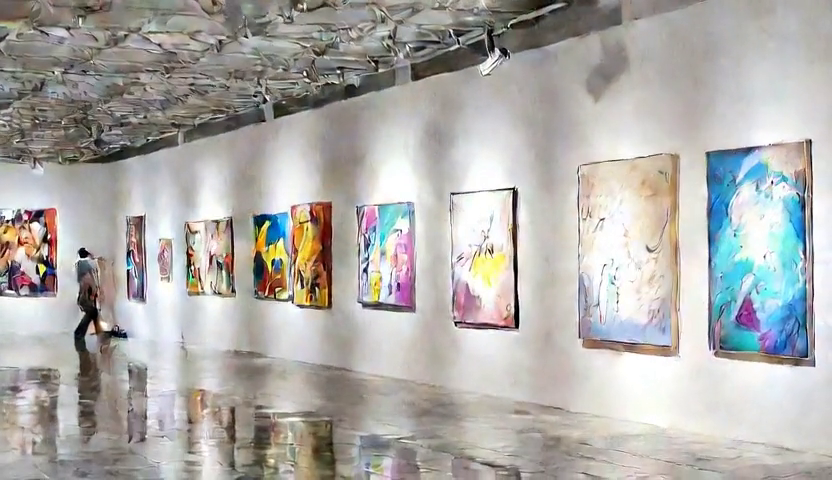} & \includegraphics[width=0.30\linewidth]{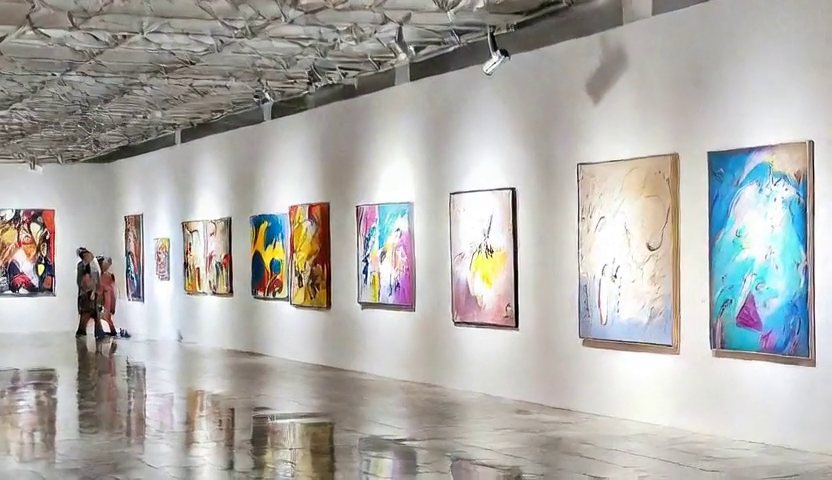}\\[2pt]
 & \multicolumn{2}{c}{Q-DiT}\\[1pt]
 & \includegraphics[width=0.30\linewidth]{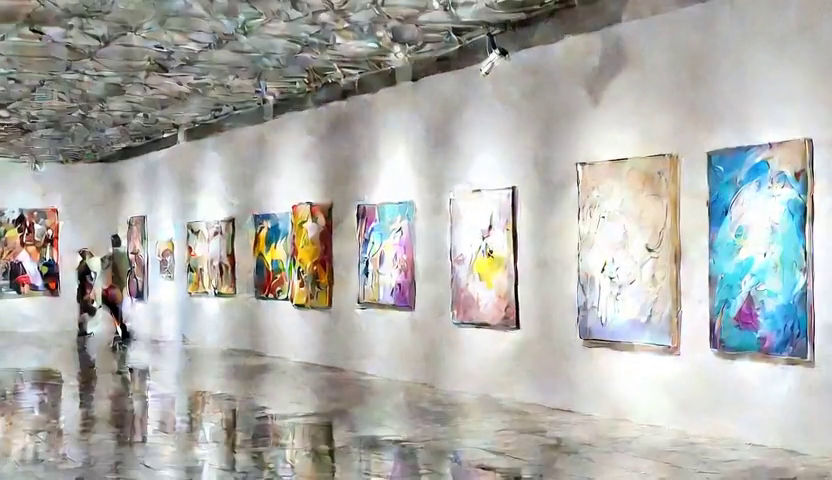} & \includegraphics[width=0.30\linewidth]{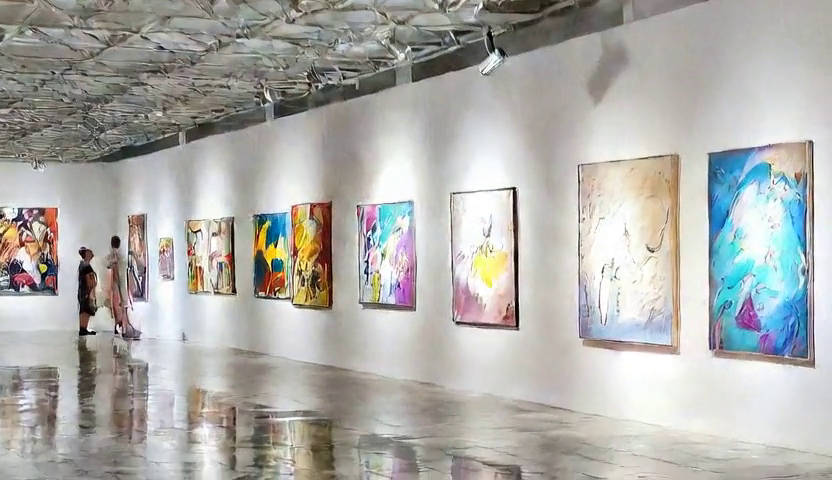}\\[2pt]
 & \multicolumn{2}{c}{MACaQ}\\[1pt]
 & \includegraphics[width=0.30\linewidth]{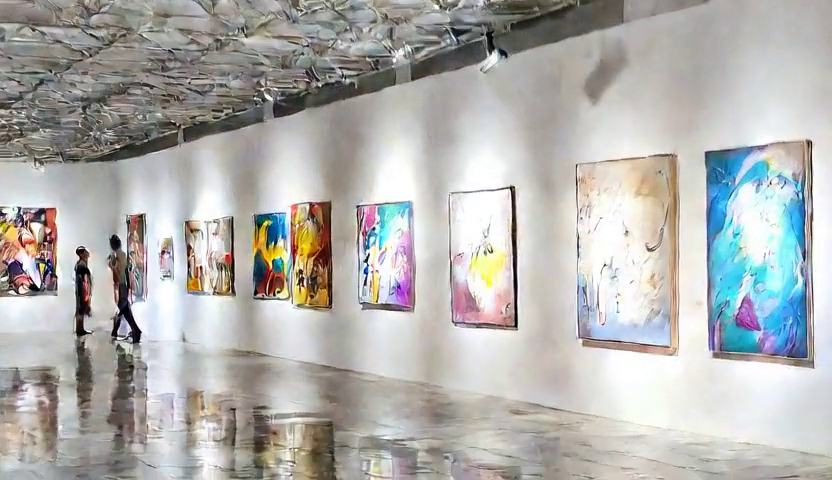} & \includegraphics[width=0.30\linewidth]{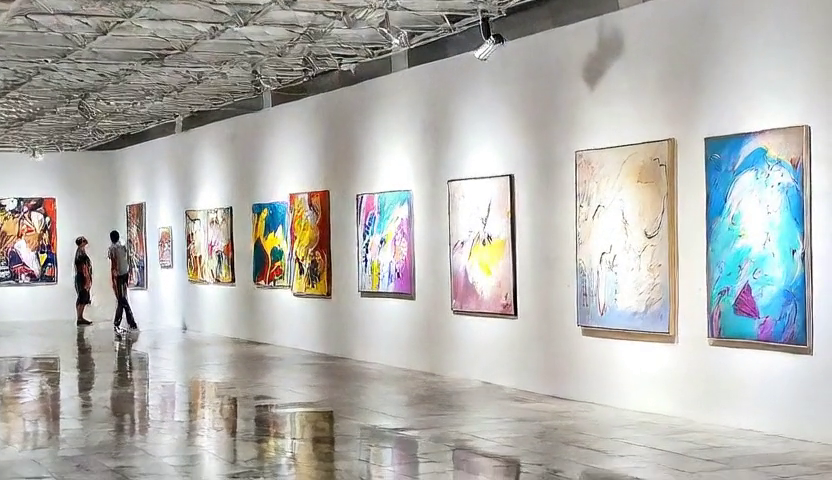}\\[2pt]
\end{tabular}
\endgroup
\end{minipage}
\par\smallskip
\captionof{figure}{\textbf{Additional Wan2.2-TI2V-5B comparisons.} Two examples with BF16 references and all evaluated W4A4 quantizers, with and without D4D. Each example shows the same video frame across methods, using the same conditioning image, prompt, and seed.}\label{fig:qual-extra-wan5}
\end{minipage}\par

\par\addvspace{10pt}
\noindent\begin{minipage}{\textwidth}
\noindent
\begin{minipage}[t]{0.48\textwidth}
\centering\textbf{Example 1}\par\smallskip
\begingroup\scriptsize\setlength{\tabcolsep}{2pt}
\begin{tabular}{@{}ccc@{}}
\textbf{BF16} & \shortstack{\textbf{W4A4}\\\textbf{without D4D}} & \shortstack{\textbf{W4A4}\\\textbf{with D4D}}\\[2pt]
 & \multicolumn{2}{c}{SVDQuant$^{*}$}\\[1pt]
\includegraphics[width=0.30\linewidth]{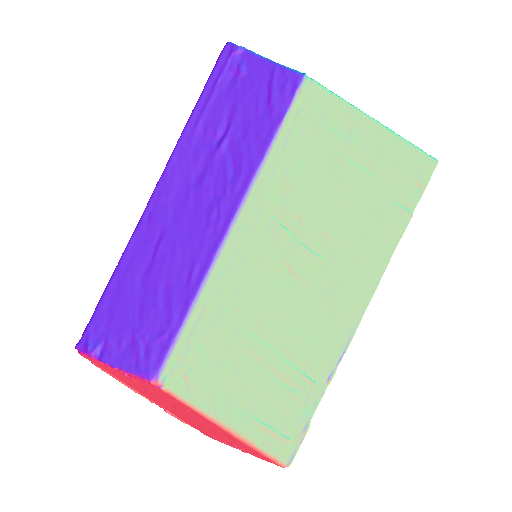} & \includegraphics[width=0.30\linewidth]{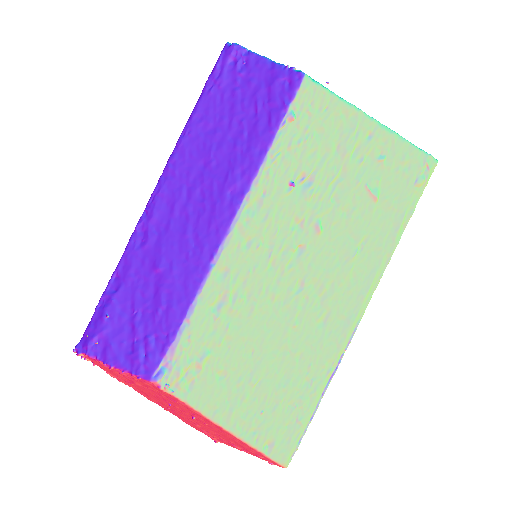} & \includegraphics[width=0.30\linewidth]{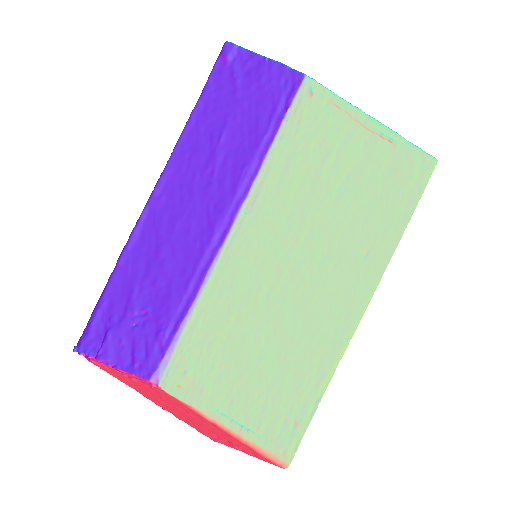}\\[2pt]
 & \multicolumn{2}{c}{MACaQ}\\[1pt]
\includegraphics[width=0.30\linewidth]{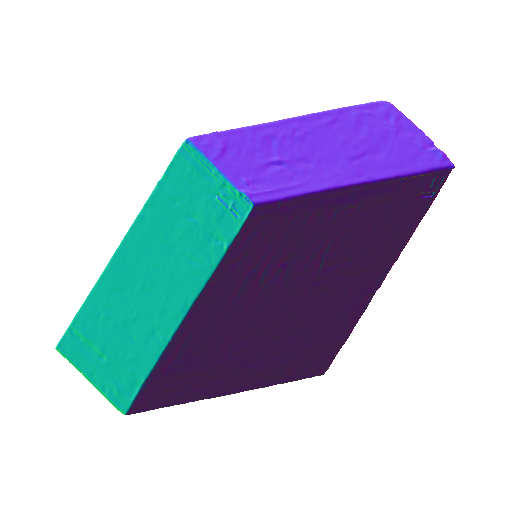} & \includegraphics[width=0.30\linewidth]{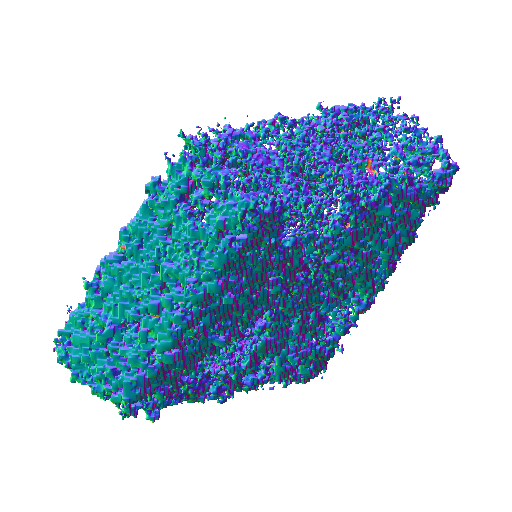} & \includegraphics[width=0.30\linewidth]{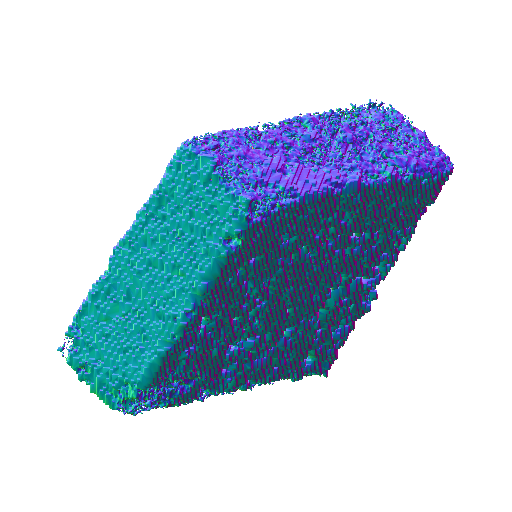}\\[2pt]
\end{tabular}
\endgroup
\end{minipage}
\hfill
\begin{minipage}[t]{0.48\textwidth}
\centering\textbf{Example 2}\par\smallskip
\begingroup\scriptsize\setlength{\tabcolsep}{2pt}
\begin{tabular}{@{}ccc@{}}
\textbf{BF16} & \shortstack{\textbf{W4A4}\\\textbf{without D4D}} & \shortstack{\textbf{W4A4}\\\textbf{with D4D}}\\[2pt]
 & \multicolumn{2}{c}{SVDQuant$^{*}$}\\[1pt]
\includegraphics[width=0.30\linewidth]{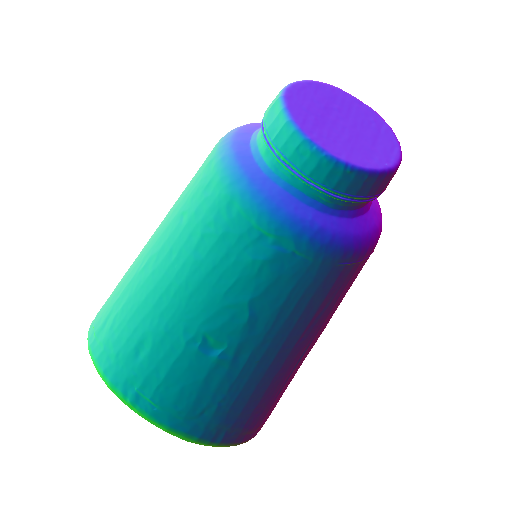} & \includegraphics[width=0.30\linewidth]{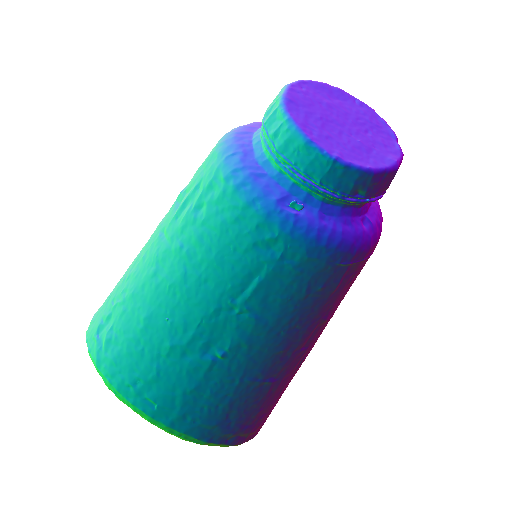} & \includegraphics[width=0.30\linewidth]{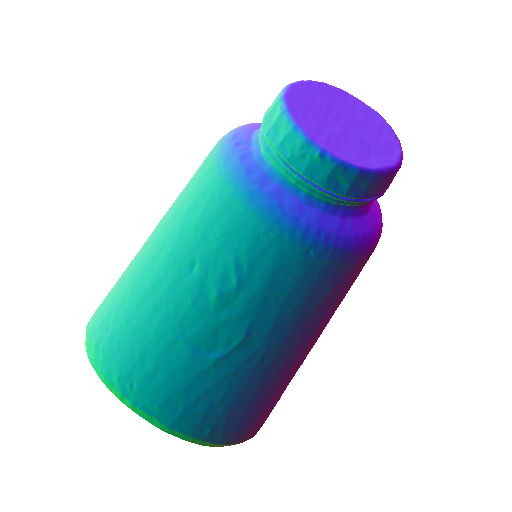}\\[2pt]
 & \multicolumn{2}{c}{MACaQ}\\[1pt]
\includegraphics[width=0.30\linewidth]{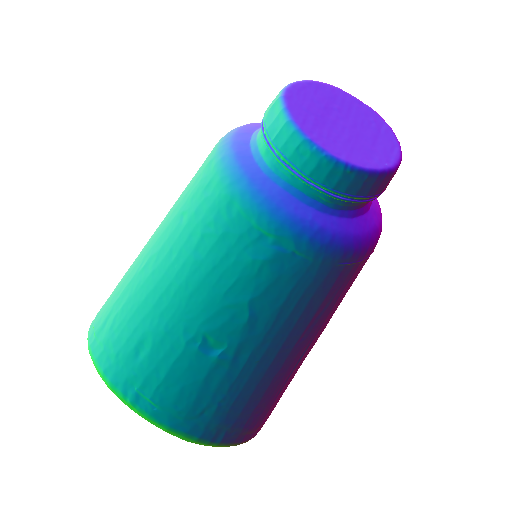} & \includegraphics[width=0.30\linewidth]{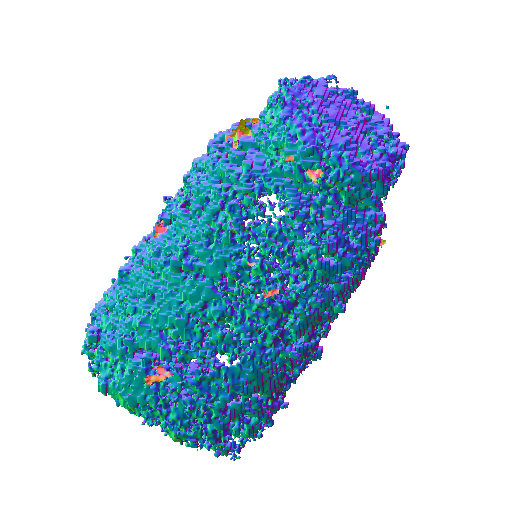} & \includegraphics[width=0.30\linewidth]{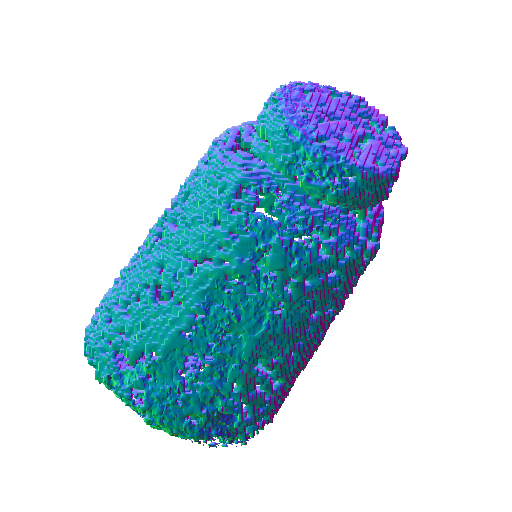}\\[2pt]
\end{tabular}
\endgroup
\end{minipage}
\par\smallskip
\captionof{figure}{\textbf{Additional Pixal3D comparisons.} Two examples with BF16 references and W4A4 quantizers, with and without D4D. Each row uses the same surface-normal view across methods. The asterisk denotes our SVDQuant adaptation.}\label{fig:qual-extra-pixal}
\end{minipage}\par

\par\addvspace{10pt}
\noindent\begin{minipage}{\textwidth}
\noindent
\begin{minipage}[t]{0.48\linewidth}\centering\textbf{Example 1}\par\smallskip
\begingroup\scriptsize\setlength{\tabcolsep}{2pt}\begin{tabular}{@{}ccc@{}}
\textbf{BF16} & \shortstack{\textbf{W4A4}\\\textbf{without D4D}} & \shortstack{\textbf{W4A4}\\\textbf{with D4D}}\\[2pt]
 & \multicolumn{2}{c}{SVDQuant$^{*}$}\\[1pt]
\includegraphics[width=0.30\linewidth]{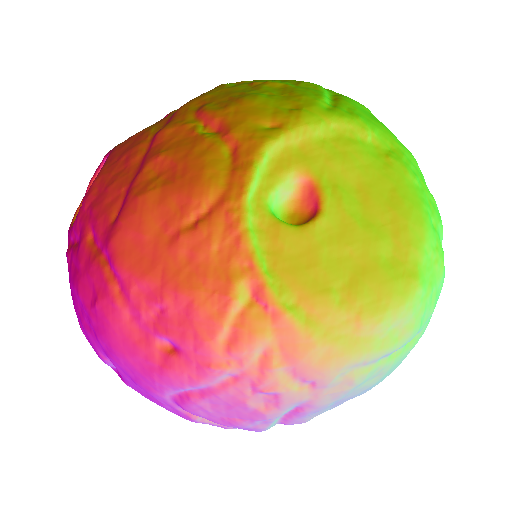} & \includegraphics[width=0.30\linewidth]{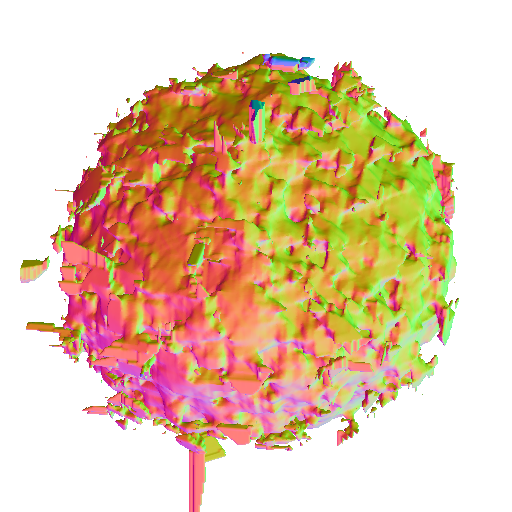} & \includegraphics[width=0.30\linewidth]{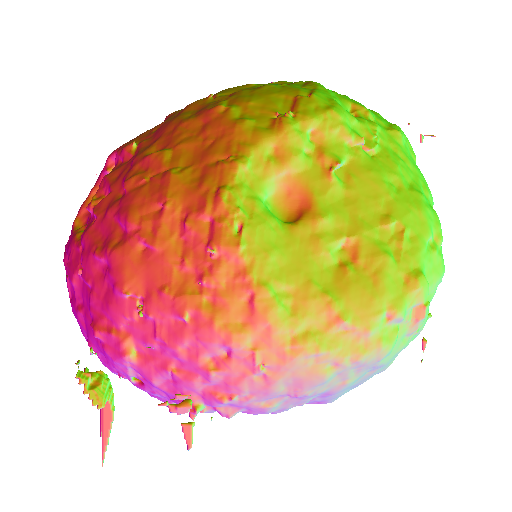}\\[2pt]
 & \multicolumn{2}{c}{MACaQ}\\[1pt]
 & \includegraphics[width=0.30\linewidth]{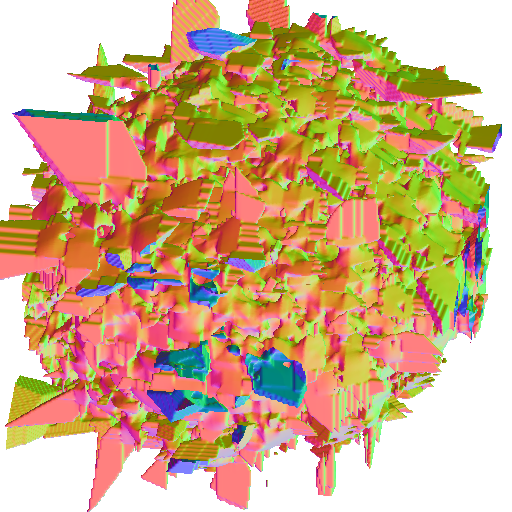} & \includegraphics[width=0.30\linewidth]{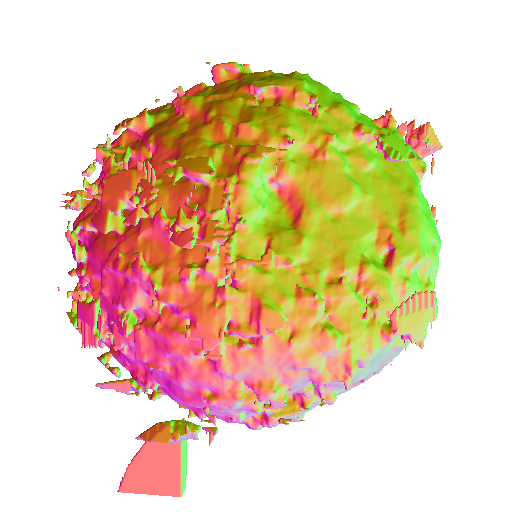}\\[2pt]
\end{tabular}\endgroup\end{minipage}
\hfill
\begin{minipage}[t]{0.48\linewidth}\centering\textbf{Example 2}\par\smallskip
\begingroup\scriptsize\setlength{\tabcolsep}{2pt}\begin{tabular}{@{}ccc@{}}
\textbf{BF16} & \shortstack{\textbf{W4A4}\\\textbf{without D4D}} & \shortstack{\textbf{W4A4}\\\textbf{with D4D}}\\[2pt]
 & \multicolumn{2}{c}{SVDQuant$^{*}$}\\[1pt]
\includegraphics[width=0.30\linewidth]{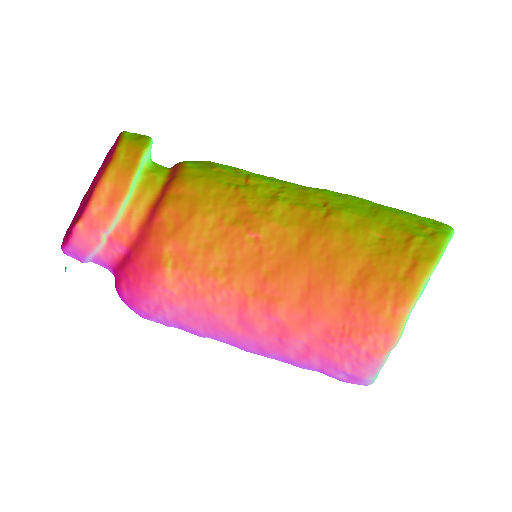} & \includegraphics[width=0.30\linewidth]{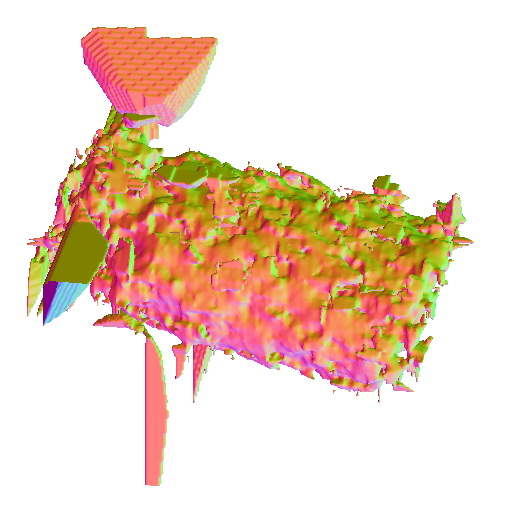} & \includegraphics[width=0.30\linewidth]{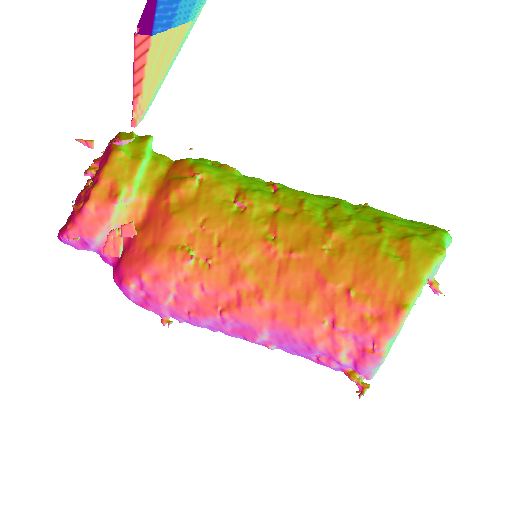}\\[2pt]
 & \multicolumn{2}{c}{MACaQ}\\[1pt]
 & \includegraphics[width=0.30\linewidth]{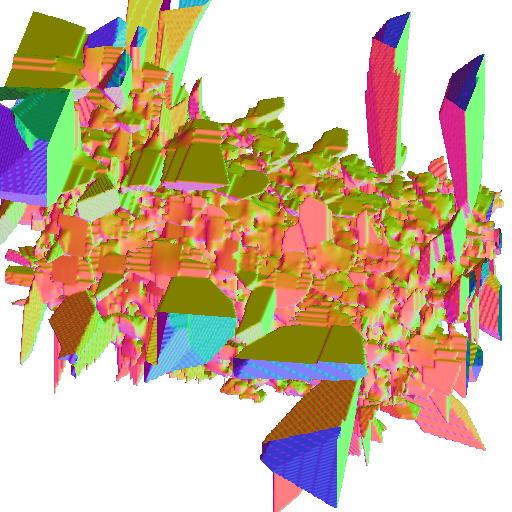} & \includegraphics[width=0.30\linewidth]{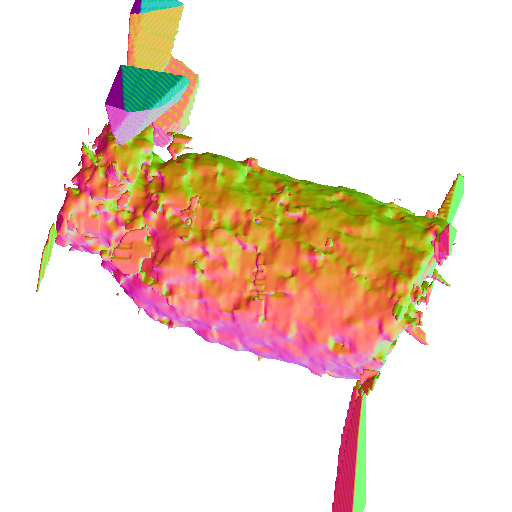}\\[2pt]
\end{tabular}\endgroup\end{minipage}
\par\smallskip
\captionof{figure}{\textbf{Additional 3DTopia-XL comparisons.} One surface-normal view per example, with and without D4D. Both base quantizers share the same BF16 reference. The asterisk denotes our SVDQuant adaptation.}\label{fig:qual-extra-topia}
\end{minipage}\par

\twocolumn
\section{Scope and Limitations}
\label{app:limitations}

\paragraph{Where D4D applies.}
D4D acts wherever both CFG branches of a sample pass through the same projection call with aligned tokens, the standard batched CFG of all evaluated models. Single-branch samplers, such as guidance-distilled, few-step or CFG-free models, have no branch pair and run their base quantizer unchanged; the same holds at steps where a sampler skips the unconditional branch. The benefit grows with the guidance scale; at $s=1$ the analysis predicts no benefit but a small loss (App.~\ref{sec:app_derivation}, App.~\ref{app:guidance-strength}). Serving the two branches on different devices would require exchanging activations at every projection and is not considered here. Within a model, text tokens keep branch coordinates because the two branches have different text inputs, and in the 3D models cross-attention keeps branch coordinates (App.~\ref{app:protocols}).

\paragraph{Analysis and evaluation.}
The analysis of Sec.~\ref{sec:bounds} predicts the local error reduction and its dependence on $\rho$ and $s$. After propagation through the denoiser, the measured reduction per projection is smaller (App.~\ref{sec:app_propagation}). The premise of the analysis, a small branch difference, holds on all evaluated models (App.~\ref{sec:app_branch_stats}), but the analysis does not predict the size of the end-to-end gain. D4D targets activation error; weight error is shared by both branches. End-to-end results are W4A4; lower activation error at 6 and 8 bits is shown locally (Fig.~\ref{fig:guided-error-gain}). Following common practice in diffusion PTQ, e.g.\ SVDQuant and ViDiT-Q, quality is measured mainly as fidelity to the matched BF16 output, complemented by ImageReward and VBench dimensions.

\end{document}